\documentclass[10pt,twoside,journal,compsoc]{IEEEtran}
\ifCLASSOPTIONcompsoc
  \usepackage[nocompress]{cite}
\else
  \usepackage{cite}
\fi
\ifCLASSINFOpdf
  \usepackage[pdftex]{graphicx}
\else
  \usepackage[dvips]{graphicx}
\fi
\usepackage[utf8]{inputenc} 
\usepackage[T1]{fontenc}    
\usepackage{hyperref}       
\usepackage{url}            
\usepackage{booktabs}       
\usepackage{amsfonts}       
\usepackage{nicefrac}       
\usepackage{microtype}      
\usepackage{algorithm}
\usepackage{subfigure}
\usepackage{listings}
\usepackage{wrapfig}
\usepackage{booktabs}
\usepackage{multirow}
\usepackage{amsmath}
\usepackage{threeparttable}
\usepackage{amsthm}
\usepackage{color}
\usepackage{pythonhighlight}
\newtheorem{Theorem}{Theorem}
\newtheorem{Lemma}{Lemma}

\usepackage{amssymb}
\usepackage{xcolor}

\usepackage{algorithmic}
\begin{document}
	
%
\title{MLR-SNet: Transferable LR Schedules for Heterogeneous Tasks}
%
%
%
%

\author{Jun~Shu*,
        Yanwen~Zhu*,
        Qian~Zhao,
        Zongben~Xu,
        and~Deyu~Meng
\IEEEcompsocitemizethanks{\IEEEcompsocthanksitem Jun Shu, Yanwen Zhu, Qian Zhao, Zongben Xu and Deyu Meng (corresponding author) are with School of Mathematics and Statistics and Ministry
	of Education Key Lab of Intelligent Networks and Network Security, Xi'an Jiaotong University, Shaanxi, P.R.China.\protect\\
Email: xjtushujun@gmail.com, zywwyz@stu.xjtu.edu.cn, {timmy.zhaoqian,
	zbxu, dymeng}@mail.xjtu.edu.cn.\protect\\
* indicates equal contribution}}


%
%

\markboth{Journal of \LaTeX\ Class Files,~Vol.~14, No.~8, August~2020}
{Journal of \LaTeX\ Class Files,~Vol.~14, No.~8, August~2020}
%



\IEEEtitleabstractindextext{%
\begin{abstract}
The learning rate (LR) is one of the most important hyper-parameters in stochastic gradient descent (SGD) algorithm for training deep neural networks (DNN). However, current hand-designed LR schedules need to manually pre-specify a fixed form, which limits their ability to adapt practical non-convex optimization problems due to the significant diversification of training dynamics. Meanwhile, it always needs to search proper LR schedules from scratch for new tasks, which, however, are often largely different with task variations, like data modalities, network architectures, or training data capacities. To address this learning-rate-schedule setting issues, we propose to parameterize LR schedules with an explicit mapping formulation, called \textit{MLR-SNet}. The learnable parameterized structure brings more flexibility for MLR-SNet to learn a proper LR schedule to comply with the training dynamics of DNN. Image and text classification benchmark experiments substantiate the capability of our method for achieving proper LR schedules. Moreover, the explicit parameterized structure makes the meta-learned LR schedules capable of being transferable and plug-and-play, which can be easily generalized to new heterogeneous tasks. We transfer our meta-learned MLR-SNet to query tasks like different training epochs, network architectures, data modalities, dataset sizes from the training ones, and achieve comparable or even better performance compared with hand-designed LR schedules specifically designed for the query tasks. The robustness of MLR-SNet is also substantiated when the training data are biased with corrupted noise. We further prove the convergence of the SGD algorithm equipped with LR schedule produced by our MLR-Net, with the convergence rate comparable to the best-known ones of the algorithm for solving the problem.
\end{abstract}

}

\maketitle
\IEEEdisplaynontitleabstractindextext

%
\IEEEpeerreviewmaketitle

\IEEEraisesectionheading{\section{Introduction}\label{introduction}}

\IEEEPARstart{S}{tochastic} gradient descent (SGD) and its many variants \cite{robbins1951stochastic,duchi2011adaptive,zeiler2012adadelta,tieleman2012lecture,kingma2014adam}, have been served as the cornerstone of modern machine learning with big data. It has been empirically shown that DNNs achieve state-of-the-art generalization performance on a wide variety of tasks when trained with SGD \cite{zhang2016understanding}. Recent researches observe that SGD tends to select the so-called flat minima, which seems to generalize better in practice, partially explaining its underlying working mechanism  \cite{hochreiter1997flat,keskar2016large,dinh2017sharp,wu2018sgd,izmailovaveraging,li2018visualizing}.

Scheduling learning rate (LR) for the SGD algorithm is one of the most widely studied aspects to help improve the training for DNNs. Specifically, it has been experimentally studied how the LR \cite{jastrzkebski2017three} essentially influences minima solutions found by SGD. This issue is also investigated from the theoretical perspective. For example, Wu et al., \cite{wu2018sgd} theoretically analyzed that LR plays an important role in minima selection from a dynamical stability perspective. Furthermore, they used stochastic differential equations to prove that the higher the ratio of the LR to the batch size, the flatter minimum inclines to be selected. Besides, He et al., \cite{he2019control} provided PAC-Bayes generalization bounds for DNN trained by SGD, which are highly correlated with LR. In summary, it is being more widely recognized that designing a proper LR schedule tends to highly influence the generalization performance of DNN training result \cite{bengio2012practical,schaul2013no,nar2018step,liu2020stochastic}.

\begin{figure*}[t]\vspace{-0mm}
	\centering
	\vspace{-5mm}
	\subfigcapskip=-0mm
	
	\subfigure[\scriptsize{Pre-set LR schedule on image dataset}]{
		\label{fig1d} 
		\includegraphics[width=0.32\textwidth]{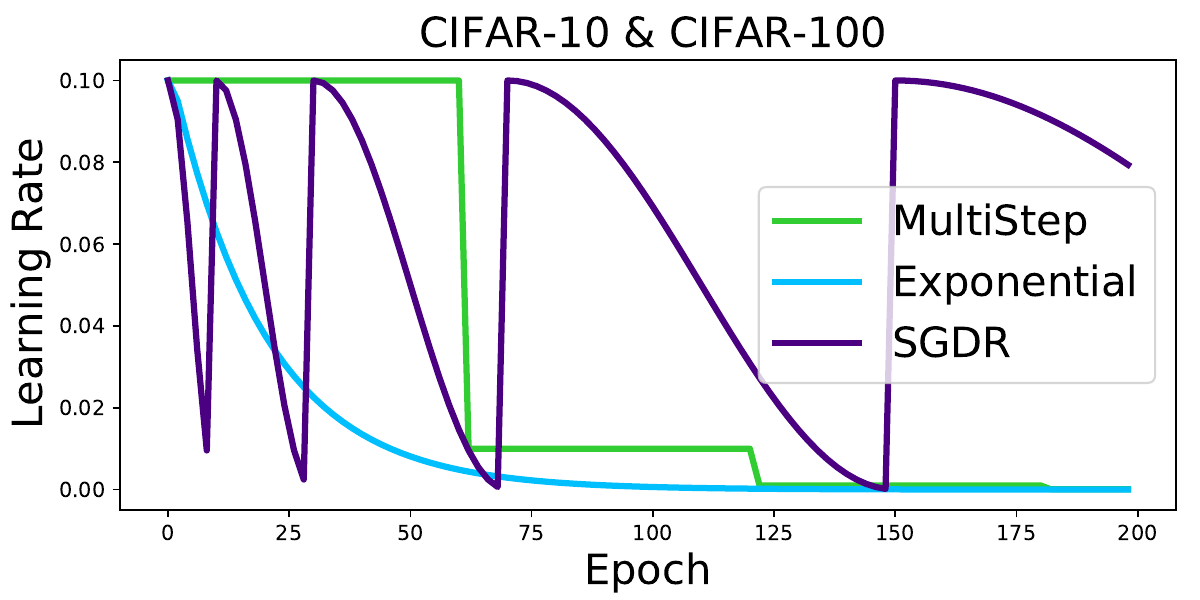}}
	\subfigure[\scriptsize{Pre-set LR schedule on text dataset}]{
		\label{fig1e} 
		\includegraphics[width=0.32\textwidth]{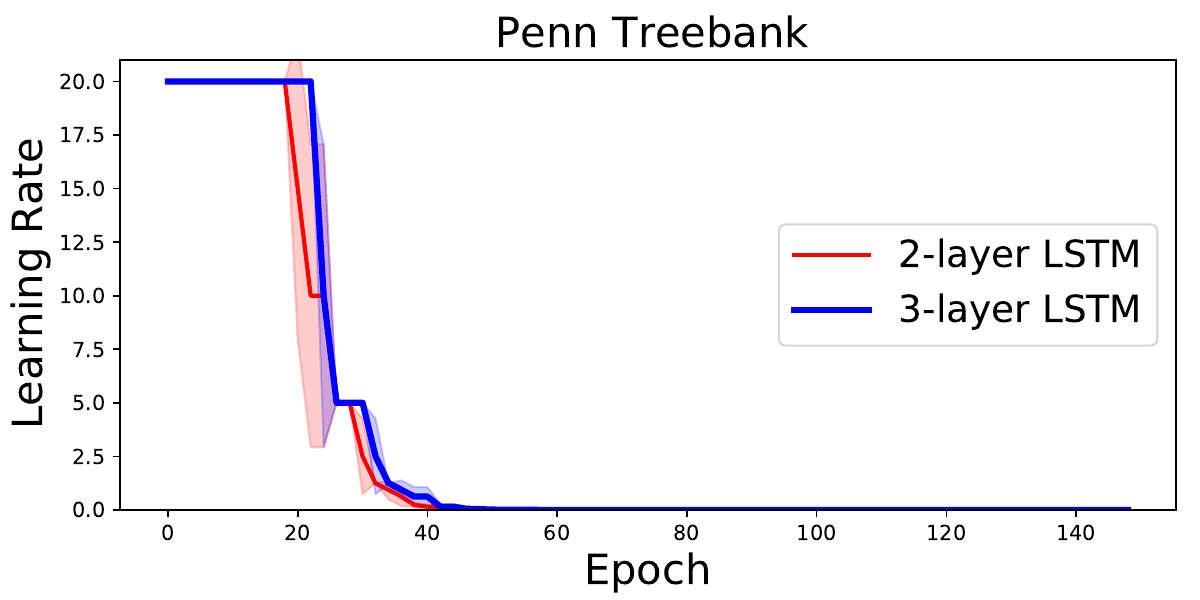}}
	\subfigure[\scriptsize{Diagram of MLR-SNet principle}]{
		\label{fig1a} 
		\includegraphics[width=0.32\textwidth]{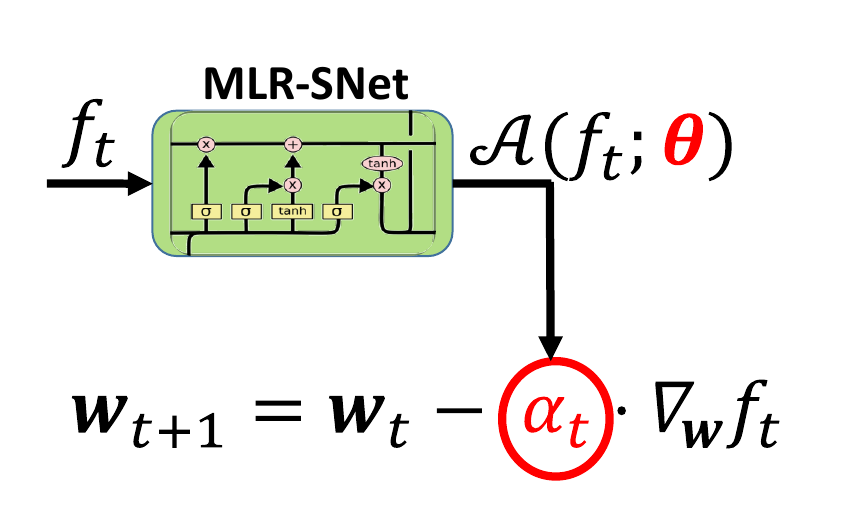}} \\ \vspace{-1mm}
	\subfigure[\scriptsize{LR schedule learned on image dataset}]{
		\label{fig1c} 
		\includegraphics[width=0.32\textwidth]{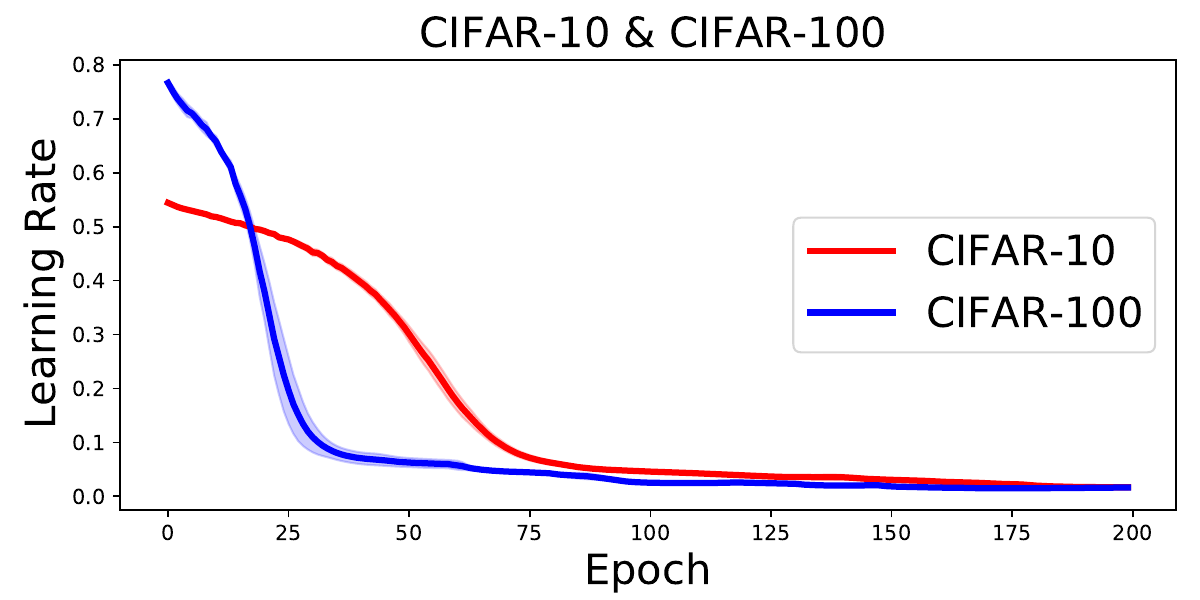}}
	\subfigure[\scriptsize{LR schedule learned on text dataset}]{
		\label{fig1b} 
		\includegraphics[width=0.32\textwidth]{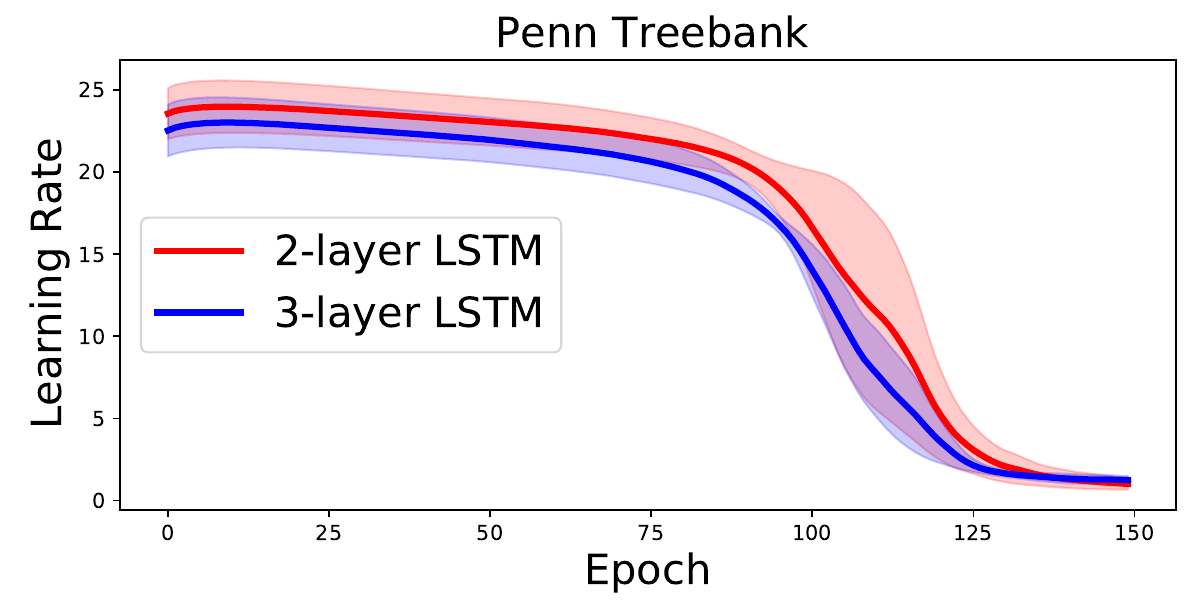}}
	\subfigure[\scriptsize{Predicted LR schedules by MLR-SNet}]{
		\label{fig1f} 
		\includegraphics[width=0.32\textwidth]{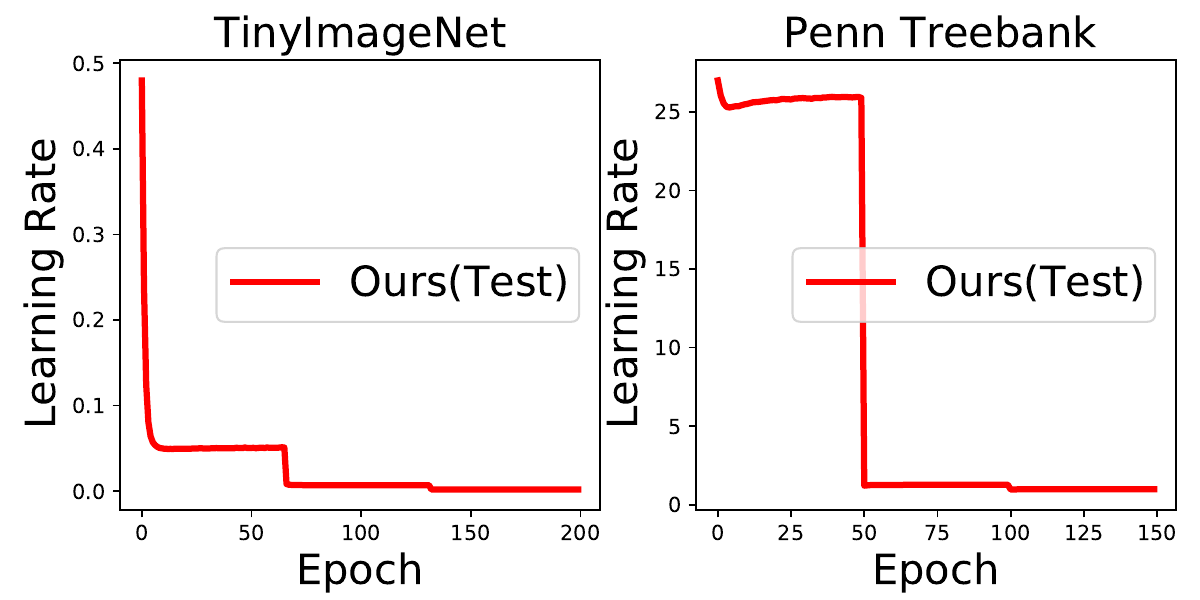}}  \vspace{-1mm}
	\caption{Pre-defined LR schedules used in our paper for (a) image and (b) text classification experiments. (c) Visualization of how we input current loss $f_t$ to MLR-SNet, which then outputs a proper LR $\alpha_t$ to help SGD find a better minima. LR schedules meta-learned by the proposed MLR-SNet on (d) image and (e) text classification experiments (meta-training stage). (f) The predicted LR schedules, learned from CIFAR-10, on image (TinyImageNet) and text (Penn Treebank) classification datasets (meta-test stage).
	}\label{ss} \vspace{-1mm}
\end{figure*}

There mainly exist three kinds of hand-designed LR schedules: (1) Pre-defined LR schedule policies. Typical ones include decaying and cyclic LR \cite{gower2019sgd,loshchilov2016sgdr} (as depicted in Fig. \ref{fig1d} and \ref{fig1e}), with a good training efficiency in practice. This line of methods have been mostly used in current DNN training, and become the default setting across the current popular deep learning libraries like Pytorch \cite{paszke2019pytorch}. Some theoretical works have further proved that the decaying schedule can yield faster convergence \cite{ge2019step,davis2019stochastic} or avoid strict saddles \cite{lee2017first,panageas2019first} under some mild conditions. 
(2) Adaptive gradient descend methods. Typical methods in this category include AdaGrad \cite{duchi2011adaptive}, RMSProp \cite{tieleman2012lecture}, and Adam \cite{kingma2014adam}, often using the adaptive LR for each model parameters based on some gradient information.
(3) LR search methods. The main idea is to borrow LR search strategies, , such as Polyak’s update rule \cite{rolinek2018l4}, Frank-Wolfe algorithm \cite{berrada2018deep}, and Armijo line-search \cite{vaswani2019painless}, used in traditional optimization approaches \cite{nocedal2006numerical} to DNN training, by searching LR adaptively in each updating step. 

Although above LR schedules can achieve competitive results on certain learning tasks, they still have evident deficiencies in practice. On the one hand, these policies need to manually pre-specify the formulation of the LR schedules, inevitably suffering from the limited flexibility to adapt to the complicated DNN optimization problems due to the significant variation of its training dynamics.
On the other hand, when solving new heterogeneous tasks, it always needs to redesign proper LR schedules from scratch, as well as to tune their involved hyperparameters.
This process is often time and computation expensive, which tends to further raise their application difficulty in real problems.

To alleviate the aforementioned issues, this paper aims to develop a model to learn a plug-and-play LR schedule under the meta-learning framework. 
The main idea is to parameterize the LR schedule as an LSTM network \cite{hochreiter1997long}, which is capable of dealing with such a long-term information dependent problem. As shown in Fig. \ref{fig1a}, with a parameterized structure, the proposed model has the capacity to fit an explicit loss-LR dependent relationship to adapt the complicated training dynamics.
We learn the LSTM network from data in a meta-learning manner, which is able to adaptively predict the LR schedule for a SGD algorithm to help improve the DNN training performance. We call
this method Meta-LR-Schedule-Net (\textbf{\emph{MLR-SNet}} for brevity). Meanwhile, the parameterized structure makes it possible to transfer the meta-learned LR schedule to be readily used in new query tasks.
In a nutshell, this paper mainly makes the following five-fold contributions.

(1) The MLR-SNet is proposed to learn an adaptive LR schedule for SGD algorithm, which is capable of dynamically adjusting LR during the DNN training process based on current training loss as well as the information delivered from past training histories stored in the MLR-SNet. Due to the explicit parameterized formulation of the MLR-SNet, it can be more flexible than hand-designed policies to find a proper LR schedule for specific learning tasks.

(2) The proposed model is model-agnostic, and can be applied to the SGD implementation on general DNN models. That is naturally feasible since the proposed MLR-SNet is with general loss information as its inputs, which is independent from the structure of the DNN models. The MLR-SNet is thus able to be generally applied to different DNN training problems, e.g., image and text classification problems, as shown in Fig.\ref{fig1c} and \ref{fig1b}. It can be seen that the meta-learned LR schedules have similar tendency as specifically pre-defined ones, as depicted in Fig. \ref{fig1d} and \ref{fig1e}, but with more adaptive variations at their locality. This validates the capability and efficacy of our method for adaptively scheduling LR.


(3) With an explicit parameterized structure, it is possible to readily transfer the meta-trained MLR-SNet for helping schedule LR of SGD on new heterogeneous tasks. Different from hand-designed LR schedules often requiring to re-design the LR schedules or re-tune the hyperparameters for new query tasks, the meta-learned MLR-SNet is plug-and-play, and without additional hyper-parameters to tune. To verify this point, we transfer the meta-learned MLR-SNet to different training epochs, datasets and network architectures, and achieve comparable performance with the corresponding best hand-designed LR schedules in the test data. Since it is directly employed as a off-the-shelf LR-schedule setting function, it is with similar computational complexity as the hand-designed LR schedules. Besides, it has been empirically verified that the generalization performance of meta-learned MLR-SNet is slightly related to the size of meta-training dataset, while relatively weakly related to the similarity between meta-training and meta-test tasks and DNN models.
This reveals the potential of transferring meta-learned LR schedules to improve the DNN training for the unseen tasks, and hopeful to save large labor and computation cost for DNN training in more real applications.

(4) The MLR-SNet is meta-learned to improve the generalization performance of the learned model on unseen data. We validate that with sound guidance of clean data as meta-data, our MLR-SNet can help achieve better robustness when training data are biased with corrupted noise than hand-designed LR schedules.

(5) We theoretical prove that the DNN models trained with the SGD algorithm, using LR schedules produced by our MLR-SNet, can obtain a convergence guarantee. Meanwhile, we can also prove the convergence guarantee for our MLR-SNet updated by the Adam algorithm guided by the validation loss under some mild conditions.

The paper is organized as follows. Section \ref{relate} reviews the related works. Section \ref{section2} presents the MLR-SNet model as well as its learning algorithm. Section \ref{result} demonstrates the experimental evaluations to validate the adaptability, transferability and robustness of the MLR-SNet, as compared with current LR schedules policies.  Section \ref{analysis} provides some analysis on MLR-SNet, e.g., its convergence and computational complexity. The paper is finally concluded.

\section{Related Work}\label{relate}
\textbf{Meta learning for optimization.} Meta learning, or learning to learn has a long history in psychology \cite{ward1937reminiscence,lake2017building}. Meta learning for optimization can date back to 1980s-1990s \cite{schmidhuber1992learning,bengio1991learning}, aiming to meta-learn the optimization process of learning itself. Inspired from such beneficial attempts, many researches were proposed to meta-learn the optimization process of different learning tasks. The early work is proposed by Schmidhuber et al.\cite{schmidhuber1992learning}, developing an end-to-end differentiable system to jointly train both the network and the learning algorithm by gradient descent, making the network able to modify its own weights. Bengio et al. \cite{bengio1991learning} also proposed to learn parameterized local neural net update rules that avoids back-propagation. Furthermore, Hochreiter et al. \cite{hochreiter2001learning} jointly train two networks, in which the output of back-propagation from one network was feed into an additional learning network to attain the learning algorithm.

Recently, \cite{andrychowicz2016learning,ravi2016optimization,chen2017learning,wichrowska2017learned,li2017learning,lv2017learning} have attempted to scale this idea to larger DNN optimization problems. The main idea is to construct a meta-learner as the optimizer, which takes the gradients as input and outputs the whole updating rules. These approaches tend to make selecting appropriate training algorithms, scheduling LR and tuning other hyper-parameters in an automatic way. The meta-learner of these approaches can be updated by minimizing the generalization error on the validation set. Furthermore, \cite{li2017learning} utilized reinforcement learning and \cite{ravi2016optimization} used test error of few-shot learning tasks to train the meta-learner.
Except for solving continuous optimization problems, some works employ these ideas to other optimization problems, such as black-box functions \cite{chen2017learning}, few-shot learning \cite{li2017meta,finn2017model}, model's curvature \cite{park2019meta}, evolution strategies \cite{houthooft2018evolved}, combinatorial functions \cite{rosenfeld2018learning}, MCMC Proposals \cite{wang2018meta}, etc.

Though faster in decreasing training loss than traditional optimizers in some cases, the learned optimizers by this line of methods always could not generalize well to varying problems from the training ones, especially longer horizons \cite{lv2017learning} and larger scale optimization problems \cite{wichrowska2017learned}. Moreover, these methods could not guarantee to output a proper descent direction in each iteration for DNN training, since they set the whole updating rules in SGD as the training variables, which might too flexible to soundly guide the training tendency of DNN parameters especially for meta-tested tasks. Comparatively, our proposed method attempts to learn an adaptive LR schedule for SGD algorithm, while sufficiently preserve the original gradient knowledge of the trained/tested problems. This not only makes the training afford of such meta-learning task capable of being largely alleviated and more stably executed, but also makes the meta-learned LR schedules easily and more accurately transferable to new heterogeneous tasks.

\textbf{HPO and LR schedule adaptation.} Hyper-parameter optimization (HPO) was historically investigated by selecting proper values for algorithm hyper-parameters to obtain better performance on validation set (see \cite{hutter2019automated} for an overview). Typical methods include grid search, random search \cite{bergstra2012random}, Bayesian optimization \cite{snoek2012practical}, gradient-based methods \cite{franceschi2017forward,shu2020learning,shu2020meta}, etc. Recently, some works attempt to find a proper LR schedule under the framework of gradient-based HPO, which can be solved by a bilevel optimization problem \cite{franceschi2017forward,baydin2017online}.
However, most HPO techniques for this task tends to directly learn the algorithm hyper-parameters against certain task while not predict their underlying setting rules across different tasks, making them easily fall into short-horizon bias and trapped into bad minima \cite{wu2018understanding}. Comparatively, our MLR-SNet is set as an explicit and concise function form to deliver the effective LR schedule setting principle among heterogeneous tasks, making it with better generality for general meta-tested tasks.

\textbf{Transfer to heterogeneous tasks.} Transfer learning \cite{pan2009survey} aims to transfer knowledge obtained from source task to help the learning on the target task. Most transfer learning approaches assume the source and target tasks consist of similar instances, features or model spaces \cite{yang2020transfer}, which greatly limits their application range. Recently, meta learning \cite{finn2017model} aims to learn common knowledge/methodology shared over observed tasks, such that the learned knowledge/methodology is expected to be transferred to unseen tasks. Similarly, our method aims to realize such a methodology-level transfer learning for the LR-schedule setting task, i.e., learn a general LR schedule predictor which is plug-and-play and easy to transfer to new query tasks. Such task-transferable capability, however, is not possessed by conventional hand-designed LR schedules and HPO methods.


%

\section{MLR-SNet} \label{section2}
The problem of training DNNs can be formulated as the following non-convex optimization problem,
\begin{align}
	\min_{w\in \mathbb{R}^d} f^{Tr}(D_{Tr};w)  := \frac{1}{N}\sum_{i=1}^N f_i^{Tr}(w),
\end{align}
where $f_i^{Tr}$ is the training loss function for data samples $i\in D_{Tr} = \{1,2,\cdots,N\}$, which characters the deviation of the model prediction from the data labels, and $w\in \mathbb{R}^d$ represents the parameters of the model (e.g., the weight matrices in the trained DNN) to be optimized. SGD \cite{robbins1951stochastic,polyak1964some} and its variants, including Momentum \cite{tseng1998incremental}, Adagrad \cite{duchi2011adaptive}, Adadelta \cite{zeiler2012adadelta}, RMSprop \cite{tieleman2012lecture}, Adam \cite{kingma2014adam}, are often used for DNN training. In general, these algorithms can be expressed as the following formulation,

\vspace{-4mm}\small
\begin{align}\label{eq2}
	w_{t+1} = w_{t} + \Delta w_{t}, \Delta w_{t} = \mathcal{O}_t(\nabla f^{Tr}_w(D_{Tr};w_t),\mathcal{H}_t;\Theta_t),
\end{align}\normalsize\vspace{-5mm}

\noindent where $w_{t}$ is $t$-th updating model parameters, $\nabla f^{Tr}_w(D_{Tr};w_t)$ denotes the gradient of $f^{Tr}$ at $w_t$, $\mathcal{H}_t$ represents the historical gradient information, and $\Theta_t$ is the hyperparameter of the optimizer $\mathcal{O}$, e.g., LR, in the current interation.
To present our method's efficiency, we focus on the following vanilla SGD algorithm in this paper\footnote{For different learning tasks, the commonly used optimizers are different. For example, image tasks often use SGD with Momentum, while text tasks always employ SGD or Adam. To guarantee the chosen optimizer able to be applied to various tasks, we learn the LR schedules for the vanilla SGD in this paper. We further validate that MLR-SNet can be applied to other optimizers, e.g., Adam (refer to Section \ref{adam}). },
\begin{align}\label{eq3}
	w_{t+1} = \xi_t(w_{t},\alpha_t)= w_{t} - \alpha_{t} \nabla_{w} f^{Tr}(D_t; w_t),
\end{align}
where $\nabla_{w} f^{Tr}(D_t; w_t) = \frac{1}{|D_t|}\sum_{i\in D_t}\nabla_w f_i^{Tr}(w_t)$, $D_t\!\subset\! D_{Tr}$ denotes the batch samples randomly sampled from the training dataset $D_{Tr}$, $|D_t|$ denotes the batch size, $\nabla_w f_i^{Tr}(w_t)$ denotes the gradient of sample $i$ computed at $w_t$ and $\alpha_{t}$ is the LR at $t$-th iteration.

\vspace{-0mm}
\subsection{Existing LR Schedule Strategies}\vspace{-0mm}
As \cite{bengio2012practical} demonstrated, the choice of LR plays a central role for effective DNN training with SGD. In this part, we will recall LR schedules proposed in the previous works.

The following presents the commonly used pre-defined LR schedules for current DNN training:
\begin{align}
\begin{split} \label{eqsgd}
&(\mathrm{Fixed})  \ \ \   \alpha_t  = \textcolor{blue}{\alpha_0}, \\
\begin{split}
&(\mathrm{MultiStep})  \ \ \   \alpha_t  = \textcolor{blue}{\alpha_0}  \times (\textcolor{blue}{\gamma_{M}})^{i}, l_{i-1}\leq E_{cur} \leq l_{i},\\
&\ \ \ \ \ \ \ \ \ \ \ \ \ \ \ \ \ \ \ \ \  \ \ \ \mathrm{for \ given \ epochs} \ \textcolor{blue}{l_0,l_1,\cdots,l_n},                                \end{split}\\
&(\mathrm{Exponential})\ \ \  \alpha_t  = \textcolor{blue}{\alpha_0} \times (\textcolor{blue}{\gamma_{E}})^{E_{cur}-1}, \\
&(\mathrm{SGDR}) \   \alpha_t  \!=\! \textcolor{blue}{\alpha_{\min}} \!+\!0.5(\!\textcolor{blue}{\alpha_{\max}}\!-\!\textcolor{blue}{\alpha_{\min}}\!)\!\left(\!1\!+\!cos(\!\frac{E_{cur}}{\textcolor{blue}{E_{per}}}\!\pi\!)\!\right),
\end{split}
\end{align}
where $\alpha_0$ denotes the initial LR and $\alpha_t$ denotes the LR at $t$-iteration, $[\alpha_{\min},\alpha_{\max}]$ specifies a range for LR setting of SGDR. $E_{cur}$ accounts for how many epochs have been performed, and $E_{per}$ denotes that after $E_{per}$ epochs SGDR restarts to decrease the LR, and it generally sets $E_{per}=E_{0}\times (T_{Mul})^k$ for the $k$-th restart. $\gamma_{M},\gamma_{E}<1$ denote the decay factors for MultiStep and Exponential, respectively.

Compared with pre-defined formulation of LR schedules, adaptive gradient methods like Adam \cite{kingma2014adam} can adaptively adjust LR by making use of (an approximation of) second order gradient information, involving the initial global LR required to be tuned. Besides, some methods extend classical line search methods in convex optimization to the training algorithm on DNNs, such as Polyak’s update rule \cite{rolinek2018l4}, Frank-Wolfe algorithm \cite{berrada2018deep}, and Armijo line-search \cite{vaswani2019painless}, etc.

Though these methods achieve competitive results on some learning tasks, they still possess certain drawbacks: (1) The pre-defined LR schedules suffer from the limited flexibility to adapt the highly variable training dynamics for the complicated deep learning optimization problems.
(2) It always needs to repetitively redesign proper LR schedules from scratch for new query tasks, as well as to tune their involved hyperparameters.
This process is time and computation expensive, and always requires expert prior knowledge to the problem, which tends to further raise their application difficulty in real problems.

Inspired by current meta-learning developments \cite{finn2017model,shu2018small,shu2019meta}, some researches proposed to learn a generic optimizer from data \cite{andrychowicz2016learning,ravi2016optimization,chen2017learning,wichrowska2017learned,li2017learning,lv2017learning}. The main idea among them is to learn a meta-learner as the optimizer to guide the learning of the whole updating rules. For example, \cite{andrychowicz2016learning} tries to replace Eq.(\ref{eq2}) with the following formulation,
\begin{align}\label{eq5}
	w_{t+1} = w_{t} + g_t,[g_t,h_{t+1}]^T=m(\nabla_t,h_t;\phi),
\end{align}
where $g_t$ is the output of a LSTM net $m$, parameterized by $\phi$, whose state is $h_t$.
This strategy has been expected to make selecting appropriate training algorithms, scheduling LR and tuning other hyper-parameters in a unified and automatic way.
Though faster in decreasing training loss than the traditional optimizers in some cases, the learned optimizer, however, might not always generalize well to more variant and diverse problems, like longer horizons \cite{lv2017learning} and large scale optimization problems \cite{wichrowska2017learned} since the framework is too flexible to be relatively easy to overfit training tasks.

Rather than the entire learning rules, a natural compromise for the task is to focus on the LR schedules while keep to use the gradient knowledge across the meta-training/testing stages. Inspired by this motivation, recently some methods \cite{franceschi2017forward,baydin2017online} consider the following constrained optimization problem to search the optimal LR schedule $\alpha^*$  such that the produced models are associated with small validation error,
\begin{align}\label{eq6}
\begin{split}
    &\min_{\alpha=\{\alpha_0,\cdots,\alpha_{T-1}\}} f^{Val}(D_{Val};w_T ),
    \\ \ &s.t. \ w_{t+1} = \xi_t(w_{t},\alpha_t), \ t=0,1,\cdots,T-1,
\end{split}
\end{align}
where $f^{Val}$ denotes the validation loss function, $D_{Val}=\{1,2,\cdots,M\}$ denotes hold-out validation set, $\alpha_t$ is to-be-solved LR hyper-parameter, $\xi_t:\mathbb{R}^d\times\mathbb{R}_+\rightarrow \mathbb{R}^d$ is a stochastic weight update dynamics, like the updating rule of the vanilla SGD in Eq.(\ref{eq3}), and $T$ is the maximum iteration step.
Though achieving comparable results on some tasks with hand-designed LR schedules and meta-learned optimizers, when generalized to new tasks, the meta-learned LR schedules keep constant. This makes it hardly well adapt to the task variations, and thus lead to possible performance degradation. Namely, it still requires to re-learn the LR schedules especially for new heterogeneous tasks, which is also time and computation expensive.

\begin{figure}[t] \vspace{-0mm}
	\begin{minipage}{0.49\textwidth}
		\subfigcapskip=-1mm
		\subfigure[\scriptsize{Computational graph of MLR-SNet}]{
			\label{figneta} 
			\includegraphics[width=0.65\textwidth]{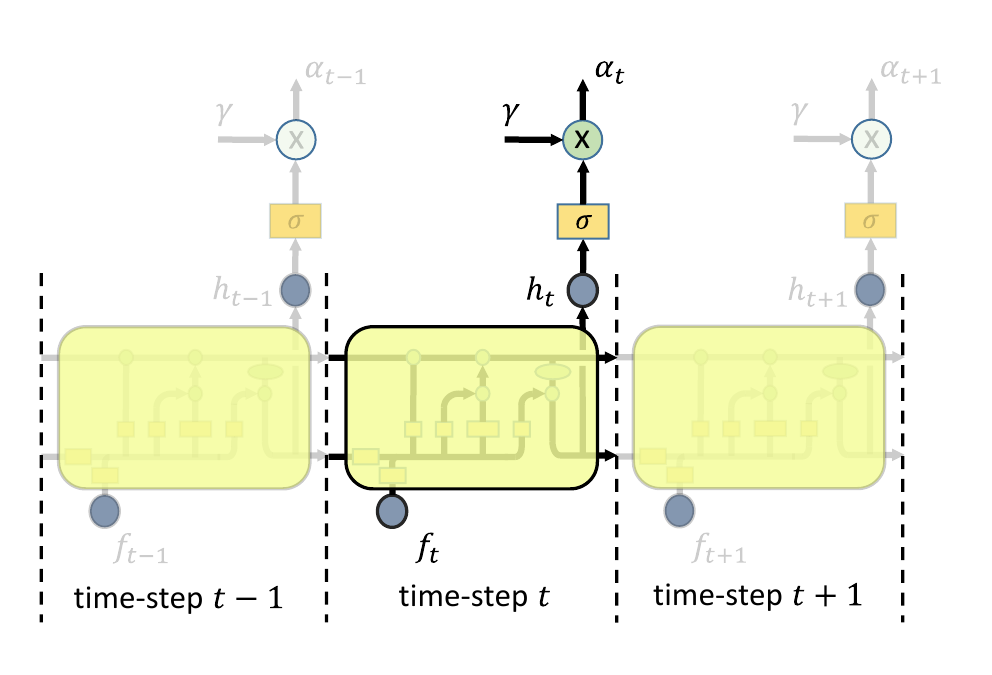}} \hspace{-3mm}
		\subfigure[\scriptsize{One step of MLR-SNet}]{
			\label{fignetb} 
			\includegraphics[width=0.33\textwidth]{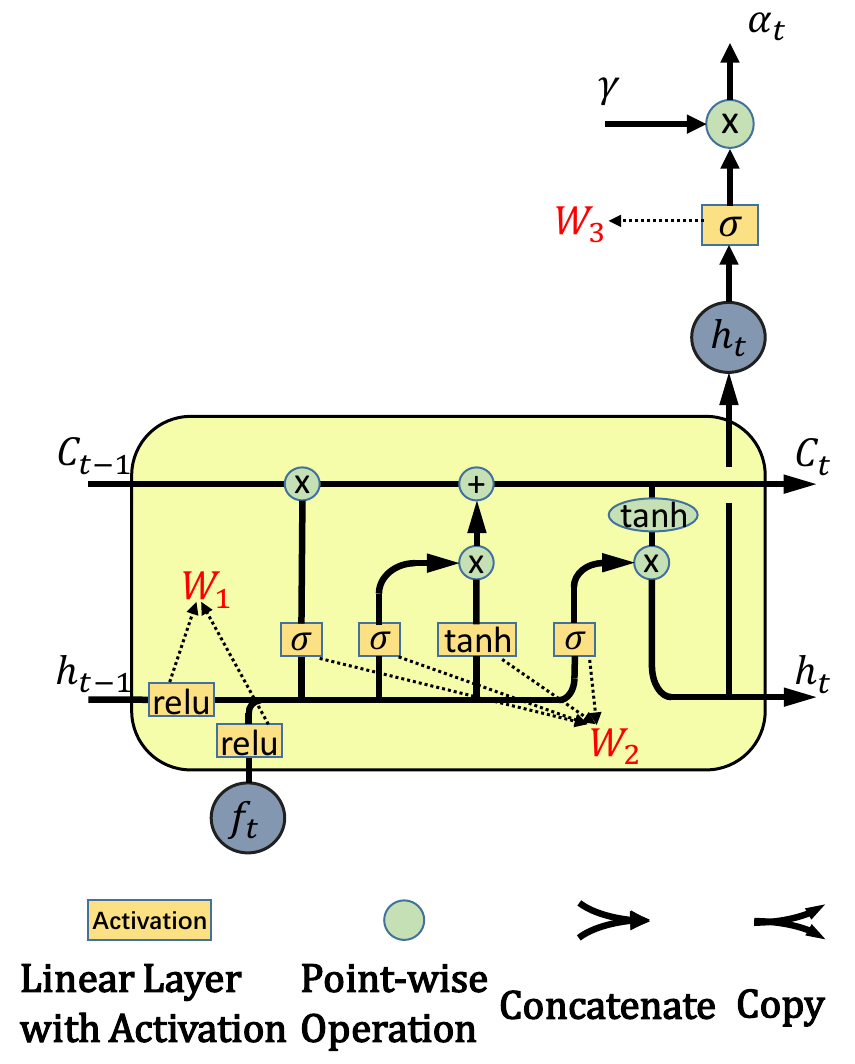}} \vspace{-3mm}
		\caption{The structure and computational graph of our proposed MLR-SNet.}\vspace{-1mm}
		\label{fignet}
	\end{minipage} \ \ \ \ \
\end{figure}

\subsection{Proposed Meta-LR-Schedule-Net Method}
To address aforementioned issues, we propose to design a meta-learner with an explicit mapping formulation to parameterize LR schedules as shown in Fig.\ref{fig1a}, called Meta-LR-Schedule-Net (\textbf{MLR-SNet} for brevity). The parameterized structure can bring two benefits: 1) It gives a fine flexibility to learn a proper LR schedule to comply with the significantly changed training dynamics of DNNs; 2) It makes the meta-learned LR schedules become transferable and plug-and-play, able to be readily applied to new heterogeneous tasks, without requiring to re-learn or tune additional hyperparameters.
\subsubsection{Formulation of MLR-SNet}
The computational graph of MLR-SNet is depicted in Fig.\ref{figneta}. Let $\!\mathcal{A}(\cdot,\cdot;\phi)\!$ denote MLR-SNet. Then the updating equation of the vanilla SGD algorithm in Eq.(\ref{eq3}) can be rewritten as:
\begin{align}\label{eq101}
\begin{split}
	 w_{t+1}& = w_{t} - \mathcal{A}(f_t,\theta_t;\phi)\nabla_{w} f^{Tr}(D_t; w_t),\\ \text{where}\  f_t&=f^{Tr}(D_t;w_t), \ \theta_t = (h_t,c_t)^T,
\end{split}
\end{align}
where $\mathcal{A}(f_t,\theta_t;\phi)$ outputs the LR ($\alpha_t$) at the $t$-th iteration, $\phi$ is the parameter of MLR-SNet, $f_t$ is the loss of the batch samples $D_t$ at the $t$-th iteration, and $\theta_t=\{h_{t-1},c_{t-1}\}$, where $h_t,c_t\in R^{d'}$ denote the output and state of the LSTM cell at the $t$-th iteration ($t=0,\cdots,T-1$), $d'$ represents the dimension of the state vectors (i.e., the size of hidden nodes). At each SGD iteration, $\mathcal{A}(f_t,\theta_t;\phi)$ can learn an explicit loss-LR dependent relationship, such that the net can adaptively predict LR according to the current input loss $f_t$, as well as the historical training information $\theta_t$ stored in the net. For every iteration step, the whole forward computation process can be written as (as shown in Fig. \ref{fignetb}):
\begin{equation}
\begin{array}{c}
\left(\!
\begin{array}{c}
I_t\\
F_t\\
O_t\\
g_t\\
\end{array}
\right) \! =\!
\left(
\begin{array}{c}
\sigma\\
\sigma\\
\sigma\\
\tanh\\
\end{array}
\!\right) W_2
\left(\begin{array}{c}
{\rm relu}\\
{\rm relu}\\
\end{array} \right)  W_1 \left(  \begin{array}{c}
h_{t-1}\\
f_t\\
\end{array}\right)\\
c_t = F_t \odot c_{t-1} + I_{t}\odot g_{t}\\
h_t = O_t \odot \tanh(c_t)\\
p_t = \sigma(W_3h_t)\\
\alpha_t = \gamma \cdot p_t \\
\end{array},
\end{equation}
where $I_t,F_t,O_t$ denote the Input, Forget and Output gates in the current iteration, and $\sigma, {\rm tanh}, {\rm relu}$ denote the Sigmoid, Tanh and ReLU activation functions, respectively. The MLR-SNet parameter is $\phi \!=\! (W_1,W_2,W_3)$, where $W_1 \in \mathbb{R}^{d' \times (d'+1)}, W_2 \in \mathbb{R}^{4d' \times 2d'}, W_3 \in \mathbb{R}^{1 \times d'}$.
Different from the vanilla LSTM, the input $h_{t-1}$ and the training loss $f_t$ are preprocessed by a fully-connected layer $W_1$ with ReLU activation function. Then it works as the LSTM and obtains the output $h_t$.
Subsequently, the predicted value $p_{t}$ is obtained by a linear transform $W_3$ on the $h_{t}$ with a Sigmoid activation function. Finally, we introduce a scale factor $\gamma$
to guarantee the final predicted LR located in the interval of $[0,\gamma]$. In our paper, we set $\gamma\!=\!\frac{f_0^{1/2}\log|f_0*C|}{4C^{1/4}}$, where $f_0$ denotes the initial loss, and $C$ accounts for the number of classes.
Albeit simple, this net is known to be capable of finely dealing with such long-term information dependent problem, and thus expected to learn a proper LR schedule to comply with the training dynamics of DNNs.

\textbf{Remark.} On the one hand, different from Eq.(\ref{eq6}) directly learning the LR schedules themselves, we use the MLR-SNet parameterized by $\phi$ to learn the LR schedules. This parameterized meta-learner helps extract the latent methodology of how to design a proper LR schedule for generally handling a DNN training problem, rather than only the hyper-parameters for a specific problem. Therefore, the meta-learned MLR-SNet can be readily transferred to new DNN training tasks for designing the LR schedules. On the other hand, compared with learning the whole updating rules as represented in Eq.(\ref{eq5}), our MLR-SNet learns the most important LR schedules for SGD algorithm while keep using the gradient knowledge of the learned problem, making it relatively easier to learn and under better control. This can explain why MLR-SNet always tends to make the DNN training procedure more robust and efficient in experiments.

\subsubsection{Learning Algorithm of MLR-SNet}
\textbf{(1) Meta-Train: adapting to the training dynamics of DNN.} The MLR-SNet can be meta-trained to improve the generalization performance on unseen validation data for DNN training by solving the following optimization problem:
\begin{align}\label{eq7}
\begin{split}
	&\min_{\theta} f^{Val}(D_{Val};w_T(\phi)),\\ \ s.t. \ w_{t+1}(\phi) &= \xi_t(w_{t},\mathcal{A}(f_t,\theta_t;\phi)), \ t=0,\cdots,T-1.
\end{split}
\end{align}
where $f_t = f^{Tr}(D_t;w_t)$ and $\xi_t(w_{t},\alpha_t)$ corresponds to Eq. (\ref{eq3}). Now the important question is how to efficiently meta-learn the parameter $\phi$ for the MLR-SNet. We employ the online approximation technique in \cite{shu2019meta} to jointly update $\phi$ and model parameter $w$ to explore a proper LR schedule with better generalization for DNNs training. However, the step-wise optimization for $\phi$ is still expensive to handle large-scale datasets and huge DNN structures. To address this issue, we attempt to update $\phi$ once after updating $w$ several steps ($T_{val}$). The updating process can then be formulated as:
\begin{algorithm}[t]
	\vspace{0mm}
	\renewcommand{\algorithmicrequire}{\textbf{Input:}}
	\renewcommand{\algorithmicensure}{\textbf{Output:}}
	\caption{The Meta-Train Algorithm of MLR-SNet}
	\label{alg}
	\begin{algorithmic}[1]  \small
		\REQUIRE  Training data $D_{Tr}$, validation set $D_{Val}$, max iterations $T$, updating period $T_{val}$.
		\ENSURE  Model parameter $w_T$ and MLR-SNet parameter $\phi_s,s\in S\subset \{1,\cdots, T\}$
		\STATE Initialize model parameter $w_0$, MLR-SNet cell $\theta_0 = (h_0,c_0)^T$, and MLR-SNet parameter $\phi_0$.
		\FOR{$t=0$ {\bfseries to} $T-1$}
		\STATE $D_t \leftarrow$ SampleMiniBatch($D_{Tr}$) with batch size $|D_t|$.
		\IF{$t\ \% \ T_{val}=0$, }
		\STATE $D_t^{(v)} \!\leftarrow\!$ SampleMiniBatch($D_{Val}$) with batch size $|D_t^{(v)}|$.
		\STATE Update $\phi_{t+1}$ by Eq. (\ref{eq9}).
		\ENDIF
		\STATE Update $w_{t+1}$ by Eq. (\ref{eq10}).
		\ENDFOR
	\end{algorithmic}
\end{algorithm}

\textbf{Updating $\phi$}. When it does not satisfy the updating conditions, $\phi$ keeps fixed; otherwise, $\phi$
will be updated using the model parameter $w_{t}$ and MLR-SNet parameter $\phi_{t}$ obtained in the last step by minimizing the validation loss defined in Eq.(\ref{eq7}). Adam algorithm can be utilized to optimize the validation loss, expressed as:
\begin{align}\label{eq9}
	\phi_{t+1} = \phi_{t} + Adam(\nabla_{\theta} f^{Val}(D_t^{(v)};\hat{w}_{t+1}(\theta));\eta_t),
\end{align}
where $Adam$ denotes the Adam algorithm, whose input is the gradient of validation loss with respect to MLR-SNet parameter $\phi$ on mini-batch samples $D_t^{(v)}$ from $D_{Val}$. $\eta_t$ denotes the LR of Adam. 
$\hat{w}_{t+1}(\phi)$\footnote{Notice that $\hat{w}_{t+1}(\phi)$ here is a function of $\phi$ to guarantee the gradient in Eq.(\ref{eq9}) to be able to be feasibly computed.} is virtually formulated on a mini-batch training samples $D_t$ from $D_{Tr}$ as follows:
\begin{small}
\begin{align}
\hat{w}_{t+1}(\phi) \!=\! w_t - \mathcal{A}(f^{Tr}(D_t,w_t),\theta_t;\phi)\!\cdot\! \nabla_{w} f^{Tr}(D_t,w)\big|_{w_t}.
\end{align}
\end{small}\vspace{-4mm}

\textbf{Updating $w$}. Then, the updated $\phi_{t+1}$ is employed to ameliorate the model parameter $w$, i.e.,
\begin{small}
\begin{align}\label{eq10}
{w}_{t+1} \!=\! w_t - \mathcal{A}(f^{Tr}(D_t,w_t),\theta_t;\phi_{t+1})\!\cdot\! \nabla_{w} f^{Tr}(D_t,w)\big|_{w_t}.
\end{align}
\end{small}
The whole algorithm in the meta-training stage can then be summarized in Algorithm \ref{alg}. All computations of gradients can be efficiently implemented by automatic differentiation libraries, like PyTorch \cite{paszke2019pytorch}, and easily used to general DNN architectures. It can be seen that the MLR-SNet can be gradually optimized during the learning process and adjust the LR dynamically based on the training dynamics of DNNs.\\
\begin{algorithm}[t]
	\vspace{0mm}
	\renewcommand{\algorithmicrequire}{\textbf{Input:}}
	\renewcommand{\algorithmicensure}{\textbf{Output:}}
	\caption{The Meta-Test Algorithm of MLR-SNet}
	\label{alg2}
	\begin{algorithmic}[1]  \small
		\REQUIRE  Training data $D_{Tr}^{\mu}$ for new task $\mu$, max iterations $T_{\mu}$, meta-learned MLR-SNet $\mathcal{A}(\cdot,\cdot;\phi_s), s\in S$.
		\ENSURE  Model parameter $u_T$.
		\STATE Initialize model parameter $u_0$, MLR-SNet cell $\theta_0 = (h_0,c_0)^T$, and choose the subset of meta-learned MLR-SNet $\phi_s, s \in S \subset\{1,\cdots,T\}$ for test.
		\FOR{$t=0$ {\bfseries to} $T_{\mu}-1$}
		\STATE $D_t^{\mu} \leftarrow$ SampleMiniBatch($D_{Tr}^{\mu}$) with batch size $|D_t^{\mu}|$.
		\STATE Compute the loss $\!f^{Tr}(D_t^{\mu},u_t)$, and then MLR-SNet predicts the LR $\!\mathcal{A}(f^{Tr\!}(\!D_t^{\mu},u_t),\theta_t;\phi_s)\!$ for current iteration.
		\STATE Update $u_{t+1}$ by Eq. (\ref{eq11}).
		\ENDFOR
	\end{algorithmic}
\end{algorithm}
\textbf{(2) Meta-Test: generalization to new heterogeneous tasks.} After the meta-training stage, the meta-learned MLR-SNet with parameter $\phi_T$ is expected to be transferred to guild the SGD running on new DNN training tasks. To better preserve the proper LR changing dynamics during DNN training, we more prefer to keep several MLR-SNet forms with parameters $\phi_s, s \in S \subset\{1,\cdots,T\}$ (e.g., {$\phi_{T/3}, \phi_{2T/3}, \phi_T$} as employed in our experiments) and use them as LR schedules along different iterations in the meta-testing stage. The new DNN parameter $u$ for the new task is then updated by (the whole meta-test process refers to Algorithm \ref{alg2}),
\begin{align}\label{eq11}
	{u}_{t+1} \!=\! u_t - \mathcal{A}(f^{Tr}(D_n,u_t),\theta_t;\phi_s)\!\cdot\! \nabla_{u} f^{Tr}(D_n,u)\big|_{u_t},
\end{align}
where $\phi_s, s\in S$ is the parameters of the subset of the meta-learned MLR-SNets. This means that we restore several LR schedule setting rules, and dynamically employ specific ones along different range of DNN training iterations. It is seen that the meta-learned MLR-SNets so learned are plug-and-play, and involve no additional hyperparameters to tune.

\begin{figure*}[t] \vspace{-0mm}
	\centering
	\subfigcapskip=-0mm
	\subfigure[CIFAR-10 with ResNet18 SGD]{
		\label{fig2a} 
		\includegraphics[width=0.30\textwidth]{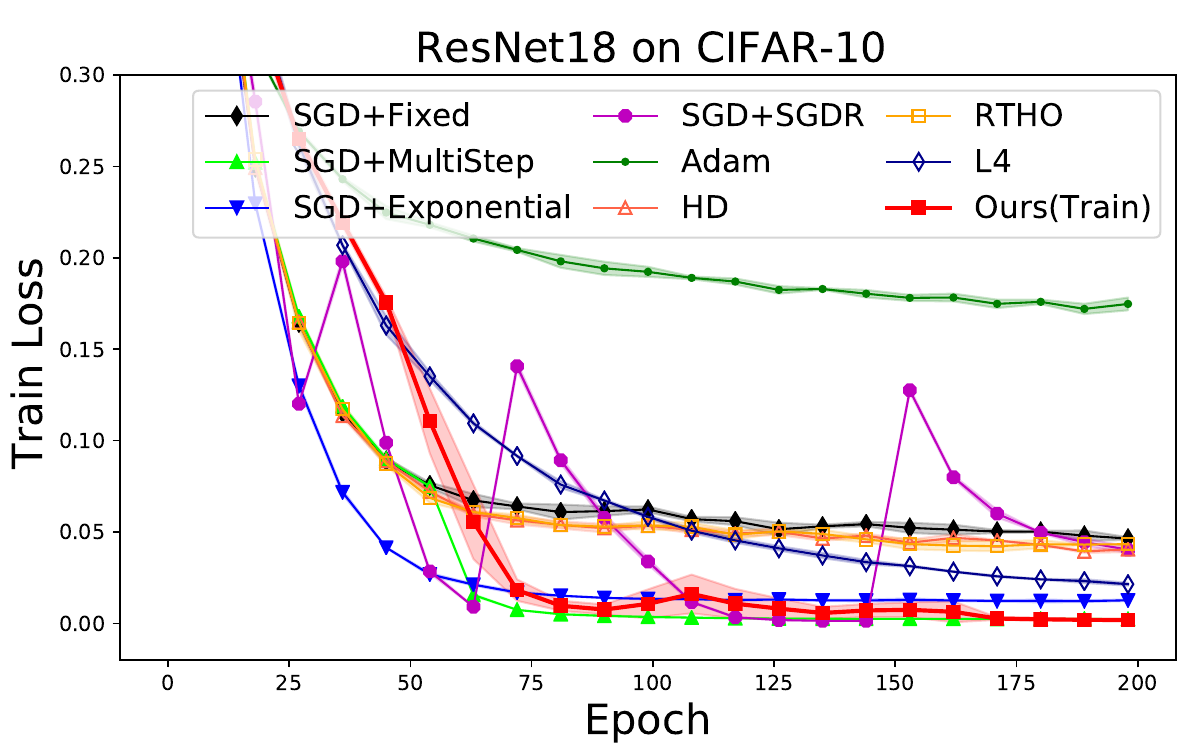}
	\includegraphics[width=0.30\textwidth]{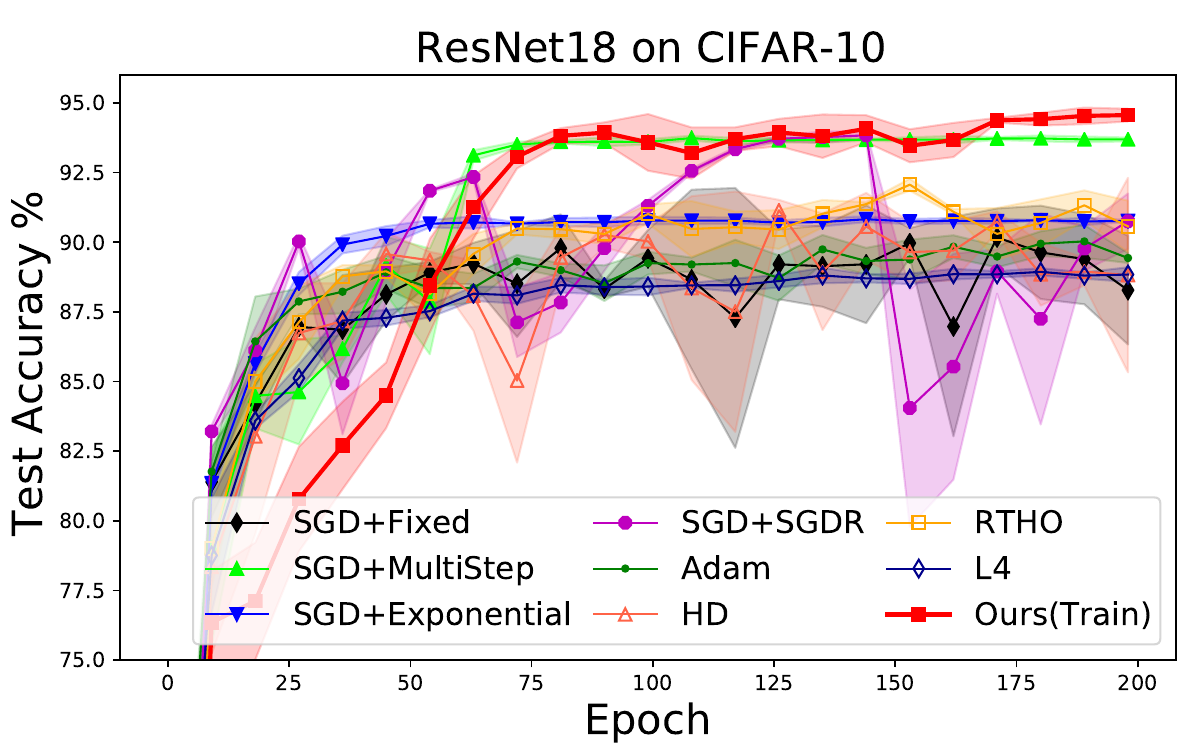}
\includegraphics[width=0.30\textwidth]{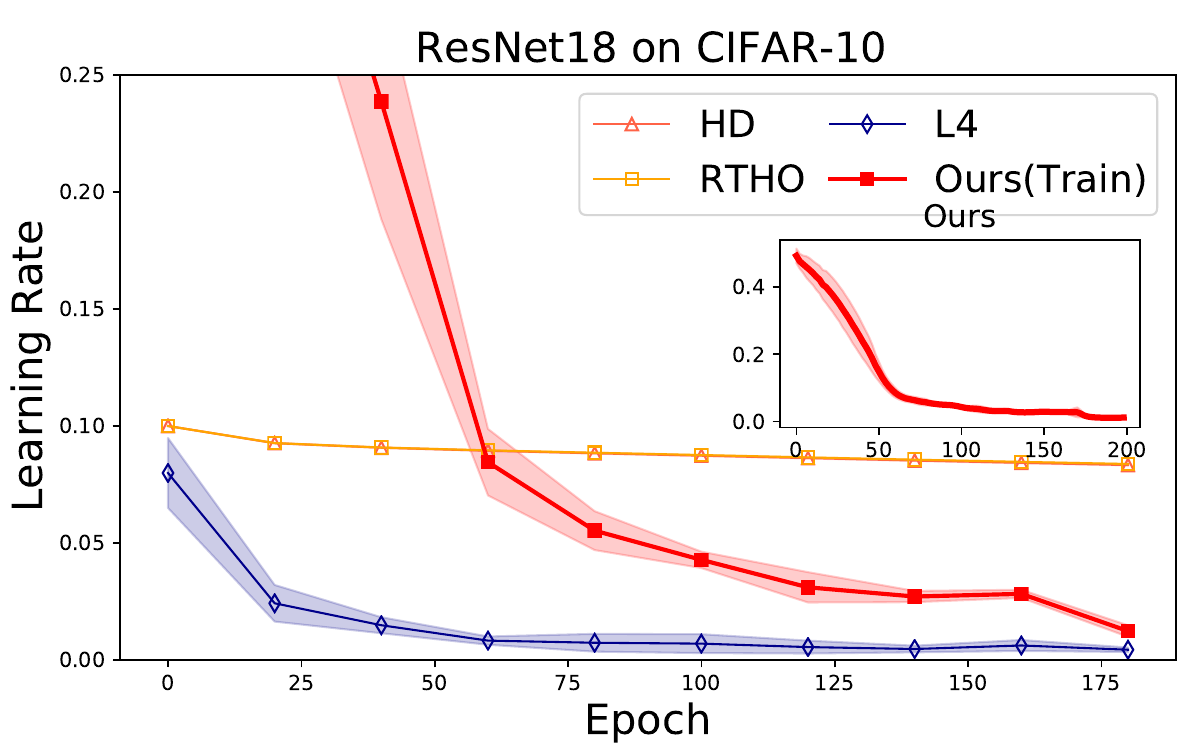}} \\ \vspace{-3mm}
		\subfigure[CIFAR-100 with WideResNet-28-10 for SGD]{
		\label{fig2c} 
		\includegraphics[width=0.30\textwidth]{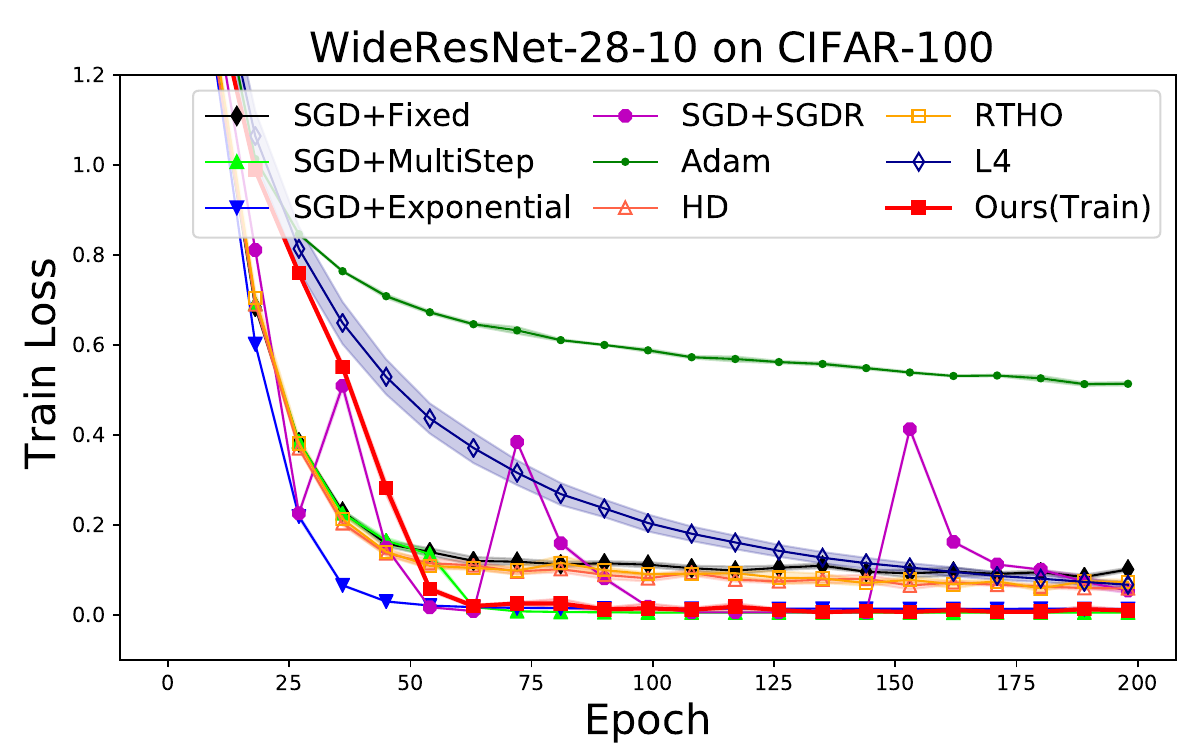}
		\label{fig2d} 
		\includegraphics[width=0.30\textwidth]{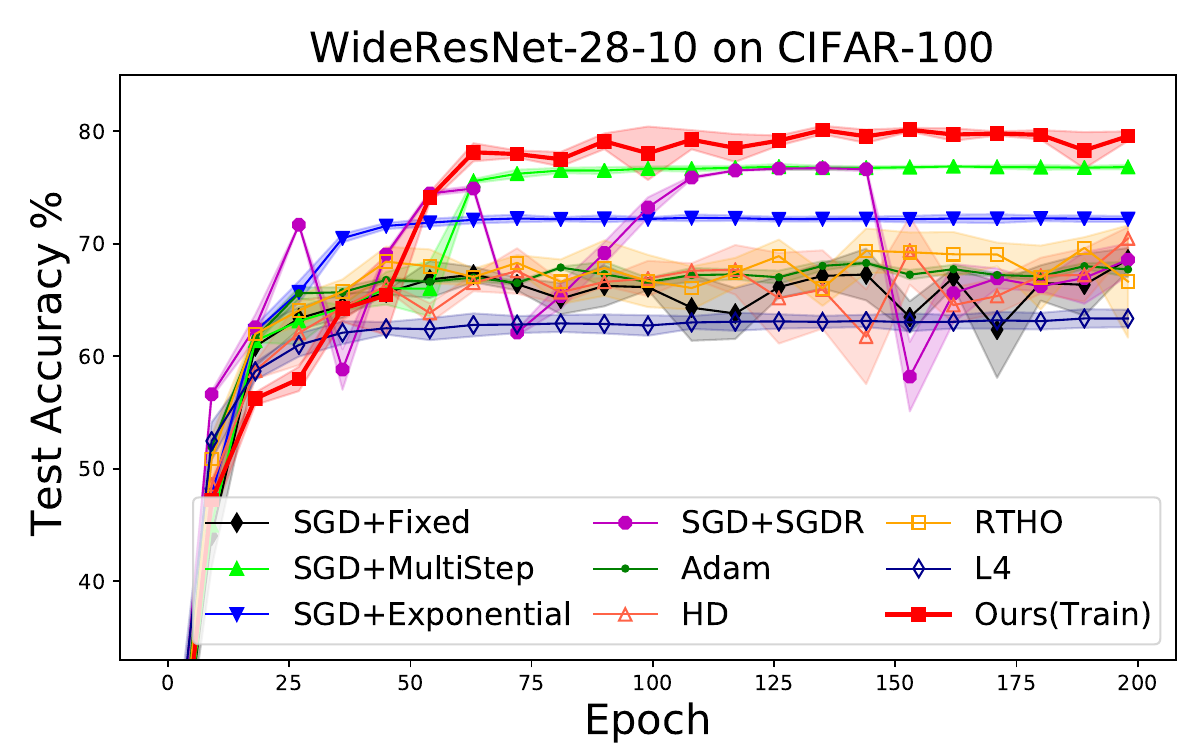}
	\includegraphics[width=0.30\textwidth]{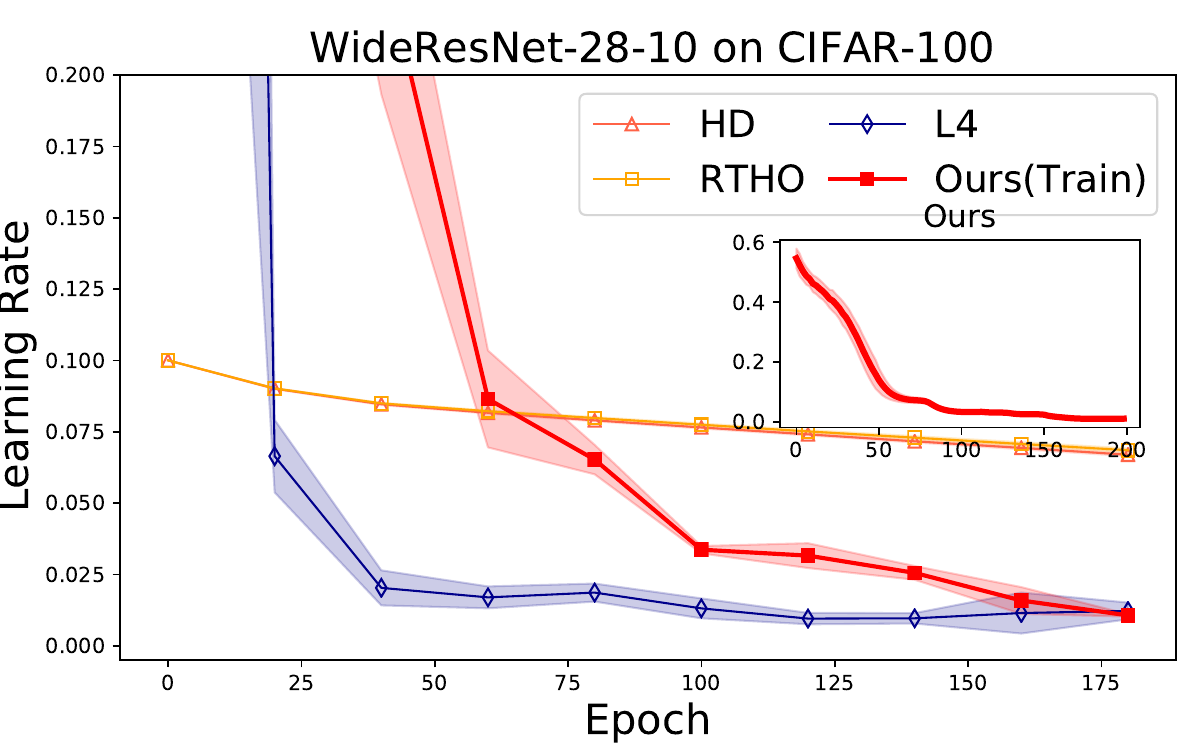}} \vspace{-3mm}
	\caption{Changing tendencies in terms of training perplexity (left column) and test accuracy (middle column) in iterations of all comparison methods on image classification datasets in the meta-train stage. The LR schedules (right column) employed by all methods are also compared. }\label{fig2}  \vspace{-2mm}
\end{figure*}

\section{Experimental Results}\label{result}
To evaluate the proposed MLR-SNet, we firstly conduct experiments to show our method can learn proper LR schedules compared with baseline methods (Section \ref{train}).
Then we transfer the meta-learned LR schedules to various tasks for meta-test to show its superiority in generalization (Section \ref{Transferability}). What influences the generalization perfromance of meta-learned LR schedules is discussed in Section \ref{how}. Finally, we show our method behaves robust and stable when training data contain different data corruptions (Section \ref{robust}).

\subsection{Meta-Train: Evaluation of the LR Schedules Meta-learned by MLR-SNet}\label{train}
In this section, we attempt to evaluate the capability of MLR-SNet to learn proper LR schedules for various tasks.

\subsubsection{Image Classification Benchmarks}\label{image}

\textbf{Datasets.} We choose CIFAR-10 and CIFAR-100 to present the efficiency of our method, which include 32$\times$32 color images arranged in 10 and 100 classes, respectively. Both datasets contain 50,000 training and 10,000 test images.

\textbf{Baselines.} The compared methods include the SGD with hand-designed LR schedules (the formulation is expressed as Eq. (\ref{eqsgd})): 1) \textbf{Fixed} LR, 2) \textbf{Exponential} decay, 3) \textbf{MultiStep} decay, 4) SGD with restarts (\textbf{SGDR})~\cite{loshchilov2016sgdr}.
Meanwhile, we compare with adaptive gradient method: 5)\textbf{Adam}, LR search method: 6) \textbf{L4} \cite{rolinek2018l4}, and current LR schedule adaptation method: 7) hyper-gradient descent (\textbf{HD}) \cite{baydin2017online}, 8) real-time hyper-parameter optimization (\textbf{RTHO}) \cite{franceschi2017forward}.
We run all experiments with 3 different seeds reporting accuracy. Our algorithm and RTHO \cite{franceschi2017forward} randomly select 1,000 clean images in the training set of CIFAR-10/100 as validation data.

\textbf{Hyperparameter setting.} We employ ResNet-18 on CIFAR-10 and WideResNet-28-10 \cite{zagoruyko2016wide} on CIFAR-100. All compared methods and MLR-SNet are trained for 200 epochs with batch size $128$. For baselines involving SGD as base optimizer, we set the initial LR as $0.1$, and weight decay as $5e^{-4}$. While for \textbf{Adam}, we just follow the default parameter setting. As for each LR schedule, \textbf{MultiStep} decays LR by $10$ every 60 epochs (i.e., $\gamma_M=0.1,l_0=0,l_1=60,l_2=120,l_3=180,l_4=200$); \textbf{Exponential} multiplys LR with $\gamma_E = 0.95$ every epoch; \textbf{SGDR} sets $\alpha_{\min}=1e^{-5},\alpha_{\max}=0.1$, and $E_0=10, T_{Mult}= 2$.
\textbf{L4}, \textbf{HD} and \textbf{RTHO} update LR every data batch, and we use the recommended setting in the original paper. \textbf{HD} and \textbf{RTHO} search different hyper-lrs from $\{1e^{-3}, 1e^{-4}, 1e^{-5}, 1e^{-6}, 1e^{-7}\}$ reporting the best performing hyper-lr.

\begin{table}[t]  \vspace{-0mm}
	\caption{Test accuracy (\%) of CIFAR datasets with SGD baselines.}\label{table11} \vspace{-2mm}
	\centering
	\resizebox{0.42\textwidth}{!}{
		\begin{tabular}{l|c|c}
			\toprule
			Optimizer 	&CIFAR-10  &CIFAR-100   \\
			\hline \hline
			SGD+Fixed	& 92.26 $\pm$ 0.12  & 70.67 $\pm$ 0.34   \\
			\hline
			SGD+MultiStep	& 93.82 $\pm$ 0.09 & 77.04 $\pm$ 0.17   \\
			\hline
			SGD+Exponential	& 90.93 $\pm$ 0.11 &  76.88 $\pm$ 0.08 \\
			\hline
			SGD+SGDR	& 93.92 $\pm$ 0.11 &  72.52 $\pm$ 0.34  \\
			\hline
			Adam	& 90.86 $\pm$ 0.15 & 68.94 $\pm$ 0.24  \\
			\hline \hline
			SGD+L4	& 89.15 $\pm$ 0.14 &  63.61 $\pm$ 0.65 \\
			\hline
			SGD+HD	& 92.34 $\pm$ 0.09 &  72.22 $\pm$ 0.30  \\
			\hline
			SGD+RTHO		& 92.60 $\pm$ 0.18 &  72.32 $\pm$ 0.47  \\
			\hline  \hline
			MLR-SNet (Meta-train)	& \textbf{94.80 $\pm$ 0.10}	  & \textbf{80.44 $\pm$ 0.17}   \\
			\bottomrule
	\end{tabular}} \vspace{-3mm}
\end{table}\vspace{0mm}

\textbf{MLR-SNet architecture.}
The architecture of MLR-SNet is illustrated in Section 3.2.
In our experiment, the size of hidden nodes (i.e., $d'$) is set as 50. The initialization of MLR-SNet follows the default setting in Pytorch. We employ Adam optimizer to train MLR-SNet, and set the LR as $1e^{-3}$, and the weight decay as $1e^{-4}$. The input of MLR-SNet is the training loss of a mini batch samples. Every iteration LR is predicted by MLR-SNet and we update it every 100 iterations ($T_{val}=100$) according to the loss on the validation data.

\begin{figure*}[t] \vspace{-0mm}
	\centering
	\subfigcapskip=0mm
	\subfigure[Penn Treebank with 2-layer LSTM for SGD]{
		\label{fig2e} 
		\includegraphics[width=0.30\textwidth]{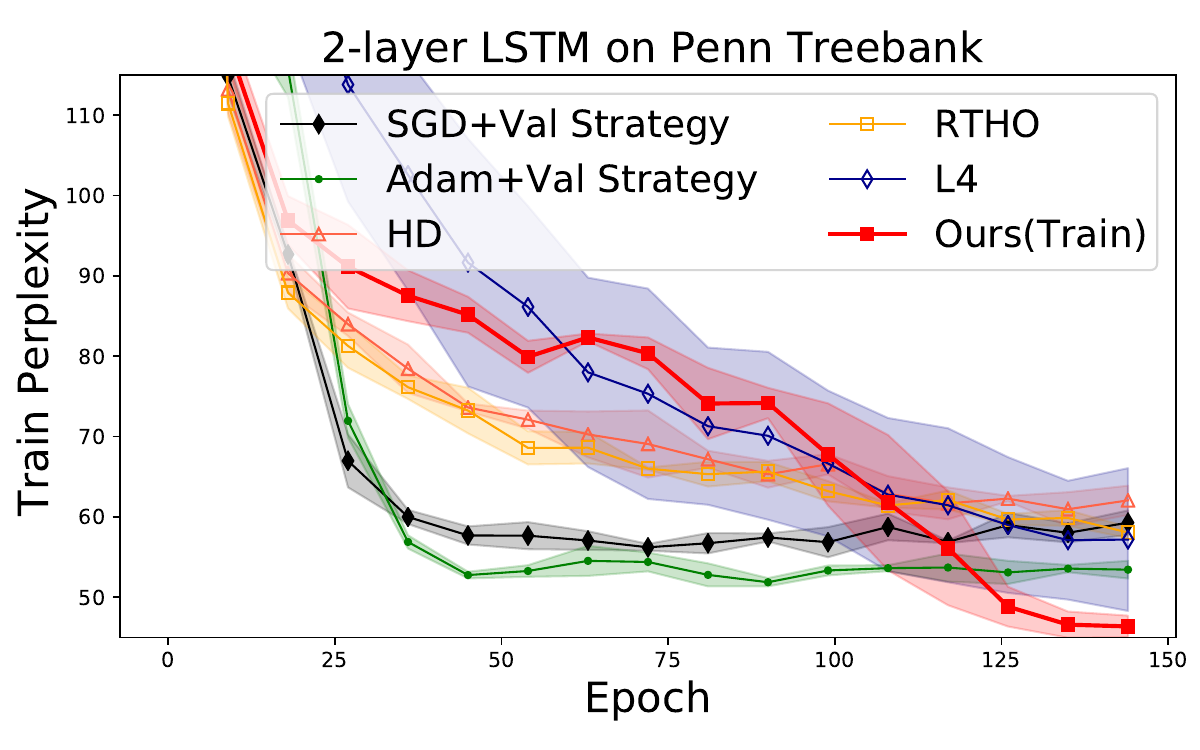}
		\label{fig2f} 
		\includegraphics[width=0.30\textwidth]{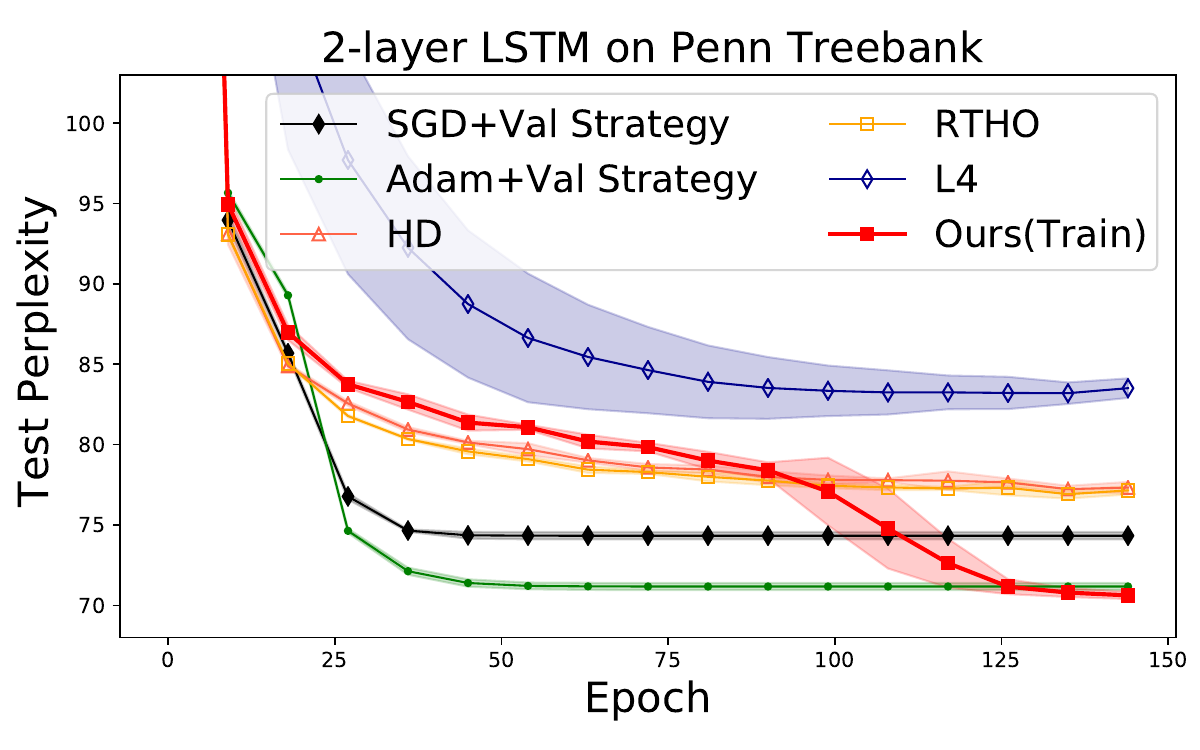}
		\includegraphics[width=0.30\textwidth]{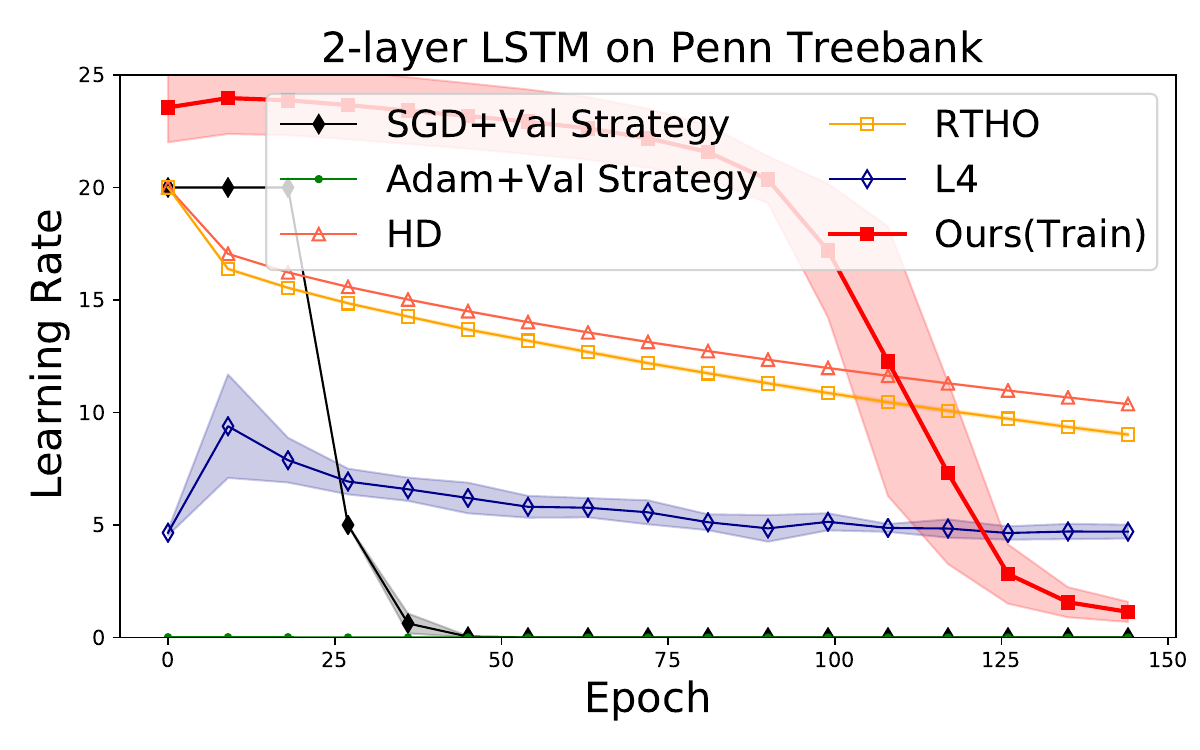}} \\ \vspace{-3mm}
	\subfigure[Penn Treebank with 3-layer LSTM for SGD]{
		\label{fig2g} 
		\includegraphics[width=0.30\textwidth]{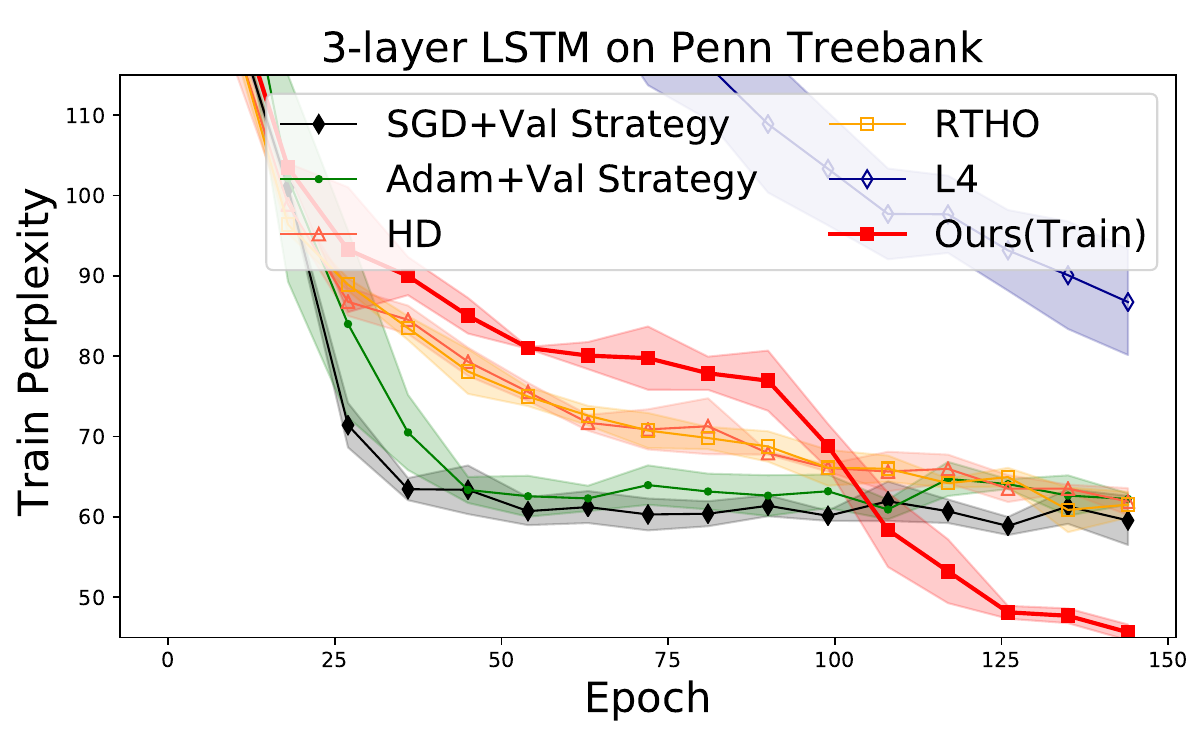}
		\label{fig2h} 
		\includegraphics[width=0.30\textwidth]{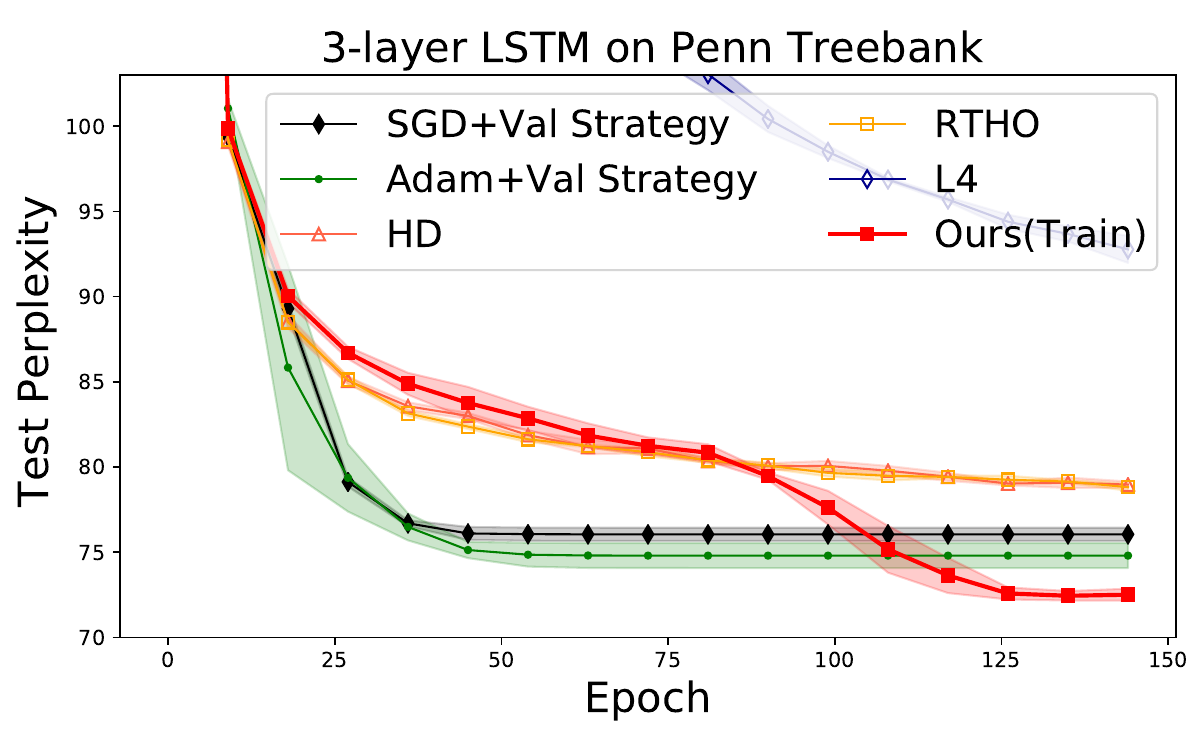}
		\includegraphics[width=0.30\textwidth]{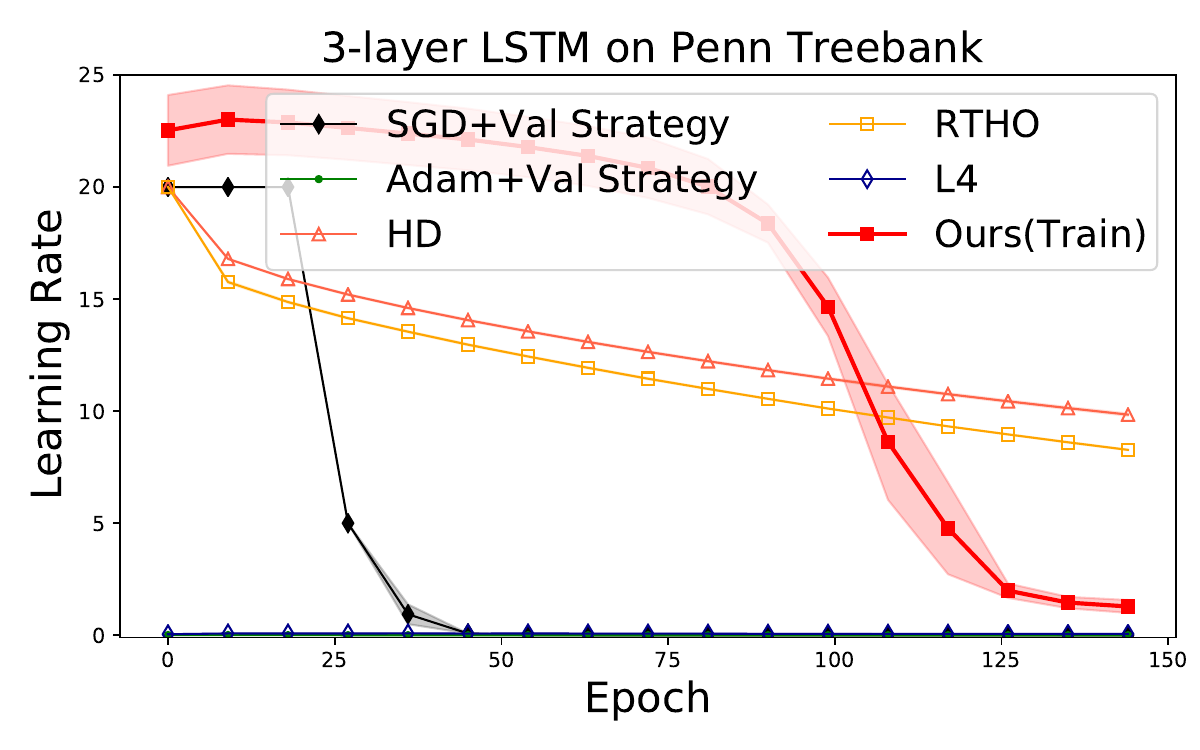}} \vspace{-3mm}
	\caption{Changing tendencies in terms of training perplexity (left column) and test perplexity (middle column) in iterations of all comparison methods on text classification datasets in the meta-train stage. The LR schedules (right column) employed by all methods are also compared.}\label{fig22}  \vspace{-2mm}
\end{figure*}

\begin{table}[t]  \vspace{-0mm}
	\caption{Test accuracy (\%) of CIFAR dataset with SGDM baselines.}\label{table21} \vspace{-2mm}
	\centering
	\resizebox{0.42\textwidth}{!}{
		\begin{tabular}{l|c|c}
			\toprule
			Optimizer 	&CIFAR-10 &CIFAR-100  \\
			\hline \hline
			SGDM+Fixed	& 87.69 $\pm$ 0.14  & 70.88 $\pm$ 0.12   \\
			\hline
			SGDM+MultiStep	& \textbf{95.08 $\pm$ 0.13} & 80.74 $\pm$ 0.19   \\
			\hline
			SGDM+Exponential	& 94.64 $\pm$ 0.05 &  78.87 $\pm$ 0.04 \\
			\hline
			SGDM+SGDR	& 95.06 $\pm$ 0.17 &  \textbf{80.93 $\pm$ 0.05}  \\
			\hline
			Adam	& 90.86 $\pm$ 0.15 & 68.94 $\pm$ 0.24  \\
			\hline \hline
			SGDM+L4	&91.03 $\pm$ 0.14 &  66.51 $\pm$ 2.83 \\
			\hline
			SGDM+HD	& 93.99 $\pm$ 0.12 &  76.80 $\pm$ 0.19  \\
			\hline
			SGDM+RTHO		& 93.17 $\pm$ 0.49 & 76.14 $\pm$ 0.29  \\
			\hline  \hline
			MLR-SNet (Meta-train)	& 94.80 $\pm$ 0.10	  & 80.44 $\pm$ 0.17   \\
			\bottomrule
	\end{tabular}}\vspace{-3mm}
\end{table}

\textbf{Results.} Fig.\ref{fig2a} and \ref{fig2c} show changing tendencies of training loss and test accuracy on CIFAR-10 and CIFAR-100 datasets in iterations of all competing methods, respectively, and Table \ref{table11} shows the corresponding classification accuracy on the test set.
It can be observed that: 1) our MLR-SNet obtains better test performance than all other competing methods, and the learned LR schedules by MLR-SNet have similar shapes as the corresponding hand-designed policies (as depicted in Fig. \ref{fig1c}), while with more elaborate variation details in locality for better adapting training dynamics. 2) The Fixed LR decreases the loss sharing the similar performance to other baselines at the early training, while fails to further decrease loss at the later training stages. This implies that this strategy could not finely adapt to such DNN training dynamics. 3) The MultiStep LR drops the LR at some epochs, and such elegant strategy overcomes the issue of Fixed LR and decreases loss substantially after dropping the LR. Thus it
obtains higher test performance. Besides, though MultiStep and MLR-SNet can decrease the loss to 0 approximately, our MLR-SNet achieves better generalization performance since the outer objective in Eq. (\ref{eq7}) tends to help learn the LR schedules to find a better minima. 4) The Exponential LR decreases loss with a faster speed at the early training steps than other baselines, while makes a slow progress due to smaller LR at the later stages. 5) The SGDR LR uses the cyclic LR, decreasing loss as fast as the Exponential LR. 6) Though Adam has an adaptive coordinate-specific LR, it behaves worse than MultiStep and Exponential LR as demonstrated in \cite{wilson2017marginal}. An extra tuning is thus necessary for better performance. 7) L4 greedily searches LR locally to decrease loss, making it fairly hard to adapt the complex DNNs training dynamics, and even with worse test performance than Fixed LR. 8) HD and RTHO perform similar as hand-designed LR schedules. Moreover, with an explicit parameterized structure, our MLR-SNet can make the learning of LR schedules more robust, and produce better test performance than HD and RTHO. 9) Since the image tasks often use SGD algorithm with Momentum (SGDM) to train DNNs, we also present the test performance of baseline methods trained with SGDM with momentum 0.9 in Table \ref{table21}.
They obtain a remarkable improvement than trained with SGD. Though not using extra historical gradient information to help optimization, our MLR-SNet is capable of achieving comparable results with baselines, since it also insightfully stores the historical LR training information in the net.

\begin{table}[t]\vspace{-0mm}
	\caption{  Test perplexity on the Penn Treebank dataset.}\label{table3}\vspace{-0mm}
	\centering
	\begin{tabular}{l|c|c}
		\toprule
		Optimizer 	&2-layer LSTM  &3-layer LSTM  \\
		\hline \hline
		SGD+Val Strategy	& 74.33 $\pm$ 0.23  & 76.05 $\pm$ 0.39   \\
		\hline
		Adam+Val Strategy	& 71.17 $\pm$ 0.23 & 74.80 $\pm$ 0.73  \\
		\hline \hline
		SGD+L4	& 82.58 $\pm$ 1.32 &  92.27 $\pm$ 0.92 \\
		\hline
		SGD+HD	& 76.90 $\pm$ 0.33 &  78.63 $\pm$ 0.08  \\
		\hline
		SGD+RTHO		& 76.69 $\pm$ 0.11 &  78.52 $\pm$ 0.16  \\
		\hline  \hline
		MLR-SNet (Meta-train)	& \textbf{70.53 $\pm$ 0.25}	  & \textbf{72.28$\pm$0.25}   \\
		\bottomrule
	\end{tabular} \vspace{-3mm}
\end{table}\vspace{0mm}

\subsubsection{Text Classification Benchmarks}
\textbf{Dataset.} We choose Penn Treebank dataset \cite{marcus19building} for evaluation, which consists of 929k training words, 73k validation words, and 82k test words, with a 10k vocabulary in total.

\textbf{Baselines.} We compare with 1) SGD, 2) Adam with LR tuned using a validation set (\textbf{SGD+Val Strategy} and \textbf{Adam+Val Strategy}). They drop the LR by a factor of 4 when the validation loss stops decreasing.
Also, we compared with 3) \textbf{L4}, 4) \textbf{HD}, 5) \textbf{RTHO}. We run all experiments with 3 different seeds reporting accuracy. Our algorithm and RTHO \cite{franceschi2017forward} regard the validation set as validation data.

\textbf{Hyperparameter setting.}
We use a 2-layer and 3-layer LSTM network which follows a word-embedding layer and the output is fed into a linear layer to compute the probability of each word in the vocabulary. Hidden size of LSTM cell is set to $512$ and so is the word-embedding size. We tie weights of the word-embedding layer and the final linear layer. Dropout is applied to the output of word-embedding layer together with both the first and second LSTM layers with a rate of $0.5$. As for training, the LSTM net is trained for 150 epochs with a batch size of $32$ and a sequence length of $35$. We set the base optimizer SGD to have an initial LR of $20$. For Adam, the initial LR is set to $0.01$ and weight for moving average of gradient is set to $0$. We apply a weight decay of $5e^{-6}$ to both base optimizers. All experiments involve a $0.25$ clipping to the network gradient norm. For both SGD and Adam, we decrease LR by a factor of 4 when performance on validation set shows no progress.  For L4, we try different $\alpha$ in $\{0.1, 0.05, 0.01, 0.005\}$ and report the best test perplexity among them. For both \textbf{HD} and \textbf{RTHO}, we search the hyper-lr lying in $\{1, 0.5, 0.1, 0.05\}$, and report the best results.

\textbf{MLR-SNet architecture.} We keep the same setting as the image classification, while we take $\frac{\mathcal{L}_{Tr}}{{\rm log} (vocabulary\ size)}$ as input of MLR-SNet to deal with the influence of large scale classes for text dataset.

\textbf{Results.} Fig.\ref{fig2e} and \ref{fig2g} show the train and test perplexity on the Penn Treebank dataset with 2-layer and 3-layer LSTM, respectively. It can be observed that: 1) The Val Strategy heuristically drops LR when the validation loss stops decreasing. This hand-designed LR schedules can decrease the loss quickly at the early training stage to find a good minima, while it is hard to further search a better solution.
2) Our MLR-SNet predicts LR according to training dynamics and updates its parameters by minimizing the validation loss, i.e., if the LR schedules produced by the MLR-SNet are of high quality, then a DNN model trained with such LR schedules should achieve low loss on a separate validation dataset. This process is a relatively more intelligent way to employ the validation dataset than Val Strategy. Thus our method achieves comparable or even better performance than Adam and SGD.
The meta-learned LR schedules of the MLR-SNet are shown in Fig.\ref{fig1b}, depicted as similar shapes as the hand-designed policies.
3) L4 often falls into a bad minima since it greedily searches LR locally.
4) Since HD and RTHO lack of an explicit parameterized structure, they directly learn LR schedules themselves by minimizing the validation loss, which tends possible to bring the optimization unstable, and lead to performance degradation. 5) When the number of LSTM's layers increases, the LR schedules predicted by MLR-SNet show more advantages for such an LSTM training problem, and bring more performance improvements compared with hand-designed LR schedules.

\textbf{Remark.} Actually, the performance of compared baselines can be approximately regarded as the best/upper performance bound. Since these strategies have been tested to work well for the specific tasks, and they are written into the standard deep learning library. For different image and text tasks, our MLR-SNet can achieve the similar or even slightly better performance compared with the best baselines. We thus believe that these experiments can demonstrate the effectiveness and generality of our proposed method.

\begin{figure}[t]\vspace{-0mm}
	\centering
	\subfigcapskip=-0mm
	\subfigure[Architectures of MLR-SNet]{
		\label{figaa} 
		\includegraphics[width=0.24\textwidth]{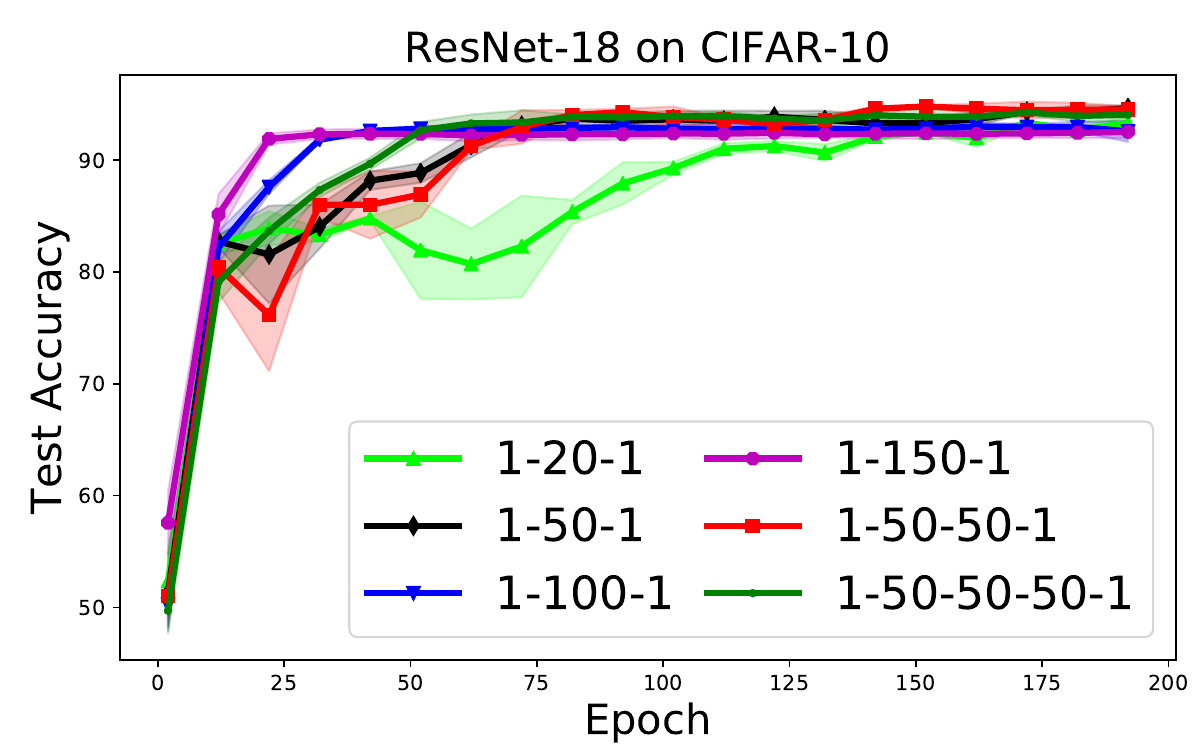}} \hspace{-3mm}
	\subfigure[LRs of Meta Optimizer]{
		\label{figbb}  
		\includegraphics[width=0.24\textwidth]{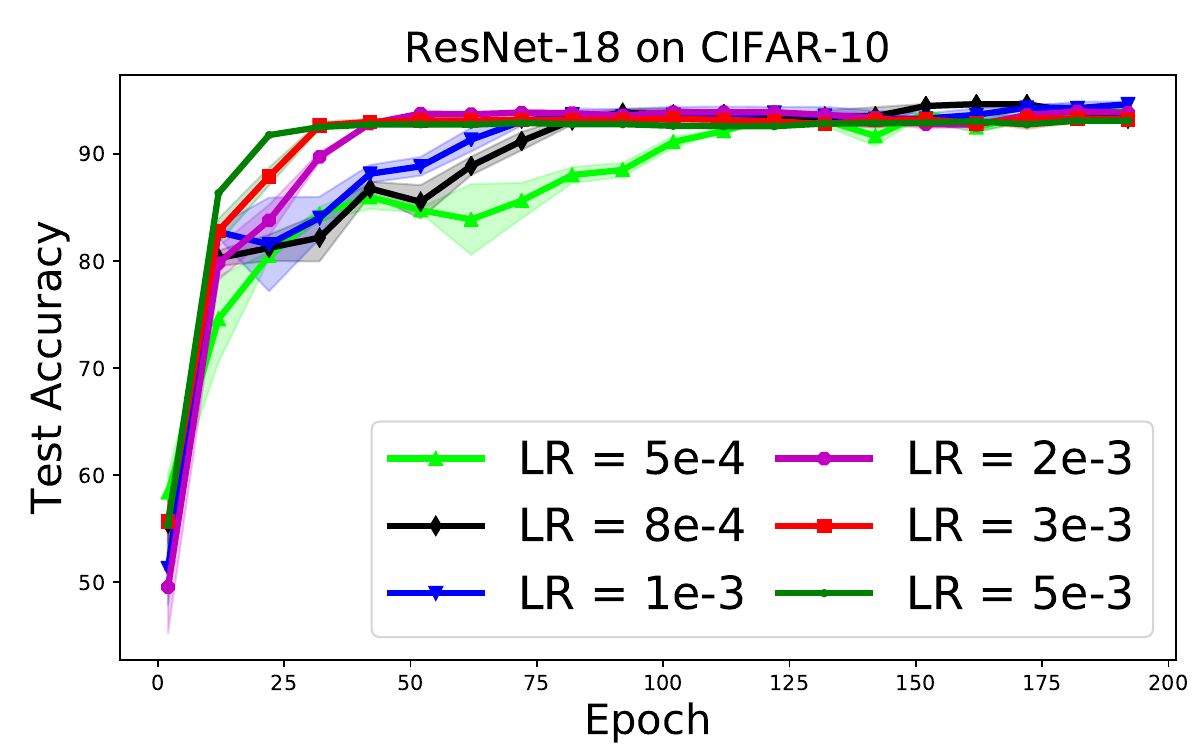}} \vspace{-3mm}
	\caption{Ablation study. Test accuracy on CIFAR-10 with ResNet-18 of (a) different architectures of MLR-SNet and (b) different LRs of meta-optimizer 'Adam'. 
}\label{figablation}  \vspace{-1mm}
\end{figure}

\begin{figure}[t] \vspace{0mm}
	\centering
	\subfigcapskip=-0mm
	\subfigure[]{\label{fig4a}
		\includegraphics[width=0.24\textwidth]{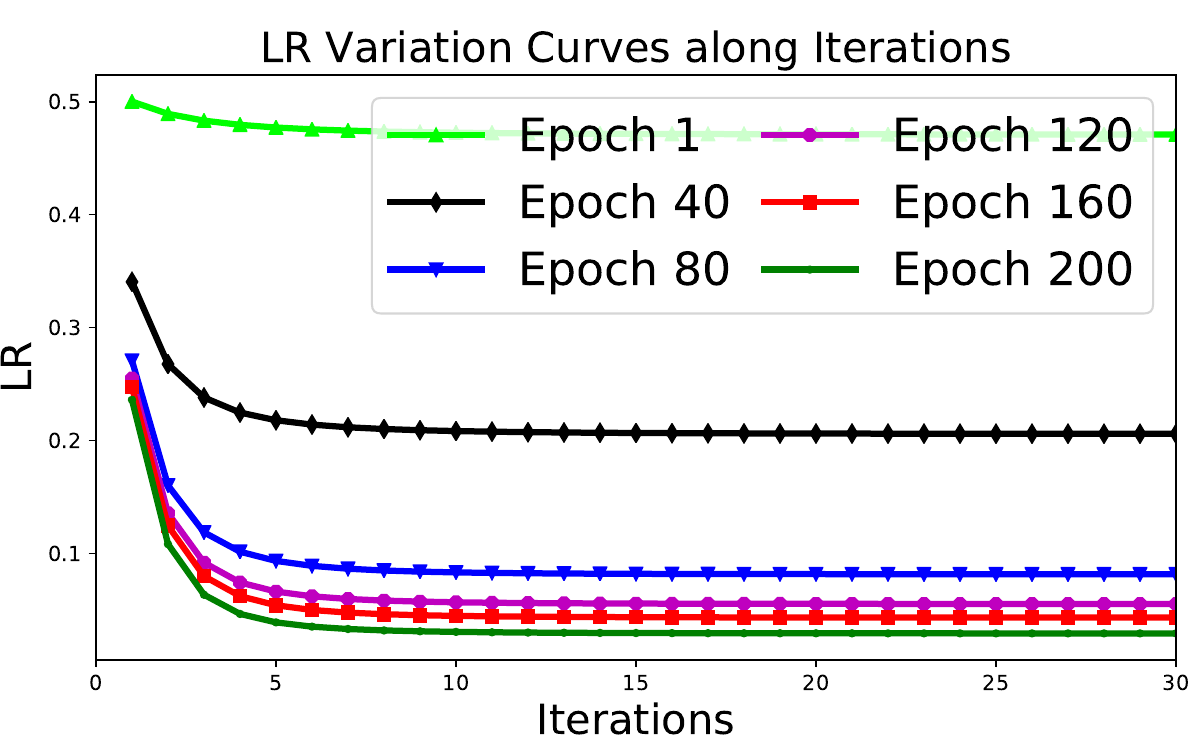} }\hspace{-3mm}
	\subfigure[]{\label{fig4b}
		\includegraphics[width=0.24\textwidth]{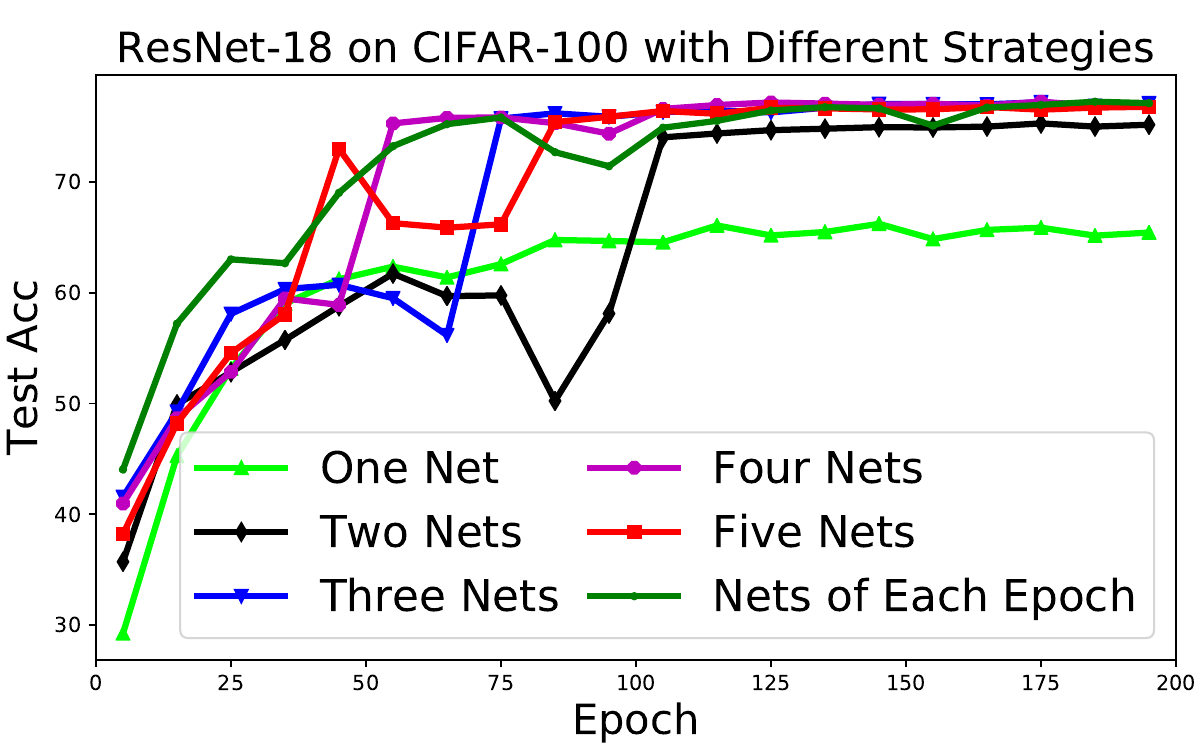}} \vspace{-5mm}
	\caption{(a) The LR variation curves along iterations with the same input loss (we set it as 5) predicted by a single meta-learned MLR-SNet obtained at certain epoch of meta-training stage. As is shown, when iteration increases, the LR is almost constant. This implies that the meta-learned MLR-SNet at certain epoch fails to predict the long trajectories LR. (b) The recording test accuracy on CIFAR-100 with ResNet-18 using different meta-test strategies. }\label{fig4} \vspace{-1mm}
\end{figure}

\vspace{-2mm}
\subsubsection{Ablation Study}
To study individual components and their importance to our proposed method, we conduct experiments above CIFAR-10 with ResNet-18 setting. Fig. \ref{figablation} summarizes the results of ablation studies, as discussed below.

\textbf{The architecture of MLR-SNet.} Fig.\ref{figaa} shows the test accuracy of MLR-SNet on CIFAR-10 with ResNet-18 of different architecture configurations. As can be seen,
our algorithm is not evidently sensitive to the configuration setting of the MLR-SNet's architecture. The depth of the MLR-SNet has unsubstantial difference on the final performance, and thus we choose the one hidden layer in our experiments attributed to its simplicity and low cost for computation. Besides, if the node size of the hidden layer is set small, e.g., 20, it will behave slower at the early training stage compared with larger node size. Therefore, we choose the node size of the hidden layer at a proper size. 
Furthermore, this property shows that our algorithm is robust and can always generally help improve the DNN training performance.

\textbf{The gobal LR of the meta-optimizer}. We adopt Adam optimizer to learn the parameter of the MLR-SNet. One tunable hyperparameter is the global LR of the meta-optimizer. Fig. \ref{figbb} shows the result to further validate whether our MLR-SNet behaves robust to the meta optimizer.
It can be seen that the MLR-SNet achieves almost the similar performance even for different global LRs. This implies that our MLR-SNet is not that sensitive for the setting of this hyper-parameter in the meta optimizer, which makes it easy to reproduce and apply to various problems. We simply set it as $1e^{-3}$ throughout all our experiments.

\begin{figure*}[t] \vspace{-0mm}
	\centering
	\subfigcapskip=-0mm
	\subfigure[Training with 100 Epochs]{
		\label{fig6a} 
		\includegraphics[width=0.30\textwidth]{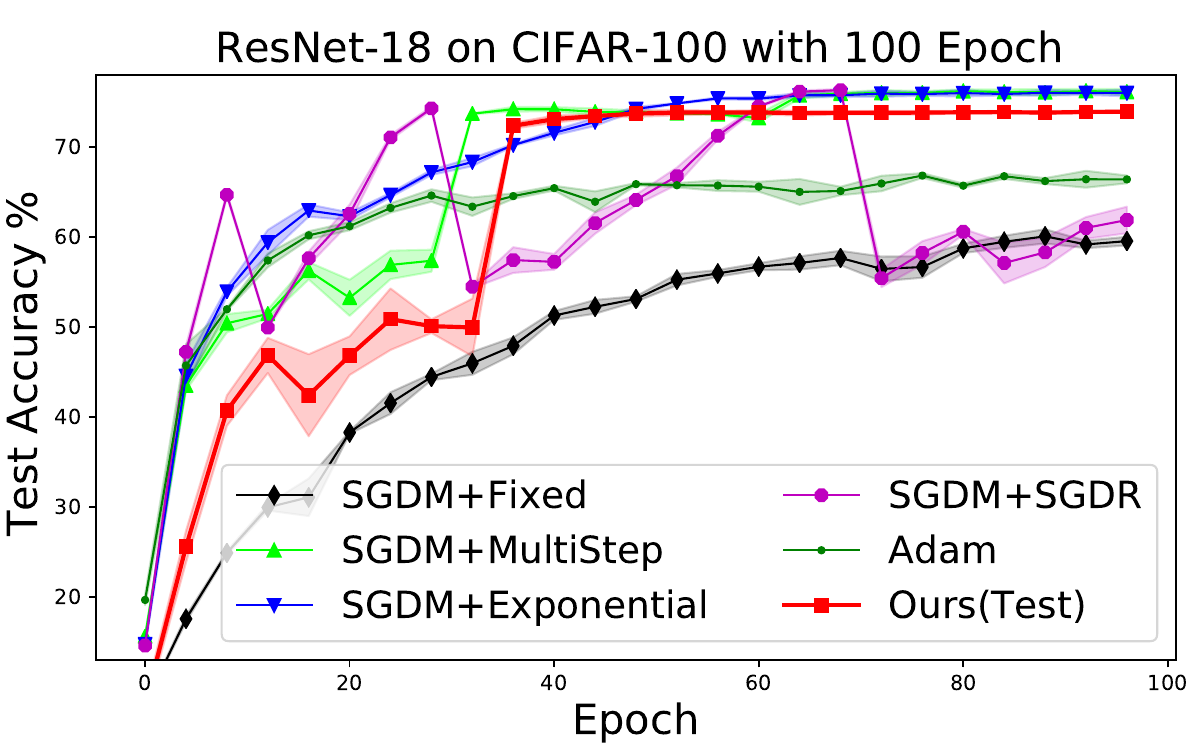}}
	\subfigure[Training with 400 Epochs]{
		\label{fig6b} 
		\includegraphics[width=0.30\textwidth]{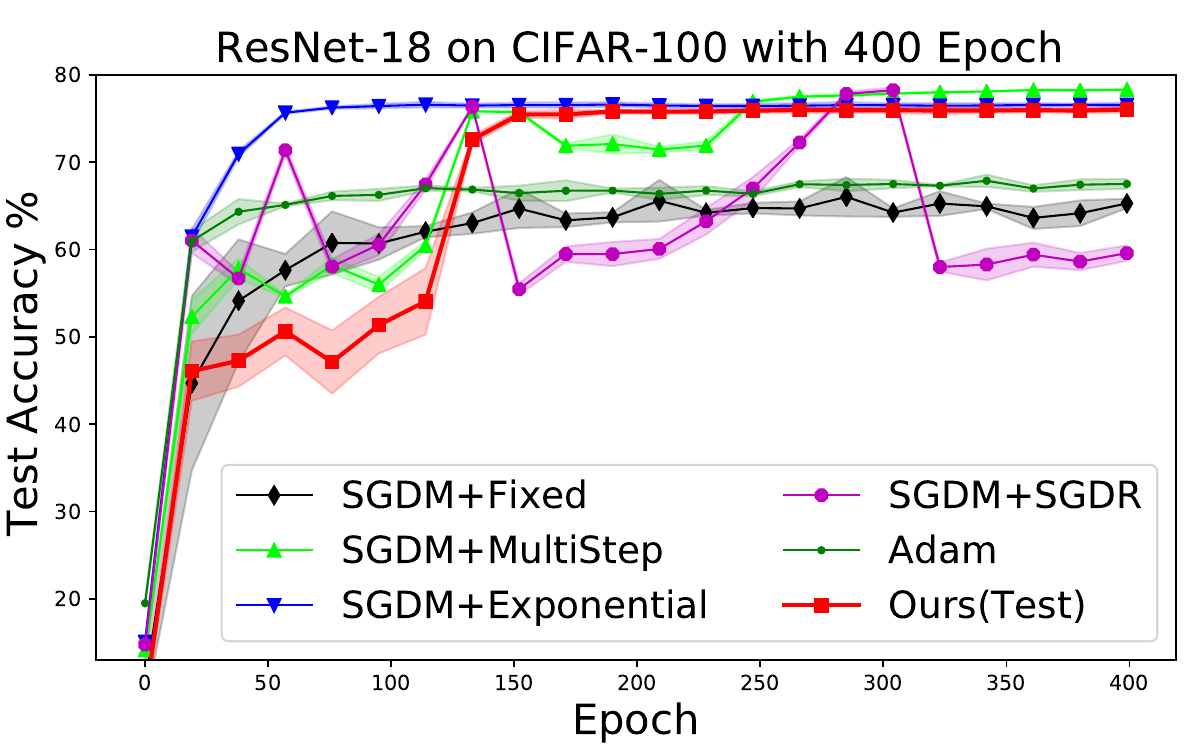}}
	\subfigure[Training with 1200 Epochs]{
		\label{fig6c}  
		\includegraphics[width=0.30\textwidth]{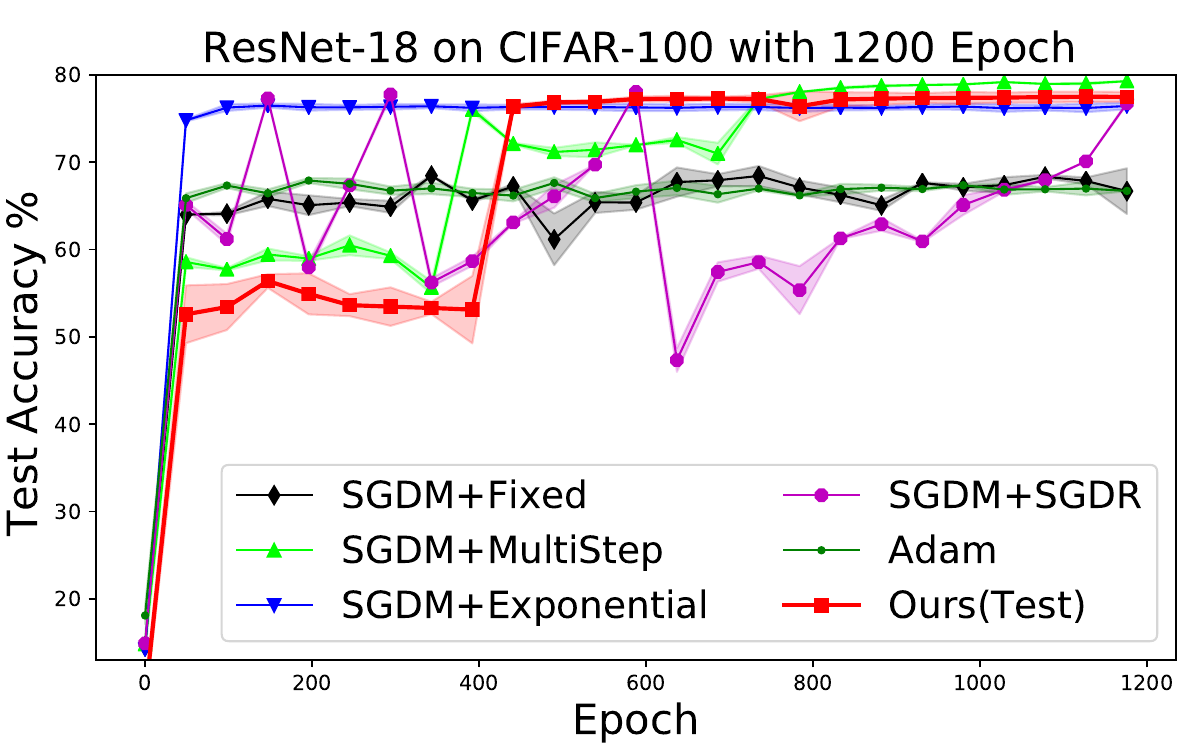}}  \vspace{-2mm}  
	\caption{Test accuracy on CIFAR-100 of ResNet-18 with \textbf{varying epochs} for our transferred MLR-SNet in the meta-test stage. }\vspace{-2mm}\label{fig6000}
\end{figure*}


\vspace{-2mm}
\subsection{Meta-Test: Transferability and Generalization capability of the LR Schedules Meta-learned by MLR-SNet}\label{Transferability}
As aforementioned, the meta-learned LR schedules are transferable and plug-and-play, attributed to its explicit parameterized mapping form. We then validate its transferability and generalization to new heterogeneous tasks.

\subsubsection{Baselines}\label{choose}
The L4, HD, RTHO methods learn the LR schedules specifically for given tasks, and they do not learn transferable structure allowing to be generalized to new tasks. We thus do not compare them in this part. The employed comparison methods for image classification include SGDM\footnote{Here we present stronger baseline results compared with trained with SGD, while our MLR-SNet still predicts LR schedules for SGD.} with hand-designed LR schedules: 1) \textbf{Fixed} LR, 2) \textbf{Exponential} decay, 3) \textbf{MultiStep} decay, and 4) \textbf{SGDR}, as well as the adaptive gradient method \textbf{Adam}. As for the text classification experiments, we compare with SGD and Adam algorithm with Val Strategy LR schedule.

We use the MLR-SNet meta-learned on CIFAR-10 with ResNet-18, as introduced in Section \ref{image}, as the plug-and-play LR schedules to directly predict the LR for SGD algorithm to new heterogeneous tasks. As discussed in Section 3.2.2, we save several meta-learned MLR-SNets at different epochs in the whole one meta-train run for helping setting LR schedules in the meta-testing stage. The motivation can be easily observed from Fig.\ref{fig4a}, which reveals that if we only use the single meta-learned MLR-SNet at certain epoch to predict LR, then the predicted LR will converge to a constant after several iterations. This implies that if we directly select one single MLR-SNet learned by our algorithm, it will raise the risk of the overfitting issue.

This thus inspired us to select more MLR-SNets learned during the meta-training iterations participating in meta-test process.  Generally, if we want to select $k$ nets for meta-test, the MLR-SNet learned at $[\frac{T*l}{k}]$-th epoch ($l=1,2,\cdots,k$) should be chosen, where $[\cdot]$ denotes ceiling operator, and T is the iteration number in training. Fig.\ref{fig4b} show the test accuracy with ResNet-18 on CIFAR-100 of different test strategies, i.e., choosing different $k$ MSR-SNets to transfer. It can be seen that once we choose more than three nets, similar performance can be obtained. We thus easily set $k$ as $3$ throughout all our experiments.

\begin{figure*}[t] \vspace{-0mm}
	\centering
	\subfigcapskip=-1mm
	\subfigure[SVHN dataset]{
		\label{fig7a} 
		\includegraphics[width=0.30\textwidth]{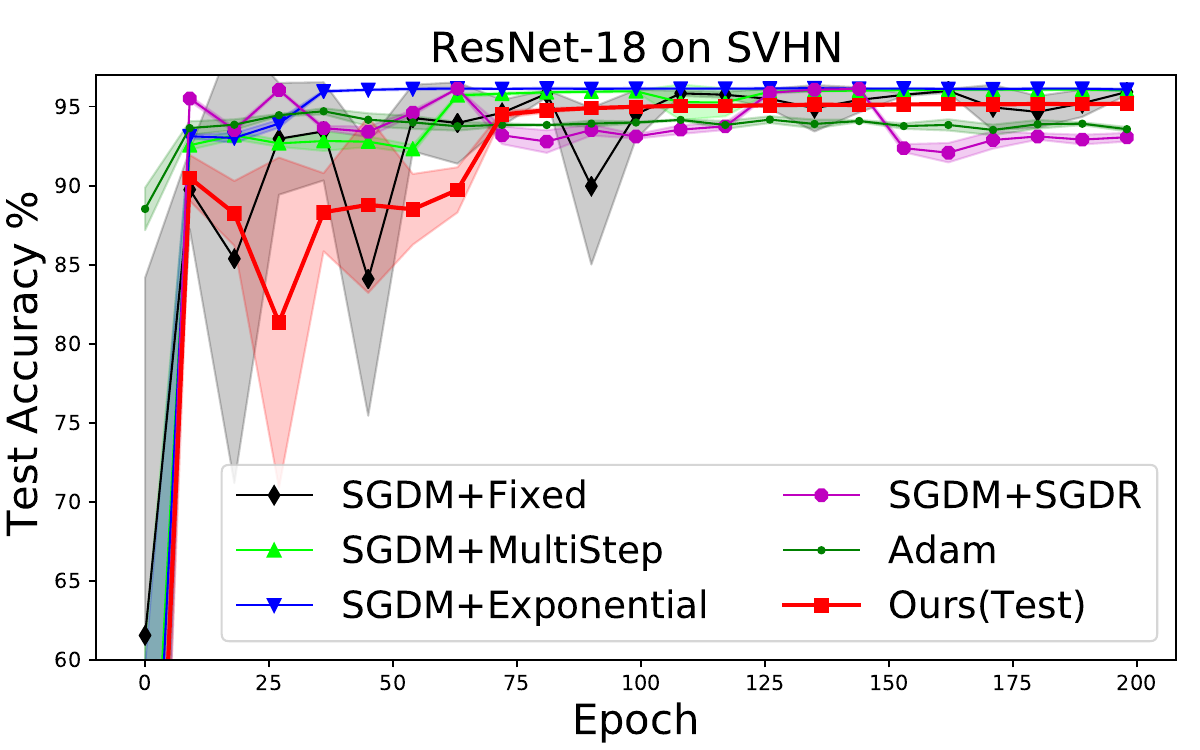}}
	\subfigure[TinyImageNet dataset]{
		\label{fig7b} 
		\includegraphics[width=0.30\textwidth]{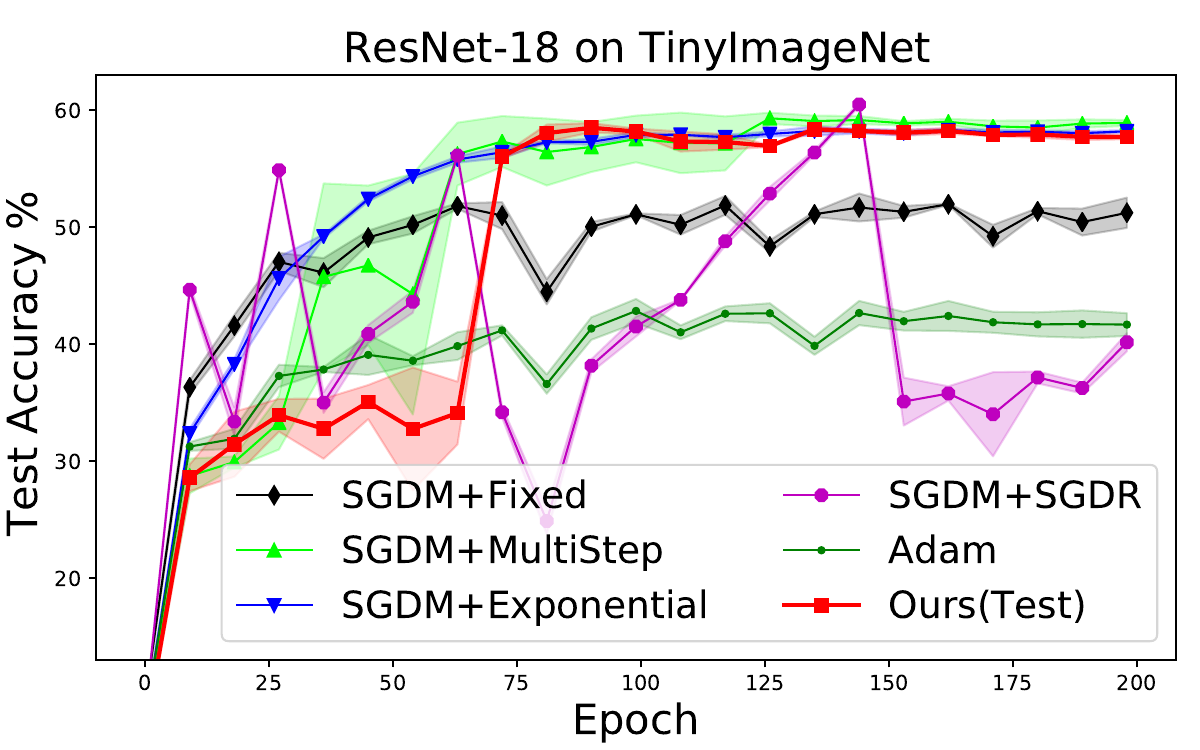}}
	\subfigure[Penn Treebank dataset]{
		\label{fig7c}  
		\includegraphics[width=0.30\textwidth]{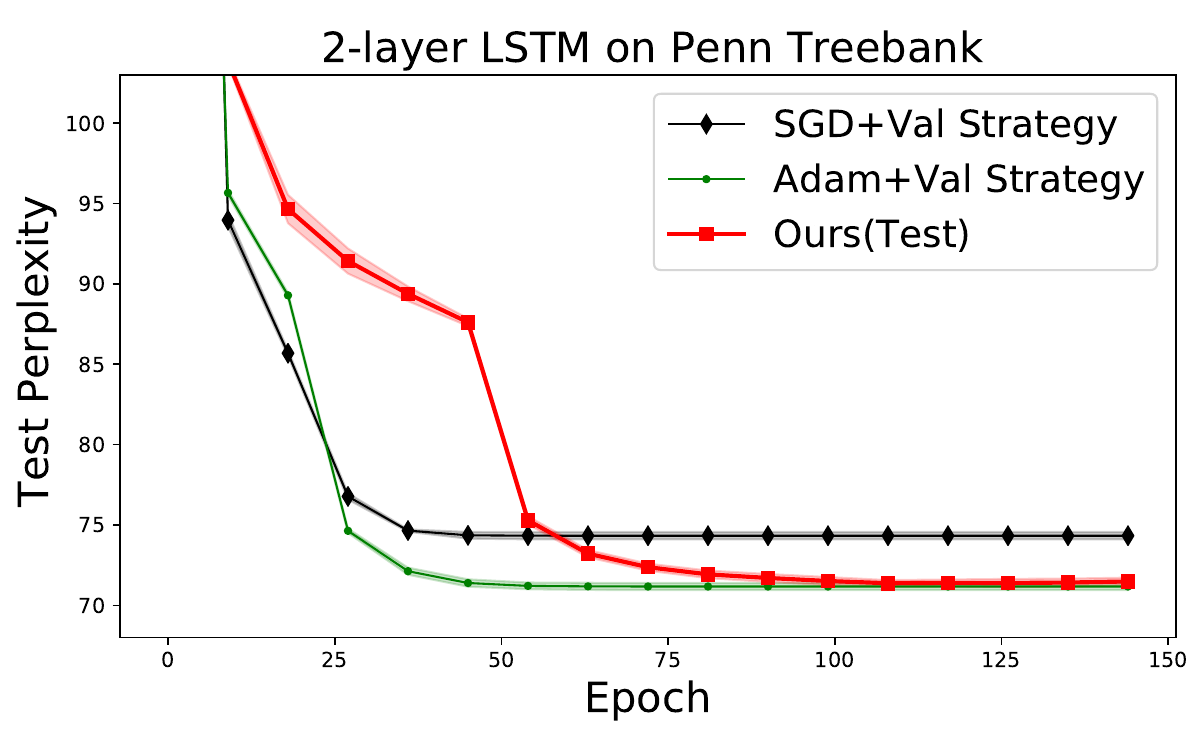}}  \vspace{-2mm}
	\caption{Test accuracy with \textbf{different datasets} for our transferred MLR-SNet in the meta-test stage. }\label{fig7} \vspace{-2mm}
\end{figure*}
\begin{figure*}[t] \vspace{-1mm}
	\centering
	\subfigcapskip=-1mm
	\subfigure[ShuffleNetV2]{
		\label{fig8a} 
		\includegraphics[width=0.31\textwidth]{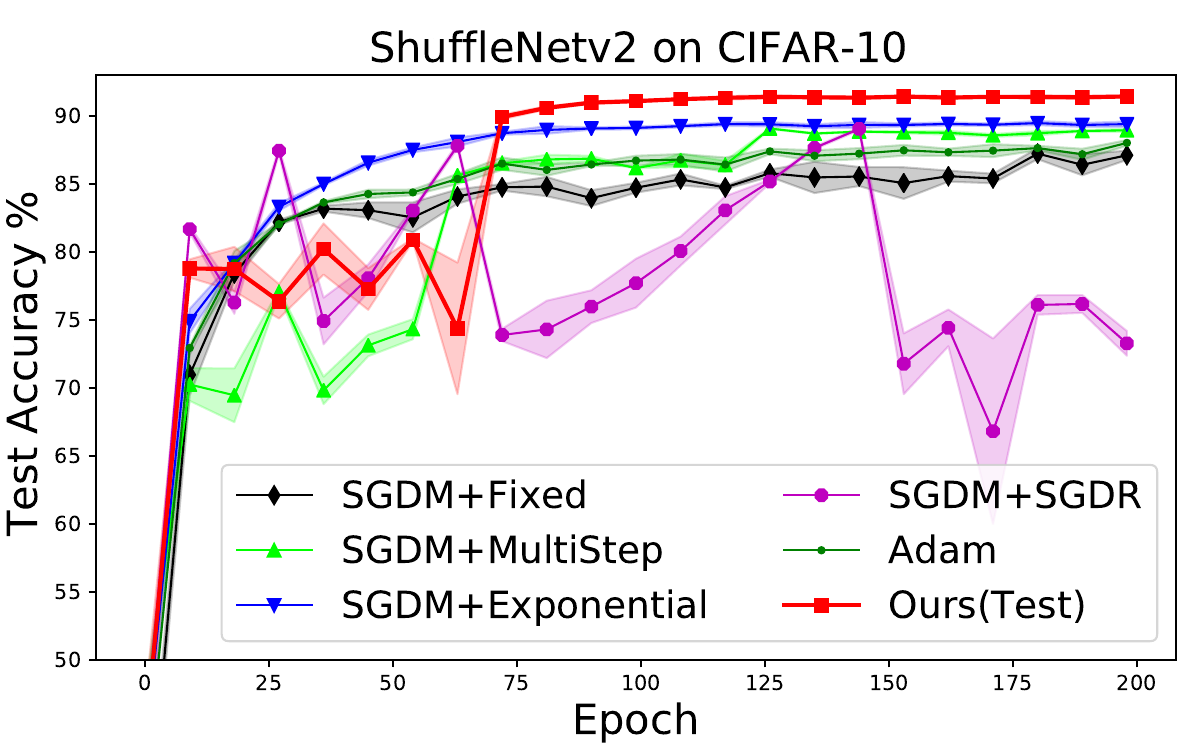}}
	\subfigure[MobileNetV2]{
		\label{fig8b} 
		\includegraphics[width=0.31\textwidth]{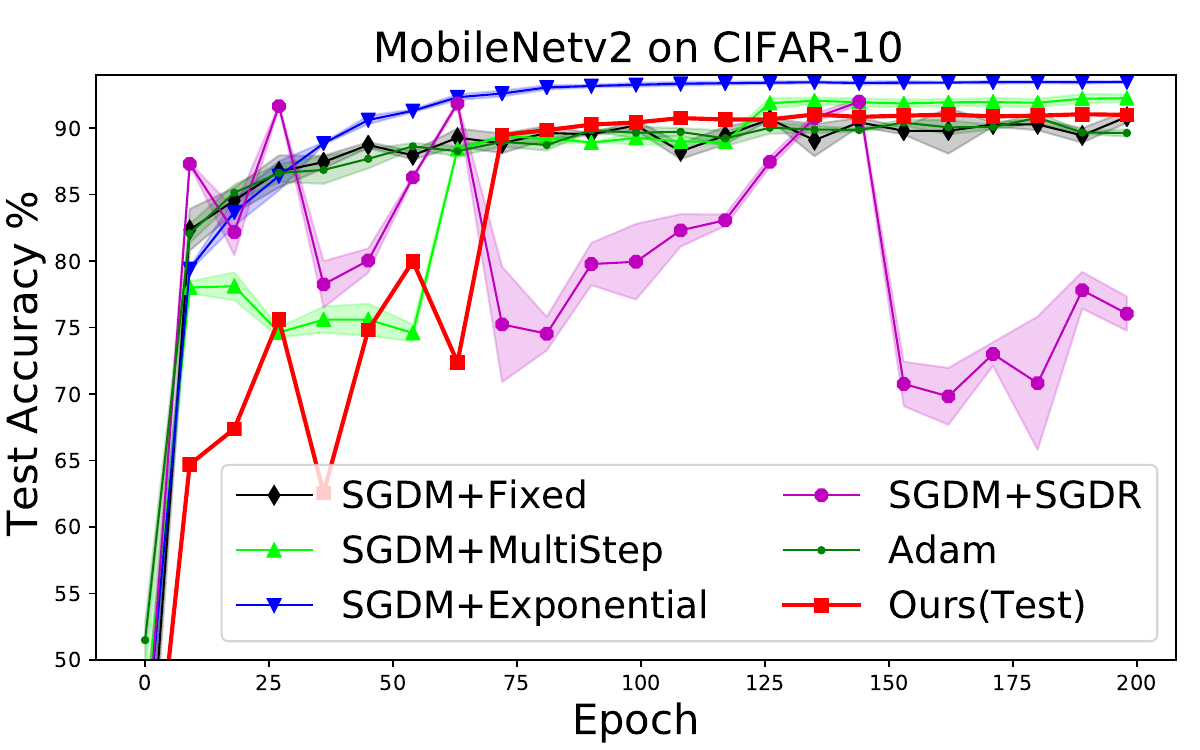}}
	\subfigure[NASNet]{
		\label{fig8c}  
		\includegraphics[width=0.31\textwidth]{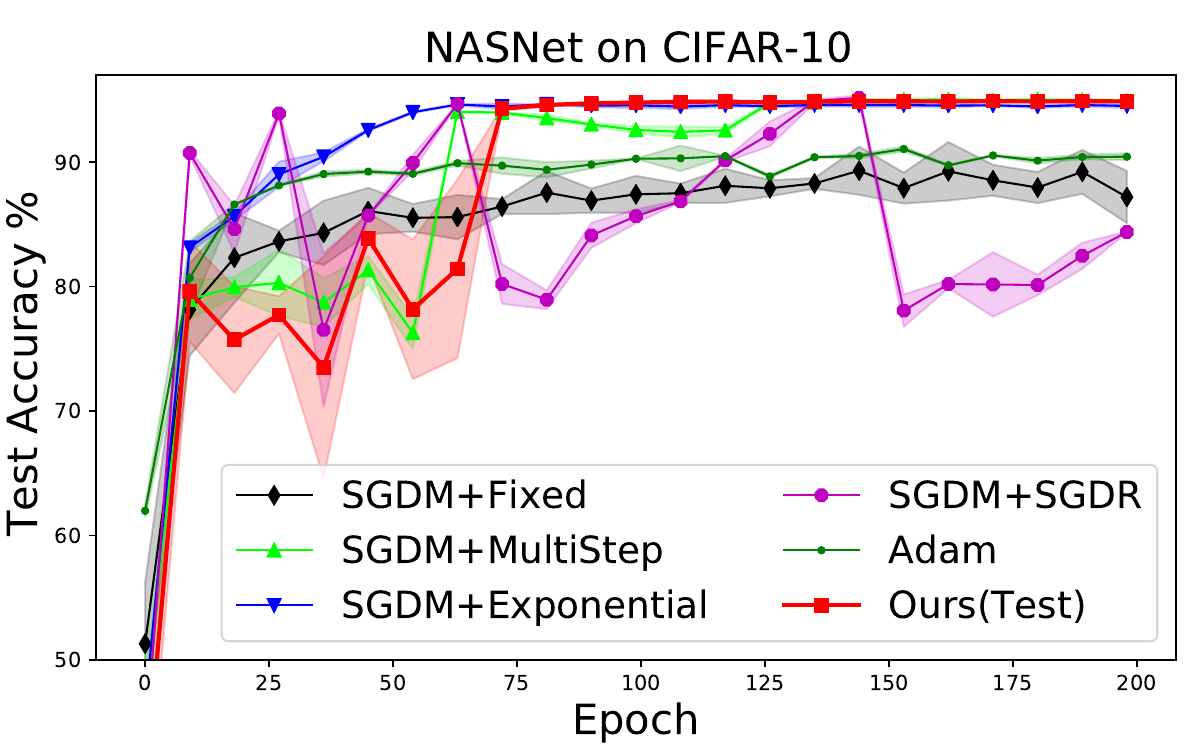}}  \vspace{-2mm}
	\caption{Test accuracy on CIFAR-10 with \textbf{different network architectures} for our transferred MLR-SNet in the meta-test stage. }\label{fig8} \vspace{-2mm}
\end{figure*}

\subsubsection{Generalization to Different Training Epochs}
The plug-and-play MLR-SNet is meta-trained with epoch 200, and we transfer it to other different training epochs, e.g., 100, 400, 1200. All the methods are trained with ResNet-18 on CIFAR-100  with batch size 128 with varying epochs.
The hyper-parameter setting for compared hand-designed LR schedules is the same as that in Section \ref{image} as illustrated above, except for MultiStep LR. For epoch 100, 400 and 1200, \textbf{MultiStep} decays LR by $10$ every 30, 120, 360 epochs, respectively. For our method, we use the transferring MLR-SNet as below: 1) For epoch 100, we employ the 3 nets at 0-33, 33-67, 67-100 epoch, respectively; 2) For epoch 400, we employ the 3 nets at 0-133, 133-267, 267-400 epoch, respectively; 3) For epoch 1200, we employ the 3 nets at 0-400, 400-800, 800-1200 epoch, respectively.

As shown in Fig.\ref{fig6000}, our MLR-SNet has the ability to train the SGD algorithm in the meta-test stage for longer horizons and achieves comparable performance as the best baseline MultiStep LR. The Fixed LR shakes at the later stage for the longer epochs. This substantiated that the learned MLR-SNet is capable of generalized to setting LR schedules with such longer horizons problems.

\subsubsection{Generalization to Different Datasets}
We transfer the LR schedules meta-learned on CIFAR-10 to SVHN \cite{netzer2011reading}, TinyImageNet \footnote{It can be downloaded at https://tiny-imagenet.herokuapp.com.}, and Penn Treebank \cite{marcus19building} datasets to validate the generalization of our method to different datasets, especially varying data modalities. For SVHN and TinyImageNet datasets, we train a ResNet-18 with 200 epoch. For Penn Treebank classification, we train a 3-layer LSTM with 150 epoch. The hyper-parameters of all compared methods are with the same setting as CIFAR-10 and Penn Treebank introduced in Section 4.1. The results are presented in Fig.\ref{fig7}. It is worth noting that the LR schedules for image task and text task have different forms, while our MLR-SNet can still obtain a relatively stable and comparable generalization performance for different tasks with the corresponding best baseline methods.

\begin{figure*}[t] \vspace{-2mm}
	\centering
	\subfigcapskip=-1mm
	\subfigure[Task similarity with CIFAR-100]{
		\label{fig9a} 
		\includegraphics[width=0.30\textwidth]{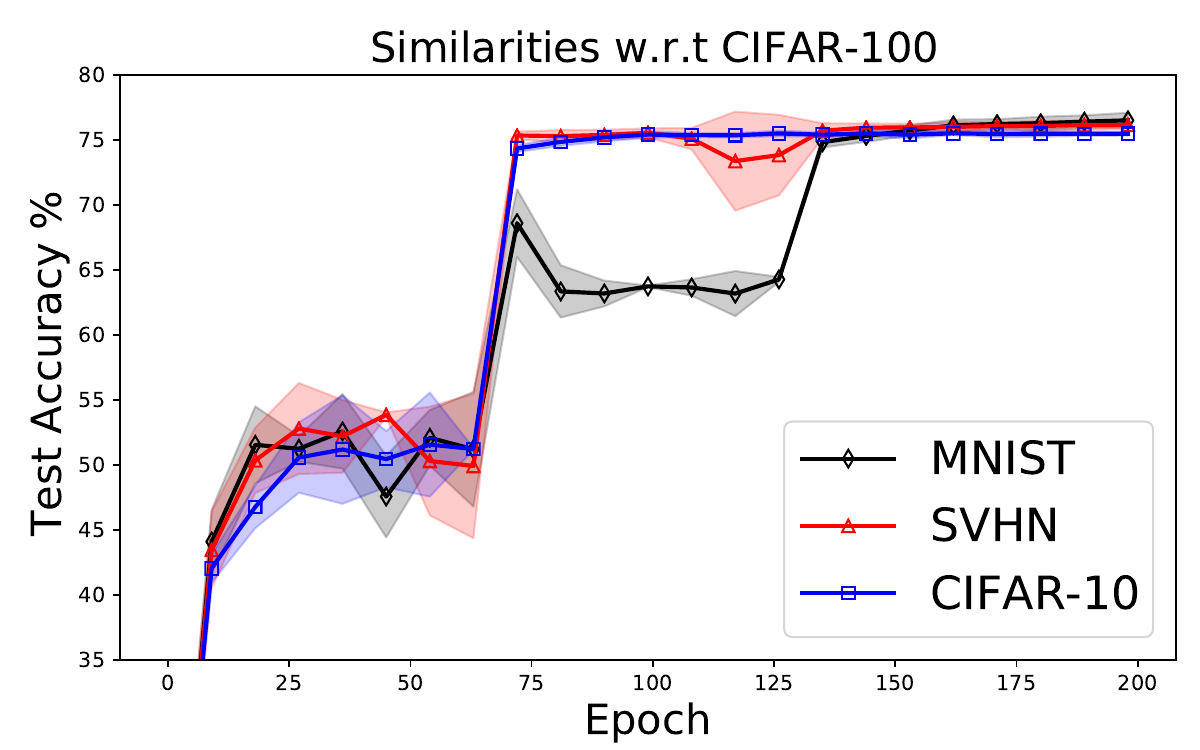}}
	\subfigure[Scale of meta-training tasks]{
		\label{fig9b} 
		\includegraphics[width=0.30\textwidth]{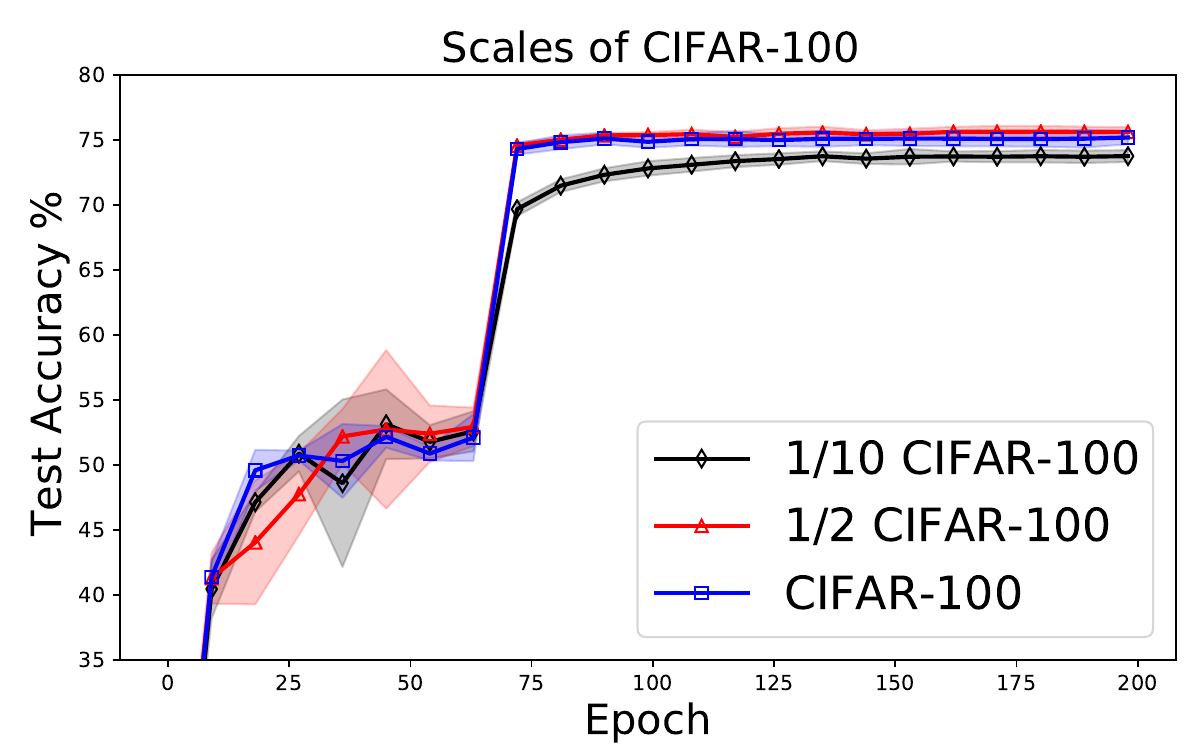}}
	\subfigure[Architectures of training models]{
		\label{fig9c}  
		\includegraphics[width=0.30\textwidth]{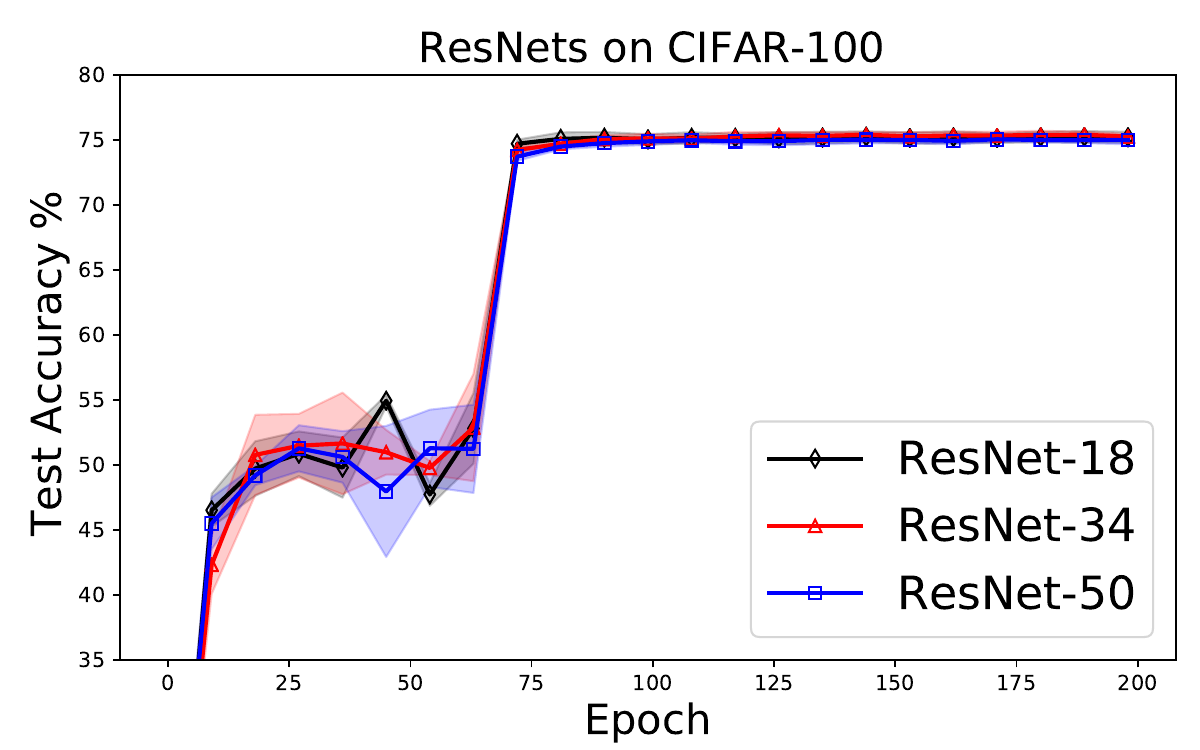}}  \vspace{-3mm}
	\caption{Illustration of meta-training tasks influencing the generalization performance of meta-Learned LR schedules. }\label{fig6} \vspace{-2mm}
\end{figure*}

\subsubsection{Generalization to Different Net Architectures}
To further validate that our method can be applied to different network architectures, we also transfer the LR schedules meta-learned on ResNet-18 to ShuffleNetV2 \cite{ma2018shufflenet}, MobileNetV2 \cite{sandler2018mobilenetv2} and NASNet \cite{zoph2018learning}\footnote{The pytorch codes of all these networks can be found on \\ https://github.com/weiaicunzai/pytorch-cifar100.}. These network architectures are different from ResNet-type network, especially the NASNet is learned from data, not the artificial constructed network. As shown in Fig.\ref{fig8}, our method can achieve comparable results and even get better performance to the best baseline method. This further shows that our MLR-SNet is able to be transferred to varying types of network training.

\subsubsection{Generalization to Large Scale Optimization Problem}
In this part, we attempt to use the meta-learned LR schedules to train DNN on ImageNet dataset \cite{deng2009imagenet}.
To our best knowledge, only \cite{wichrowska2017learned} had attempted this task among existing learning-to-optimize literatures. However, it can only be executed for thousands of steps, and then its loss begins to increase dramatically, thus not able to be implemented in the optimization process in practice. We transfer the LR schedules meta-trained on CIFAR-10 with ResNet-18 to ImageNet dataset with ResNet-50\footnote{The training codes of baseline methods can be found on \\ https://github.com/pytorch/examples/tree/master/imagenet.}. All compared methods are trained by SGDM with a momentum $0.9$, a weight decay $5e^{-4}$, an initial learning rate $0.1$ for 90 epochs, and batch size 256.
\textbf{MultiStep} decays LR by $10$ every 30 epochs; \textbf{Exponential} multiplies LR with $\gamma_E = 0.95$ every epoch; \textbf{SGDR} sets $\alpha_{\min}=1e^{-5},\alpha_{\max}=0.1$, and $E_0=10, T_{Mult}= 2$. Following \cite{wilson2017marginal}, we decay global LR by $10$ every 30 epochs for Adam.

The test accuracy on ImageNet validation set is presented in Fig.\ref{figimg}. It can be seen that the performance of our method is competitive with those hand-designed LR schedules methods, though we train the model with SGD using the LR schedules predicted by our transferred MLR-SNets. Meanwhile, the LR schedules predicted by MLR-SNet brings non-extra computation complexity in the DNN training process. This implies that our method is hopeful to be effectively and efficiently used to deal with such large scale optimization problems, making learning-to-optimize ideas towards more practical applications.

\subsection{How do Meta-Training Tasks Influence the Generalization Performance of Meta-Learned LR Schedules} \label{how}
In this section, we empirically study how meta-training tasks influence the generalization performance of meta-learned LR schedules. To conduct ablation study for answering this question, we construct three groups of meta-training tasks to character the influence factors for the generalization performance. An overview of them is shown in Table \ref{table4}. The meta-test task is set as training a ResNet-18 on full CIFAR-100 with meta-learned LR schedules. The hyperparameter setting follows those introduced in Section \ref{image}.

%

\textbf{The similarity between meta-training and meta-test tasks.} Grayscale digits (MNIST), RGB digits (SVHN) and natural photos (CIFAR-10) represent incremental similarity between meta-training and meta-test tasks. We use the three datasets to meta-learn MLR-SNet with ResNet-18, respectively. As shown in Fig. \ref{fig9a}, three transferred LR schedules meta-learned from different datasets achieve very similar final performance on the meta-test task.
This validates that such similarity difference has a relatively weak influence on the generalization of meta-learned LR schedules.

\textbf{Scale of meta-training tasks.} The scale of meta-training tasks is also taken into consideration. We uniformly sampled 50, 250, 500 samples per class in CIFAR-100 as training datasets, denoted by 1/10 CIFAR-100, 1/2 CIFAR-100 and CIFAR-100, respectively. We use the three datasets to meta-learn MLR-SNet with ResNet-18. Fig. \ref{fig9b} shows the generalization performance of three kinds of such meta-learned LR schedules. As is shown, the performance deteriorates when the size of training task set is small. If the scale of training task set is in the same order of magnitude, it tends to obtain similar generalization performance.

\textbf{Architectures of training models.} Different network architectures in the meta-training stage may produce different LR schedules. We adopt three different classifier networks, including ResNet-18, ResNet-34, and ResNet-50, to meta-learn MLR-SNet on CIFAR-100. Fig. \ref{fig9c} shows that three transferred LR schedules achieve similar generalization performance, even though they are meta-learned based on different classifier networks.

\textbf{Remark.} We have empirically verified that the generalization performance of the meta-learned LR schedules is not sensitive to the similarity between meta-training and meta-test tasks, and network architectures in the meta-training stage. This can be rationally explained by the fact that our MLR-SNet is sufficiently simple to make it less rely on the task-related information. Besides, it is also verified that the size of meta-training task could slightly influence the final generalization performance. This might possibly due to that few meta-training data could not provide enough information to fit the proper LR schedules.
Furthermore, these empirically results state that our MLR-SNet is easy to be meta-trained for achieving an admirable performance on the meta-test tasks.

\begin{table}[t]\vspace{-0mm}
	\caption{ Variants constructed from meta-training tasks.}\label{table4}\vspace{-2mm}
	\centering
	\begin{threeparttable}
	\begin{tabular}{l|c}
		\toprule
		Influence factors	& Tasks design \\
		\hline \hline
		Task similarity	& MNIST$^a$, SVHN$^a$, CIFAR-10  \\
		\hline
		Task scale	& 1/10 CIFAR-100$^b$, 1/2 CIFAR-100$^b$, CIFAR-100  \\
		\hline \hline
		Architecture	& ResNet-18, ResNet-34, ResNet-50 \\
		\bottomrule
	\end{tabular}
\begin{tablenotes}
	\item $a$: \ uniformly downsample to 50000 samples
	\item $b$: \ uniformly sample to certain proportion of full CIFAR-100
	\end{tablenotes}
\end{threeparttable}
\end{table}\vspace{-4mm}

\begin{table*}[t]\vspace{-2mm}
	\caption{Test accuracy (\%) on CIFAR-10 and CIFAR-100 training sets of different methods trained on CIFAR-10-C and CIFAR-100-C. Best and Last denote the best test result and the last epoch test result, respectively. The \textbf{Bold} and \textbf{\underline{Underline Bold}} denote the first and second best results, respectively.}\label{tablerobust1} \vspace{-2mm}
	\centering
	\begin{normalsize}
		\begin{tabular}{c|c|c|c|c|c|c|c}
			\toprule
			\multicolumn{2}{c|}{Datasets/Methods} & Fixed & MultiStep &Exponential & SGDR&Adam&Ours(Train)\\ \hline
			\multirow{2}{*}{CIFAR-10-C} & Best &  79.78$\pm$3.95        &    85.52$\pm$1.72       &  83.48$\pm$1.45       &  85.94$\pm$1.52      &  81.45$\pm$1.42       &  \textbf{86.04$\pm$1.51 }   \\
			& Last &  77.88$\pm$3.91        &  85.36$\pm$1.71        &  83.32$\pm$1.43         & 78.21$\pm$2.01        & 80.29$\pm$1.64        & \textbf{85.87$\pm$1.54  }   \\ \hline
			\multirow{2}{*}{CIFAR-100-C} &Best &  46.74$\pm$3.03        &    52.26$\pm$2.58       &  49.72$\pm$1.97     & 52.54$\pm$2.49      & 45.45$\pm$1.94       &  \textbf{52.56$\pm$2.26 }   \\
			&Last &  44.79$\pm$3.91        &  52.16$\pm$2.59        & 49.58$\pm$1.98        & 41.58$\pm$3.24      & 43.76$\pm$2.22        & \textbf{52.42$\pm$2.34  }   \\
			\bottomrule
		\end{tabular} \vspace{-2mm}
	\end{normalsize}
\end{table*}
\begin{table*}[t]
	\caption{Test accuracy (\%) on CIFAR-10 and CIFAR-100 training sets of different methods trained on CIFAR-10-C and CIFAR-100-C. Best and Last denote the best test result and the last epoch test result, respectively. The \textbf{Bold} and \textbf{\underline{Underline Bold}} denote the first and second best results, respectively.}\label{tablerobust2} \vspace{-2mm}
	\centering
	\begin{normalsize}
		\begin{tabular}{c|c|c|c|c|c|c|c}
			\toprule
			\multicolumn{2}{c|}{Datasets/Methods} & Fixed & MultiStep &Exponential & SGDR&Adam&Ours(Train)\\ \hline
			\multirow{2}{*}{CIFAR-10-C} & Best &  79.96$\pm$4.09        &    85.64$\pm$1.71       &  83.63$\pm$1.38      &  \textbf{86.10$\pm$1.44}      &  81.57$\pm$1.39       &  \underline{\textbf{85.73$\pm$1.71 }}   \\
			& Last &  77.89$\pm$4.05       & \underline{ \textbf{85.48$\pm$1.71 }}     &  83.47$\pm$1.37         & 78.46$\pm$1.92        & 80.39$\pm$1.65       &\textbf{85.62$\pm$1.76 }   \\ \hline
			\multirow{2}{*}{CIFAR-100-C} &Best &  46.91$\pm$3.08       &    52.38$\pm$2.43       &  49.90$\pm$1.93     & \textbf{52.80$\pm$2.39}     & 45.58$\pm$1.95       &  \underline{\textbf{52.51$\pm$2.38}}  \\
			&Last &  44.81$\pm$5.98        &  \underline{\textbf{52.28$\pm$2.44} }       & 49.75$\pm$1.94        & 41.68$\pm$3.33     & 43.94$\pm$2.18       & \textbf{52.35$\pm$2.46 }\\
			\bottomrule
		\end{tabular} \vspace{-2mm}
	\end{normalsize}
\end{table*}

\subsection{Robustness on Data Corruptions} \label{robust}
In this section, we further validate whether our MLR-SNet behaves robust against corrupted training data guided by a clean validation set.
To this aim, we design experiments as follows: we take CIFAR-10-C and CIFAR-100-C \cite{hendrycks2019benchmarking} as our training set\footnote{They can be downloaded at https://zenodo.org/record/2535967\#.\\ Xt4mVigzZPY and https://zenodo.org/record/3555552\#.Xt4mdSgzZPY.}, consisting of 15 types of algorithmically generated corruptions from noise, blur, weather, and digital categories. These corruptions contain Gaussian Noise, Shot Noise, Impulse Noise, Defocus Blur, Frosted Glass Blur, Motion Blur, Zoom Blur, Snow, Frost, Fog, Brightness, Contrast, Elastic, Pixelate and JPEG. All the corruptions are generated on 10,000 test set images of CIFAR-10/100 dataset, and each corruption contains 50,000 images since each type of corruption has five levels of severity. We treat CIFAR-10-C or CIFAR-100-C dataset as training set, and the original training set of CIFAR-10 or CIFAR-100 as test set. We train models with ResNet-18 for each corrupted dataset. Finally, we can obtain 15 models for CIFAR-10-C or CIFAR-100-C dataset. The average accuracy of 15 models on test data is used to evaluate the robust performance of each LR schedules strategy. All compared hand-designed LR schedules are trained with a ResNet-18 by SGDM with a momentum $0.9$, a weight decay $5e^{-4}$, an initial learning rate $0.1$ for 100 epochs, and batch size 128. \textbf{Exponential} LR  multiplies LR with $0.95$ every epoch; \textbf{MultiStep} LR decays LR by $10$ every 30 epochs; \textbf{SGDR} sets $\alpha_{\min}=1e^{-5},\alpha_{\max}=0.1$, and $E_0=10, T_{Mult}= 2$;
\textbf{Adam} just uses the default parameter setting. We update the MLR-SNet under the guidance of a small set of validation set without corruptions, to guarantee that the final learned models finely generalize to clean test set. We randomly choose 10 clean images for each class as validation set in this experiment.

Table \ref{tablerobust1} shows the mean test accuracy of 15 models ($\pm$std) on the training set of CIFAR-10 or CIFAR-100 dataset. As can be seen, our proposed MLR-SNet is capable of achieving better generalization performance on clean test data than baseline methods, which implies that our method behaves more robust and stable than the pre-set LR schedules when the learning tasks in which the distribution of training and test data are mismatched. This is due to the fact that our MLR-SNet has more flexibility to adapt the variation of the data distribution than the pre-set LR schedules, and it can find a proper LR schedule through minimizing the generalization error which is based on the knowledge specifically conveyed from the given validation data.

Furthermore, we attempt to explore the generalization for our meta-learned LR schedules. Different from the above experiments where all 15 models are trained under the guidance of a small set of validation set, we just meta-learn the MLR-SNet on Gaussian Noise corruption dataset, and then transfer the meta-learned LR schedules to other 14 corruptions datasets. We report the average accuracy of 14 models on test data to show the robust performance of our transferred LR schedules. All the methods are meta-tested with a ResNet-18 for 100 epochs with batch size 128. The hyper-parameter setting of hand-designed LR schedules keeps the same as above. Table \ref{tablerobust2} shows the mean test accuracy of 14 models on the training set of CIFAR-10 or CIFAR-100 dataset. As can be seen, our transferred LR schedules obtain the best performance in the last epoch compared with hand-designed LR schedules.
This implies that our transferred LR schedules can also perform robust and stable for the learning tasks in which the distribution of training and test data are mismatched. Besides, our transferring LR schedules are plug-and-play, and have no additional hyper-parameters to tune when transferred to new heterogeneous tasks.


\begin{figure*}[t]\vspace{-4mm}
	\centering
	\subfigcapskip=-0mm
		\subfigure[]{\label{figimg}
		\includegraphics[width=0.33\textwidth]{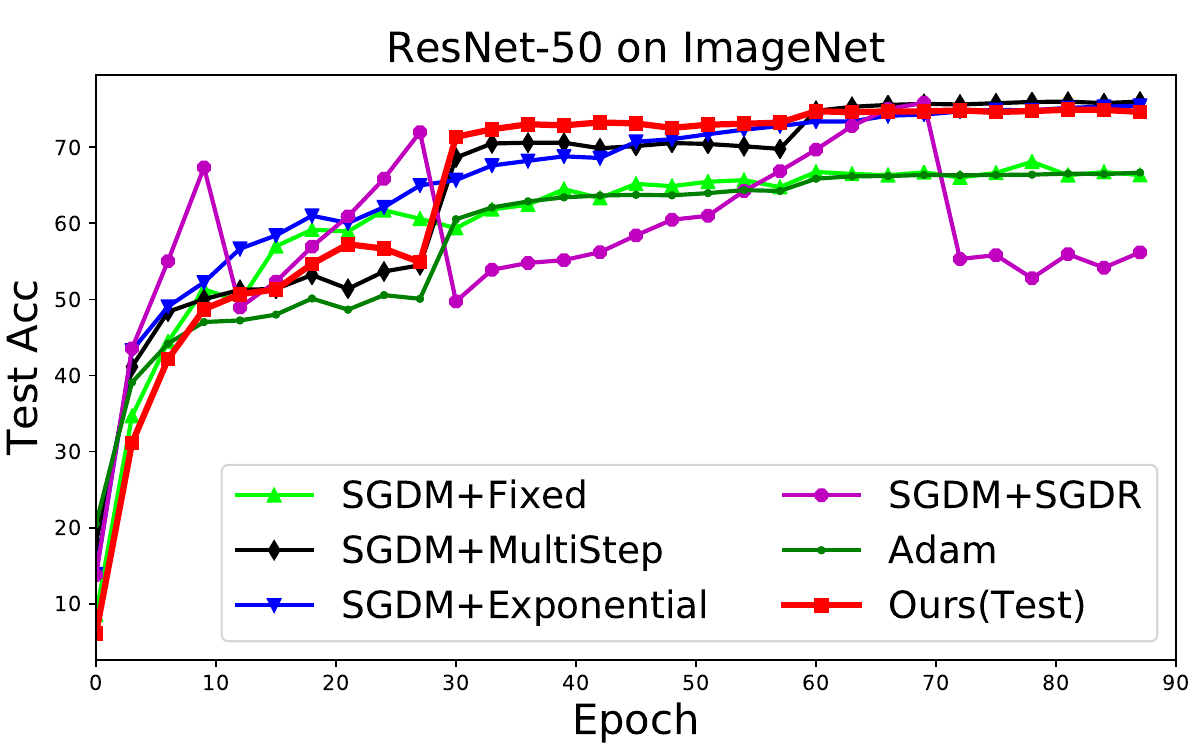}} \hspace{-3mm}
	\subfigure[]{\label{figtime}
		\includegraphics[width=0.33\textwidth]{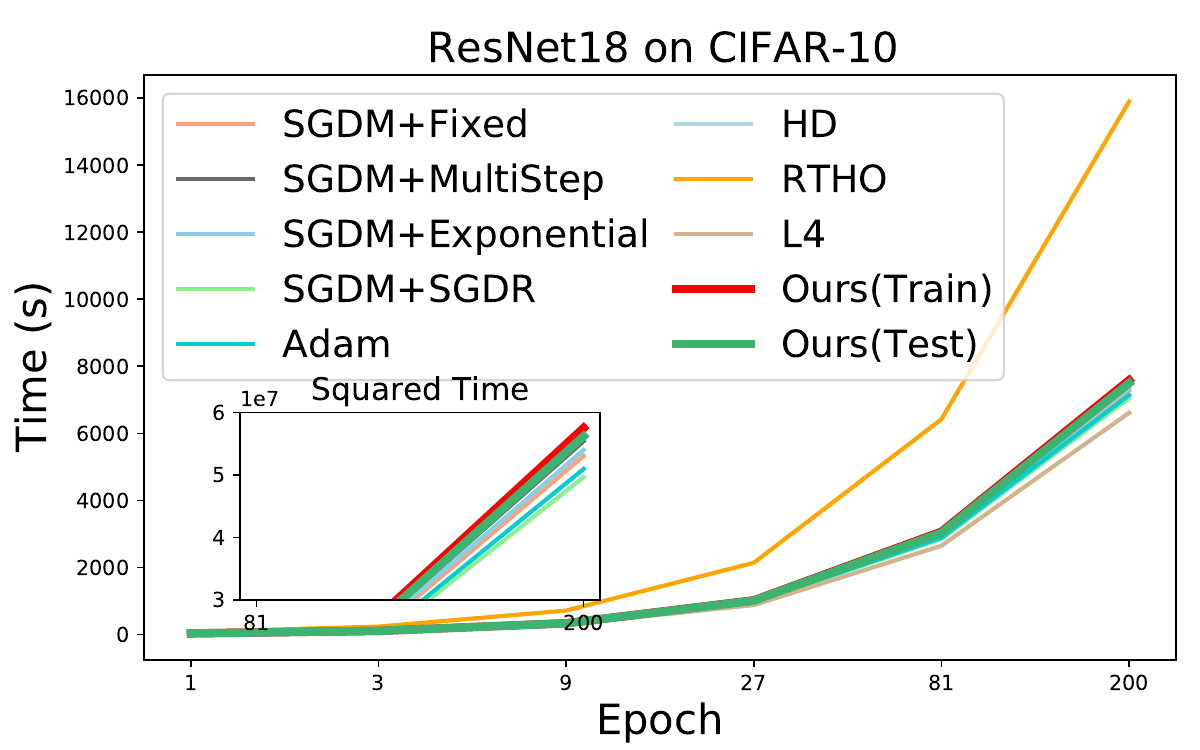}} \hspace{-3mm}
	\subfigure[]{\label{curve}
		\includegraphics[width=0.33\textwidth]{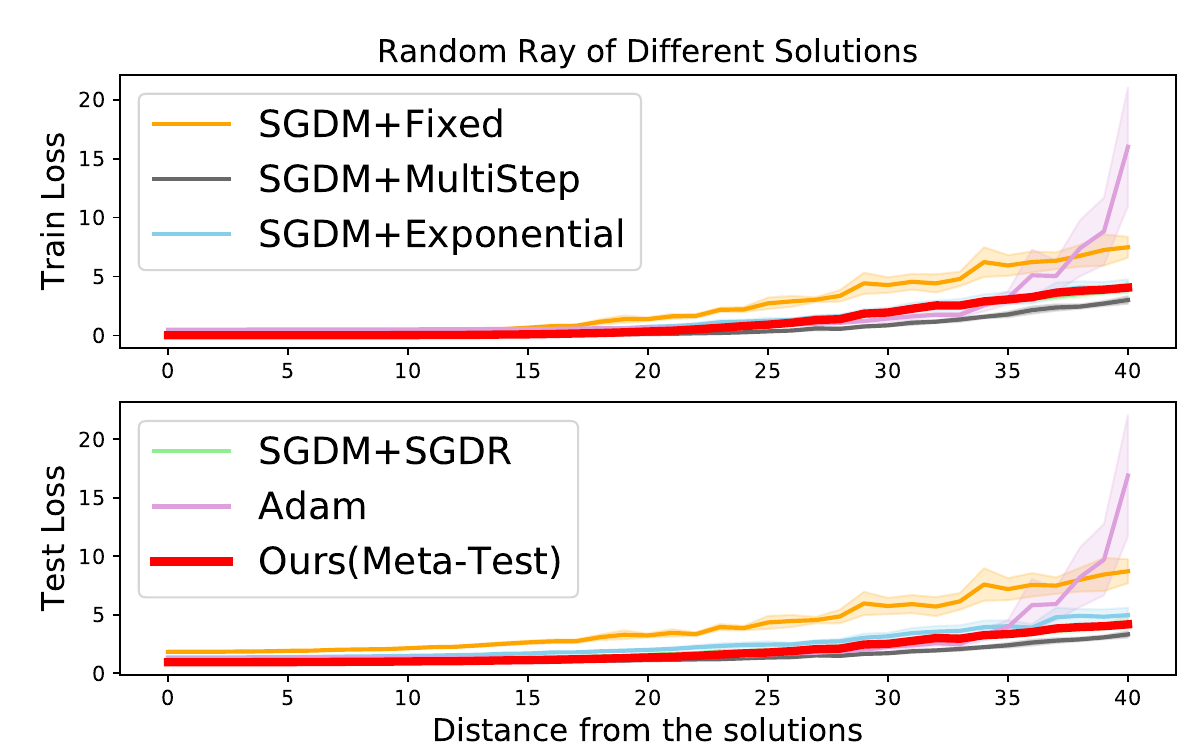}} \vspace{-2mm}
	\caption{(a)Test accuracy on ImageNet validation set with ResNet-50. (b)Computational time costed by different LR schedule methods. (c) (\textbf{Upper}) Train lossf and (\textbf{Lower}) test loss as a function of a point on a random ray starting at the solutions for different methods on CIFAR-100 with ResNet-18. } \vspace{-2mm}
\end{figure*}


\section{Further Analysis on MLR-SNet}\label{analysis}
In this section, we firstly provide the convergence guarantee for the SGD algorithm with LR schedules produced by our MLR-SNet, as well as the convergence guarantee for the meta-learning of the MLR-SNet (Section \ref{convergence}). In Section \ref{computation}, we further analyze the computational complexity for the MLR-SNet. The ``width'' of the solution is visualized in Section \ref{width}. In Section \ref{lstm-type}, we further verify that the LSTM-type meta-learner behaves more superiorly than MLP-type meta-learner . Finally, we show that the MLR-SNet can be applied to Adam optimizer in Section \ref{adam}.

\subsection{Convergence Analysis of MLR-SNet}\label{convergence}
The preliminary experimental evaluations show that our method gives good convergence performance on various tasks. We find that the meta-learned LR schedules in our experiments follow a consistent trajectory as shown in Fig.\ref{ss}, almost obeying a decay LR form. Without loss of generality, we assume that the learning rate can be represented by
\begin{align}\label{eqlr}\alpha_t = \alpha_{t-1} \beta_{t}, \ t = 1,2,\cdots,T,\end{align}
 where $\alpha_t$ denotes the learning rate predicted by MLR-SNet at the $t$-th iteration, and $\beta_{t}$ denotes the decay factor at the $t$-th iteration, $1/K \leq a \leq \beta_t \leq  b \leq 1$, where $a=(M/T)^{1/T}, b= (N/T)^{1/T}$, and $a\neq b, M,N \propto T$, and $K$ is the arbitrarily large constant. We denote by $\mathbb{E}[\cdot]$ the expectation with respect to the underlying probability space. To present the convergence results, we also assume that \footnote{They are commonly used for existing SGD convergence theories \cite{karimi2016linear,reddi2016stochastic,bottou2018optimization}.}:

(\textbf{A1}) The loss function $f(w): \mathbb{R}^d \rightarrow \mathbb{R}$  is $L$-smooth, i.e., $f$ is differentiable and its gradient $\nabla f(w)$ is  $L$-Lipschitz.

(\textbf{A2}) $f$ satisfies the $\mu$-PL condition, that is, their exists some $\mu>0$, $\frac{1}{2}\|\nabla f(w)\|^2 \geq \mu(f(w)-f^*),$ holds for any w, where $f^*$ represents the infimum of $f(w)$.

(\textbf{A3}) For $t\!=\!1,2, \cdots, T$, we assume $\mathbb{E}_t [\|v_t - \nabla f(w_t)\|^2] \leq \kappa\|\nabla f(w_t)\|^2 +\sigma$, where $\kappa,\sigma>0$, and $v_t$ is an unbiased estimate of the gradient of $f$ at point $w_t$, i.e., $\mathbb{E}_t v_t = \nabla f(w_t)$.

Firstly, we consider the case where the function is smooth and satisfies the Polyak- Lojasiewicz (PL) condition \cite{polyak1963gradient,lojasiewicz1963topological}. The proofs of all Theorems are listed in the appendix file.
\begin{Theorem}
	Assume (\textbf{A1},\textbf{A2},\textbf{A3}) hold, and the SGD is with learning rate (\ref{eqlr}), where $\alpha_0 = (L(1+\kappa))^{-1}$. Then for a given $T \geq \max\{3,M,N\}$, the $w_t$ generated from SGD satisfies
	\begin{align*}
	\mathbb{E} f(\mathbf{w}_{t+1}) \!-\!f^* \!\leq\! &  C(M)  \exp\left(-  \frac{\mu T}{KL(1+\kappa)\ln(T/M)}\right) \\
(f(\mathbf{w}_1)-f^*)	&+\frac{2K^2C(M)\ln^2(T/M)(N/M)^2}{e^2\mu^2(1-M/N) T}.
	\end{align*}
	where $C(M) =  \exp(\frac{\mu M}{KL(1+\kappa)\ln(T/M)})$.
\end{Theorem}
Theorem 1 states that SGD with learning rate produced by our MLR-Net can obtain an approximately linear convergence rate,
achieving the best-known rates for the non-convex optimization \cite{karimi2016linear}. While the assumption (A2) means that all stationary points are optimal point, which is not always true for deep learning, the following theorem  discusses the case where the PL condition is not satisfied.

\begin{Theorem}
	Assume (\textbf{A1},\textbf{A3}) hold, and the SGD is with learning rate (\ref{eqlr}), where $\alpha_0 = (cL(1+\kappa))^{-1}, c>1$. Then for $w_t$ generated using SGD, we have the following bound
	\begin{align*}
	\min_t \mathbb{E} \|\nabla f(w_t) \|^2 &\!\leq\!  \frac{2cKL(1\!+\!\kappa)\ln(T/M)}{T-M} [\mathbb{E} f(w_{1}) \!-\! \mathbb{E} f(w_{T})] \\
	&+ \mathcal{O}\left(\frac{\sigma KT}{c(1+\kappa)(T-M)}\right).
	\end{align*}
\end{Theorem}
It can be seen that when $\sigma\neq 0$, if we set $c \propto \sqrt{T}$ and $\sigma = \mathcal{O}(1)$, it would give the $\mathcal{O}(1/\sqrt{T})$ rate; when $\sigma=0$, if we set $c =\mathcal{O}(1)$, it would give the $\mathcal{O}(1/T)$.
It is worth noting that the condition $\sigma=0$ holds in many practical scenarios, e.g.,
\cite{vaswani2019fast}.
On the other hand, we provide a convergence analysis of the MLR-SNet updated by the validation loss.
\begin{Theorem}
	Assume (\textbf{A1},\textbf{A3}) hold, $f$ has $\rho$-bounded gradients with respect to training/validation data, and the $\mathcal{A}(\theta)$ is differential with a $\delta$-bounded gradient and twice differential with its Hessian bounded by $\mathcal{B}$. Assume that the learning rate $\alpha_t=\mathcal{A}(\theta_t)$ predicted by MLR-SNet obey Eq.(\ref{eqlr}).
	We suppose that the learning rate of Adam algorithm for updating MLR-SNet satisfies $\eta_t=\eta$ for all $t\in[T],\eta \leq \frac{\epsilon}{2L}$ and $1-\beta_2 \leq \frac{\epsilon^2}{16\rho^2}$, where $\beta_2,\epsilon$ are the hyperparameters of the Adam algorithm (It can be found in Appendix). Then for $\theta_t$ generated using Adam, we have the following bound:
	\begin{align}
	\min_{0\leq t \leq T} \mathbb{E}[ \|\nabla \mathcal{L}_{Val}(\hat{\mathbf{w}}_{t}(\theta_{t}))\|_2^2] \leq \mathcal{O}(\frac{1}{c^2\ln(T)}+\sigma^2 ).
	\end{align}
\end{Theorem}
It can be seen that when $\sigma\neq 0$, if we set $c \propto \sqrt{T}$, and $\sigma = \mathcal{O}(1)$, it would lead to the $\mathcal{O}(\frac{1}{T\ln(T)}+\sigma^2)$ convergence rate; when $\sigma=0$, if we set $c =\mathcal{O}(1)$, it would give the $\mathcal{O}(\frac{1}{\ln(T)})$ convergence rate. It can then be proved that the convergence of the proposed method.

\begin{figure*}[t]\vspace{-2mm}
	\centering
	\subfigcapskip=-2mm
	\subfigure[Meta-training Results]{
		\includegraphics[width=0.24\textwidth]{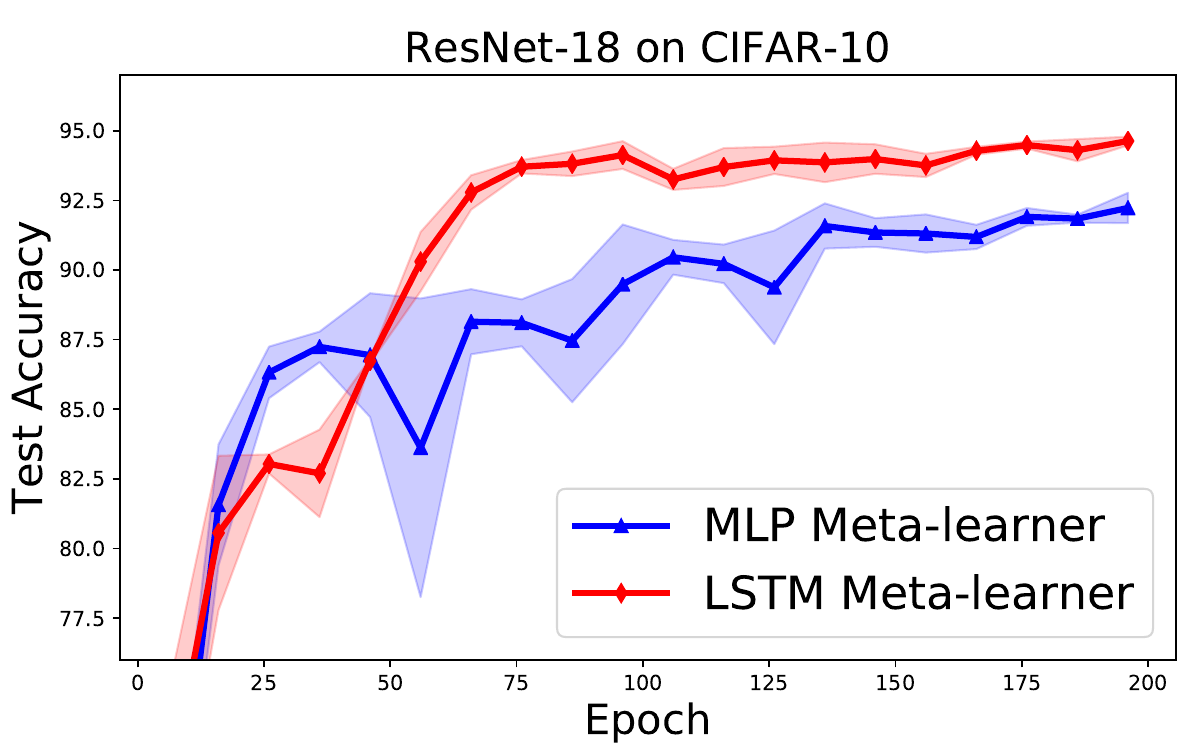}
		\includegraphics[width=0.24\textwidth]{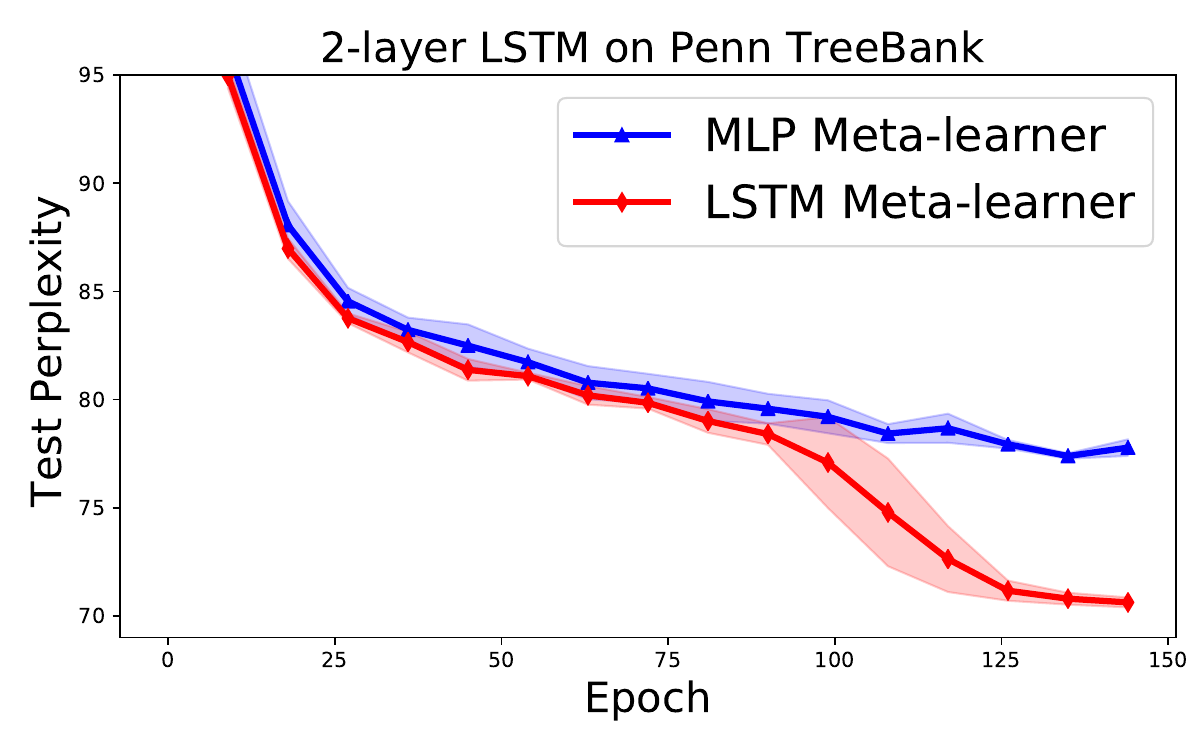}}
	\subfigure[Meta-test Results]{
		\includegraphics[width=0.24\textwidth]{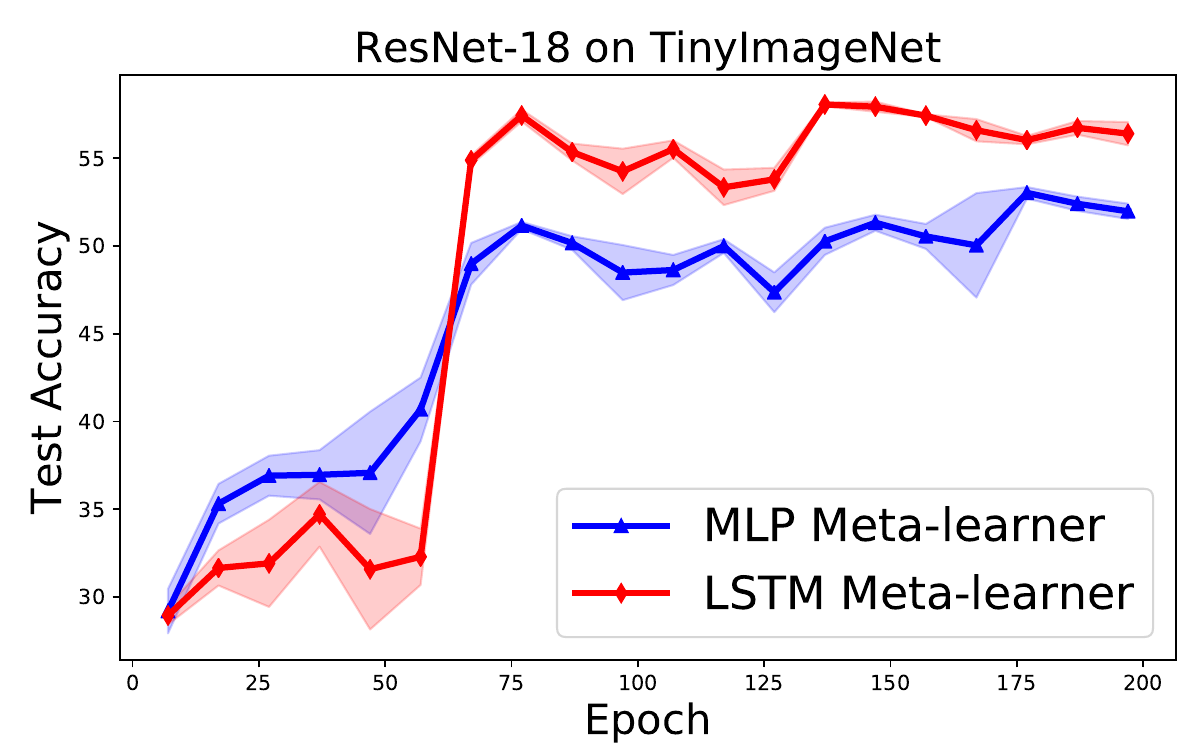}
		\includegraphics[width=0.24\textwidth]{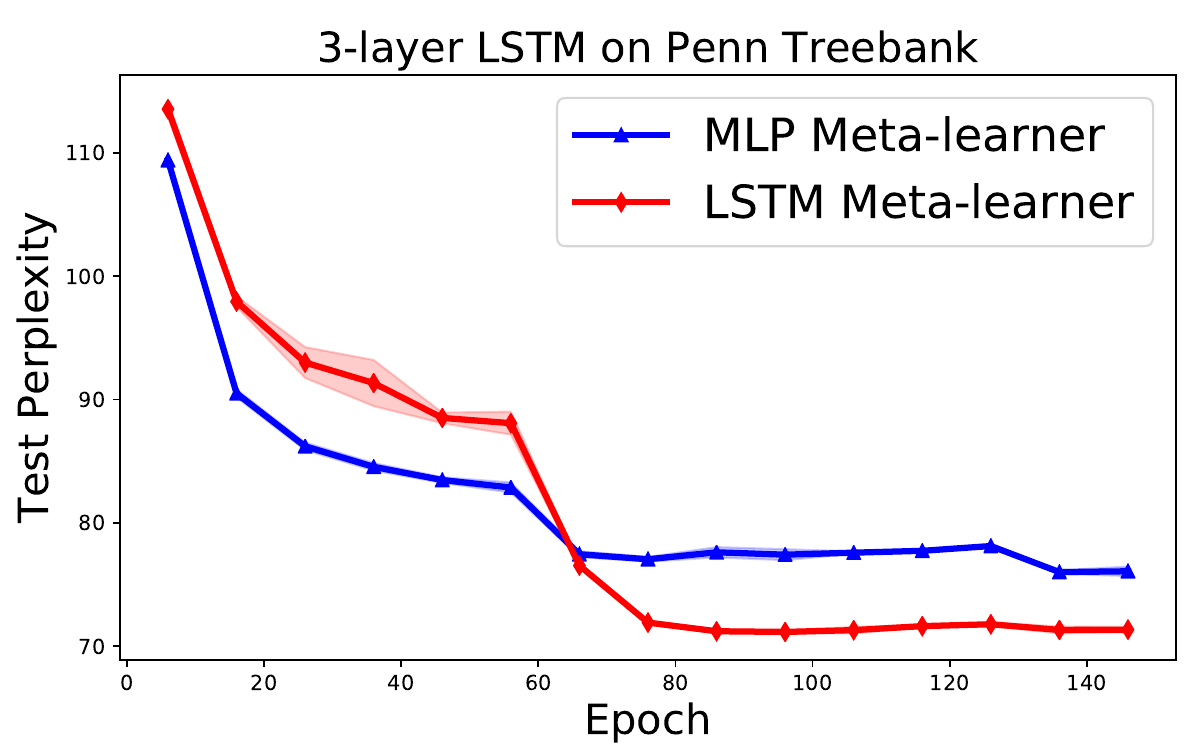}} \vspace{-2mm}
	\caption{Performance comparison of two types of meta-learners. (a) Two types of meta-learners are trained on CIFAR-10 and Penn Treebank datasets following the experiment setting in Section 4.1. The figure presents the test performance of two tasks. (b) The LR schedules meta-learned on CIFAR-10 is transferred to TinyImageNet and Penn Treebank datasets following the experiment setting in Section 4.2. The meta-test performance are shown in the figure.   }\label{figar}  \vspace{-4mm}
\end{figure*}

\subsection{Computational Complexity Analysis} \label{computation}
In the meta-training stage, our MLR-SNet learning algorithm can be roughly regarded as requiring two extra full forward and backward passes of the network (step 6 in Algorithm 1) in the presence of the normal network parameters update (step 8 in Algorithm 1), together with the forward passes of MLR-SNet for every LR. Therefore compared to normal training, our method needs about $3\times$ computation time for one iteration. Since we periodically update MLR-SNet after several iterations, this will not substantially increase the computational complexity compared with normal network training. In the meta-test stage, our transferred LR schedules predict LR for each iteration by a small MLR-SNet (step 4 in Algorithm 2), whose computational cost should be significantly less than the cost of the normal network training. To empirically show the computational complexity differences between baselines and our MLR-SNet, we conduct experiments with ResNet-18 on CIFAR-10 and report the running time for all methods. All experiments are implemented on a computer with Intel Xeon(R) CPU E5-2686 v4 and a NVIDIA GeForce RTX 2080 8GB GPU. We follow the corresponding settings in Section 4.1, and results are shown in Figure \ref{figtime}. It is seen that except that \textbf{RTHO} costs significantly more time, our MLR-SNet takes similar time to complete the meta-training and meta-test phase compared to hand-designed LR schedules. Considering its good transferability and generalization capability, it should be rational to say that it is efficient.

\begin{figure}[t]\vspace{-0mm}
	\centering
	\subfigcapskip=-0mm
	\subfigure[Meta-training Results]{
		\includegraphics[width=0.238\textwidth]{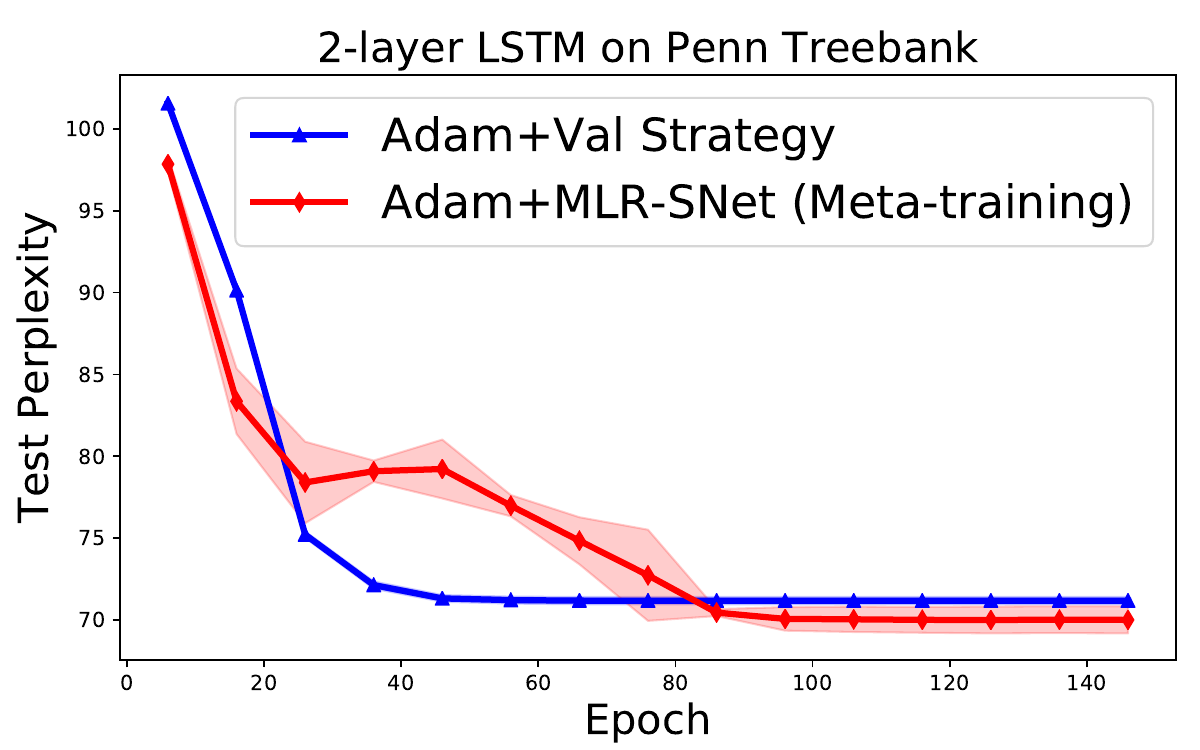} } \hspace{-4mm}
	\subfigure[Meta-test Results]{
		\includegraphics[width=0.238\textwidth]{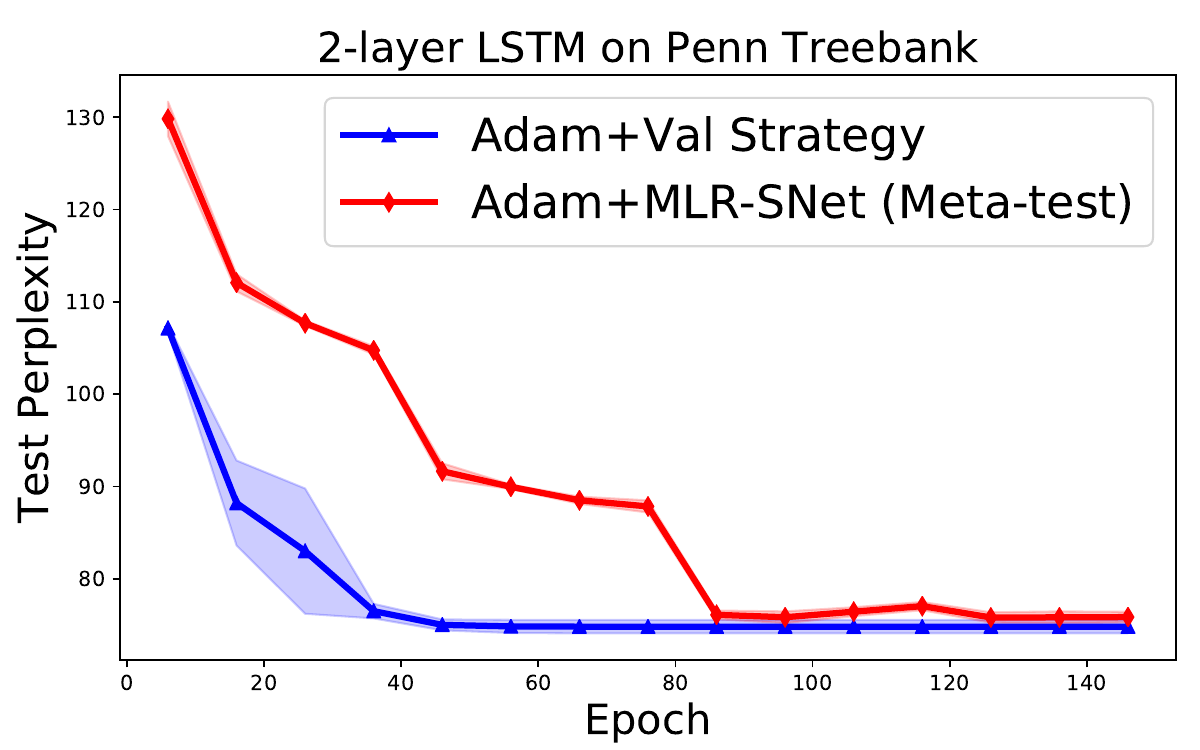} } \vspace{-4mm}
	\caption{Applying MLR-SNet on Top of Adam Algorithm. (a) The MLR-SNet for Adam is meta-trained on Penn Treebank datasets with 2-layer LSTM following the experiment setting in Section 4.1. The figure present the test perplexity. (b) The meta-learned LR schedules are transferred to train the 3-layer LSTM on Penn Treebank dataset. The test perplexity is depicted.   }\label{figadam}  \vspace{-4mm}
\end{figure}
\subsection{Visualizing the ``Width'' of Solutions}\label{width}
We further point out that visualizing the ``width'' of a given solution $w$ in a low-dimensional space may help understand why the model has fine generalization capability.
Generally, \cite{keskar2016large,dinh2017sharp} suggested that the wider optima leads to better generalization. We use the visualization technique in \cite{izmailovaveraging} to show how the
loss changes along many random directions drawn from the $d$-dimensional Gaussian distribution.
Fig.\ref{curve} visualizes the ``width'' of the solutions learned on CIFAR-100 with ResNet-18 for different LR schedules. It can be seen that our method, as well as the competitive baselines, lies a wide flat region of the train loss.
This could explain why they achieve better generalization performance. Deeper understandings on this point will be further investigated.

\subsection{Why Do We Need LSTM Meta-learner } \label{lstm-type}
We regard scheduling LR as a long-term information dependent problem, and thus we parameterize the LR schedules as an LSTM network. As we know MLP (multilayer perceptron) network can also learn an explicit mapping but ignores the temporal information, here we compare the performance of the two types of meta-learners. Fig. \ref{figar} compares the performance of two types of meta-learners for both meta-training and meta-test procedures. As is shown, the MLP meta-learner achieves better performance in the early learning stage for both meta-training and meta-test procedure. While at the later training stage, the LSTM meta-learner gradually brings a notable performance increase compared with MLP meta-learner. This might be possibly due to that the MLP meta-learner easily falls into the local optimal LR learning, while lacks of considering the overall significantly changed training dynamics. Though MLP meta-learner can also depict the loss-LR relationship, it ignores the more important training dynamics information involved for the scheduling LR. The LSTM meta-learner, however, is capable of accumulating temporal information of complicated training dynamics, and thus inclines to help find a more proper LR schedule for such DNNs training.

\subsection{Applying MLR-SNet on Top of Adam Algorithm}\label{adam}
To further demonstrate the versatility of our method, we apply the MLR-SNet on top of the Adam algorithm. Fig.\ref{figadam} shows that our method can help find better LR schedules than the Val Strategy. And the transferred LR schedules can also attain comparable performance with the hand-designed LR schedules. This implies that our framework is hopeful to learn the proper LR schedules for various optimizers.

\section{Conclusion and Discussion}
In this paper, we have proposed to learn an adaptive and transferrable LR schedule in a meta learning manner. To this aim, we have designed an LSTM-type meta-learner (MLR-SNet) to parameterize LR schedules, which gives more flexibility to adaptively learn a proper LR schedule to comply with the complex training dynamics of DNNs. Meanwhile, the meta-learned LR schedules are plug-and-play and transferrable, which can be readily transferred to schedule LR for SGD to new heterogeneous tasks.
Comprehensive experiments have been implemented, and the results substantiate the superiority of our method on various image and text benchmarks in its adaptability, transferability and robustness, as compared with current LR schedules policies. The MLR-SNet is hopeful to be useful in practical problems as it requires negligible increase in the parameter size and computation time, and small transferrable cost for new tasks. We will make further endeavor to further ameliorate our proposed method to make it as a general and useful tool for helping improve current DNN training. More practical applications will also be attempt to further verify its effectiveness in general learning tasks.

\bibliographystyle{IEEEtran}
\bibliography{IEEEabrv,mylib}

\appendix

\section{Convergence Analysis of the MLR-SNet}
\subsection{Proof of Theorem 1}
In the following we provide the proof details for the result of Theorem 1 in the maintext.
\begin{proof}
	Let $f^*$ be the infimum of $f(w)$, and then under the assumption A1, we have
	\begin{align}
	f(w_{t+1}) \leq f(w_t) -\langle \nabla f(w_t), \alpha_t v_t \rangle+\frac{L}{2}\alpha_t^2 \|v_t\|^2.
	\end{align}
	Taking expectation on both sides, we have
	\begin{align*}
	\mathbb{E} f(w_{t+1}) -\mathbb{E} f(w_{t})
	\leq&  - \langle \mathbb{E}\nabla f(w_t), \alpha_t \mathbb{E} v_t \rangle +\frac{L}{2}\alpha_t^2 \mathbb{E} \|v_t\|^2 \\
	= & -\alpha_t \mathbb{E} \|\nabla f(w_t) \|  +\frac{L}{2}\alpha_t^2 \mathbb{E} \|v_t\|^2.
	\end{align*}
	According to the assumption A3, it produces that
	\begin{align*}
	\mathbb{E} \|v_t\|^2 \leq (\kappa+1) \mathbb{E} \|\nabla f(w_t) \| + \sigma.
	\end{align*}
	Therefore, we have
	\begin{align}\label{eqlipsch}
	\begin{split}
	&\mathbb{E} f(w_{t+1}) -\mathbb{E} f(w_{t})\\
	\leq & -\alpha_t \mathbb{E} \|\nabla f(w_t) \| + \frac{L}{2}\alpha_t^2 [(\kappa+1) \mathbb{E} \|\nabla f(w_t) \|^2 + \sigma]\\
	=& -\left(\alpha_t - \frac{L(\kappa+1)}{2}\alpha_t^2 \right)\mathbb{E} \|\nabla f(w_t) \|^2 + \frac{L}{2}\alpha_t^2\sigma \\
	\leq & -\frac{1}{2} \alpha_t \mathbb{E} \|\nabla f(w_t) \|^2  + \frac{L}{2}\alpha_t^2\sigma,
	\end{split}
	\end{align}
	where the last inequality holds since $\alpha_t \leq \frac{1}{L(\kappa+1)}$.
	Let $\delta_t = \mathbb{E} f(w_t) -f^*$, and then we get
	\begin{align} \label{eqdelta}
	\delta_{t+1} \leq \delta_t -\frac{1}{2} \alpha_t \mathbb{E} \|\nabla f(w_t) \|^2  + \frac{L}{2}\alpha_t^2\sigma.
	\end{align}
	Based on the assumption A2, we can get $-\frac{1}{2}\|\nabla f(w_t)\|^2 \leq \mu \delta_t$. Now, Eq(\ref{eqdelta}) can be written as
	\begin{align*}
	&\ \ \delta_{T+1}
	\leq (1-\mu\alpha_T) \delta_T  + \frac{L}{2}\alpha_T^2\sigma\\
	& \leq (1-\mu\alpha_T)\left[ (1-\mu\alpha_{T-1}) \delta_{T-1} + \frac{L}{2}\alpha_{T-1}^2b\right] + \frac{L}{2}\alpha_T^2\sigma \\
	& =  (1-\mu\alpha_T)(1-\mu\alpha_{T-1}) \delta_{T-1} + \frac{L\sigma}{2} \left[(1-\mu\alpha_T)\alpha_{T-1}^2 + \alpha_T^2  \right] \\
	& = \cdots \\
	& = \prod_{t=1}^T(1-\mu\alpha_t) \delta_1 + \frac{L\sigma}{2} \sum_{t=1}^{T} \prod_{i=t+1}^{T} (1-\mu\alpha_i) \alpha_t^2.
	\end{align*}
	Since $1-\mu \alpha_t \leq \exp(-\mu \alpha_t), t=1,\cdots,T$, we have
	\begin{align} \label{eqsum}
	\begin{split}
	\delta_{T+1} \!\leq\!& \prod_{t=1}^{T} \exp(-\mu \alpha_t) \delta_1 \!+\! \frac{L\sigma}{2} \!\sum_{t=1}^{T}\!\prod_{i=t+1}^{T}\!  \exp(-\mu \alpha_i) \alpha_t^2 \\
	\!=\!& \exp(-\mu \sum_{t=1}^{T}\alpha_t) \delta_1 + \frac{L\sigma}{2} \!\sum_{t=1}^{T}\!  \exp(-\mu \!\sum_{i=t+1}^{T} \alpha_i\!) \alpha_t^2.
	\end{split}
	\end{align}
	Since $\alpha_t = \alpha_{t-1}\beta_t, 1/K \leq a \leq \beta_t$, then $\alpha_t \geq \alpha_0 a^t$,
	\begin{align*}
	\sum_{t=1}^T \alpha_t &\geq \alpha_0 \frac{a-a^{T+1}}{1-a} = \alpha_0  \frac{a(1-a^{T})}{1-a} \\
	&\geq \frac{\alpha_0}{K} \frac{1-a^T}{1-a} =  \frac{\alpha_0}{K} \frac{1-M/T}{1-a}\\
	&\geq \frac{\alpha_0}{K} \frac{1-M/T}{1/T \ln(T/M)} = \frac{\alpha_0 (T-M)}{K\ln(T/M)},
	\end{align*}
	where we use the result that
	$$\!1-x \!\leq\! \ln(1/x), \forall x$$ in the last inequallity. Thus we have
	\begin{align*}
	\begin{split}
	\exp(-\mu \sum_{t=1}^{T}\alpha_t)&\leq \exp\left(-\mu\alpha_0  \frac{T-M}{K\ln(T/M)}   \right)  \\
	& =C(M) \exp\left(-  \frac{\mu T}{KL(1+\kappa)\ln(T/M)}   \right),
	\end{split}
	\end{align*}
	where $C(M) = \exp(\frac{\mu M}{KL(1+\kappa)\ln(T/M)})$.	
	Observing that
	\begin{align*}
	\sum_{i=t+1}^{T} \alpha_i = \alpha_0 \frac{a^{t+1}-a^{T+1}}{1-a}\geq \frac{\alpha_0 T\left(a^{t}- a^{T}\right)}{K\ln(T/M)} ,
	\end{align*}
	we can deduce that
	\begin{align*}
	&\sum_{t=1}^{T}  \exp(-\mu \sum_{i=t+1}^{T} \alpha_i) \alpha_t^2
	\leq  \sum_{t=1}^{T}  \exp\left(-\mu \alpha_0 T \frac{a^{t}-a^T}{K\ln(T/M)} \right) \alpha_t^2 \\
	= & C(M)\sum_{t=1}^{T} \exp\left(\frac{-\mu \alpha_0 T a^t}{K\ln(T/M)} \right) \alpha_t^2 \\
	\leq &  C(M)\sum_{t=1}^{T}  \left(\frac{2K\ln(T/M)}{e\mu\alpha_0 a^{t}T}\right)^2 \alpha^2_t  \\
	\leq & C(M)\sum_{t=1}^{T}  \left(\frac{2K\ln(T/M)}{e\mu\alpha_0 a^{t}T}\right)^2 \alpha_0^2 b^{2t}  \\
	= & 4K^2C(M)\sum_{t=1}^{T} \frac{\ln^2(T/M)}{e^2\mu^2  T^2} (N/M)^{2t/T} \\
	= &  \frac{4K^2C(M)\ln^2(T/M)}{e^2\mu^2 T^2}  \frac{(N/M)^{2/T}- (N/M)^{2+2/T}}{1-(N/M)^{2/T}}. \\
	\leq &  \frac{4K^2C(M)\ln^2(T/M)}{e^2\mu^2 T^2} \frac{(N/M)^{2+2/T}}{(N/M)^{2/T}-1} \\
	= & \frac{4K^2C(M) \ln^2(T/M)}{e^2\mu^2 T^2} \frac{(N/M)^2}{1-(M/N)^{2/T}} \\
	\leq & \frac{4K^2C(M)\ln^2(T/M)}{e^2\mu^2 T^2} \frac{T(N/M)^2}{2-2M/N}  \\
	=&  \frac{2K^2C(M)\ln^2(T/M)(N/M)^2}{e^2\mu^2(1-M/N) T},
	\end{align*}
	where the second inequality holds since $\exp(-x)\!\leq\! (s/ex)^s$, $\forall x>0, \forall s>0$, and 	
	the last inequality is based on the Bernoulli inequality
	$(M/N)^{2/T} \!=\! (1+M/N-1)^{2/T} \!\leq\! 1+$\ $ \frac{2M/N-2}{T}$.
	Putting all above results together, Eq.(\ref{eqsum}) can be bounded by
	\begin{align*}
	\delta_{T+1} \leq &  C(M) \exp\left(-  \frac{\mu T}{KL(1+\kappa)\ln(T/M)}\right)\delta_1 \\
	+ &\frac{2K^2C(M)\ln^2(T/M)(N/M)^2}{e^2\mu^2(1-M/N) T}.
	\end{align*}
	Thus the conclusion holds.
\end{proof}

\subsection{Proof of Theorem 2}
In the following we provide the proof details for the result of Theorem 2 in the maintext.
\begin{proof}
	According to the proof process of Theorem 1, under the assumption \textbf{A1},\textbf{A2} and the setting that $\alpha_0= \frac{1}{L(1+\kappa)}$, it can be deduced that Eq.(\ref{eqlipsch})  holds, i.e.,
	\begin{align}
	&\mathbb{E} f(w_{t+1}) -\mathbb{E} f(w_{t}) \leq
	-\frac{1}{2} \alpha_t \mathbb{E} \|\nabla f(w_t) \|^2  + \frac{L}{2}\alpha_t^2\sigma.
	\end{align}
	Summing up above inequalities over $t=1,2,\cdots,T$, and rearranging the terms, we can obtain
	\begin{align*}
	\frac{1}{2}\sum_{t=1}^T \alpha_t \mathbb{E} \|\nabla f(w_t) \|^2 \leq \mathbb{E} f(w_{1}) - \mathbb{E} f(w_{T}) + \frac{L\sigma}{2}\sum_{t=1}^T \alpha_t^2.
	\end{align*}
	Thus, we can deduce that
	
	\begin{align*}
	\min_{0\leq t \leq T} \mathbb{E} \|\nabla f(w_t) \|^2 \leq & \frac{\sum_{t=1}^T \alpha_t \mathbb{E} \|\nabla f(w_t) \|^2}{\sum_{t=1}^T \alpha_t} \\
	\leq & \frac{2\mathbb{E} f(w_{1}) - 2\mathbb{E} f(w_{T}) + L\sigma\sum_{t=1}^T \alpha_t^2}{\sum_{t=1}^T \alpha_t}.
	\end{align*}
	Observing that
	\begin{align}\label{eqsumat}
	\begin{split}
	\sum_{t=1}^T \alpha_t^2
	\leq &\sum_{t=1}^T \alpha_0^2 b^{2t} = \alpha_0^2 \frac{b^2-b^{2T+2}}{1-b^2}  \\
	\leq &\alpha_0^2 \frac{1-b^{2T}}{1-b^2}
	= \alpha_0^2 \frac{1-(N/T)^{2}}{1-(N/T)^{2/T}}\\ = &\alpha_0^2 \frac{1-(N/T)^{2}}{1-\exp(2/T\ln(N/T))}\\
	\leq & \alpha_0^2 \frac{2\ln(T/N)}{1-1/(1-2/T\ln(N/T))}\\
	=& \alpha_0^2\left(T+2\ln(T/N)\right) = \frac{T+2\ln(T/N)}{c^2L^2(1+\kappa)^2},
	\end{split}
	\end{align}
	where the last inequality holds since $\exp(x)\leq 1/(1-x), \forall x<1$. Recall the following intermediate result of the proof in Theorem 1,
	\begin{align*}
	\sum_{t=1}^T \alpha_t \geq & \frac{\alpha_0 (T-M)}{K\ln(T/M)}= \frac{ T-M}{cKL(1+\kappa)\ln(T/M)},
	\end{align*}
	we can then obtain
	\begin{align*}
	\min_{0\leq t \leq T} \mathbb{E} \|\nabla f(w_t) \|^2 &\leq  \frac{2cKL(1+\kappa)\ln(T/M))}{T-M} \\ [\mathbb{E} f(w_{1}) - \mathbb{E} f(w_{T})]
	&+ \mathcal{O}\left(\frac{\sigma K T}{c(1+\kappa)(T-M)}\right).
	\end{align*}	Thus the conclusion holds.
\end{proof}

\subsection{Proof of Theorem 3}
In the following we provide the proof details for the result of Theorem 3 in the maintext. First we need prove a necessary lemma as follows:
\begin{Lemma}
	Suppose that the loss function $f$ is Lipschitz smooth with respect to the model parameter $w$ with constant $L$, and has $\rho$-bounded gradients with respect to the training/validation data. And the $\mathcal{A}(\theta)$ is differential with a $\delta$-bounded gradient and twice differential with its Hessian bounded by $\mathcal{B}$.
	Then it holds that the gradient of MLR-SNet parameter $\theta$ with respect to the loss is also Lipschitz smooth.
\end{Lemma}
\begin{proof}
	The gradient of MLR-SNet parameter $\theta$ with respect to the loss at data point $j$ can be written as
	\begin{small}
		\begin{align*}
		\nabla_{\theta} f_j (\hat{w}_{t}(\theta))|_{\theta_t} &= \frac{\partial f_j(\hat{w}_{t}(\theta)) }{\partial \hat{w}_{t}(\theta) } \frac{\partial \hat{w}_{t}(\theta) }{\partial \mathcal{A}(\theta)} \frac{\partial \mathcal{A}(\theta)}{\partial \theta} \\
		&= \frac{-\alpha_t}{n} \sum_{i=1}^n \left(\frac{\partial f_j (\hat{w}_{t}(\theta))}{\partial \hat{w}_{t}(\theta)} \frac{\partial \ell_i(w_t)}{\partial w_t} \right)\frac{\partial \mathcal{A}(\theta)}{\partial \theta}\big|_{\theta_t},
		\end{align*}
	\end{small}
	
	Let $G_{ij} =\frac{\partial \ell_j (\hat{w}_{t}(\theta))}{\partial \hat{w}_{t}(\theta)} \frac{\partial \ell_i(w_t)}{\partial w_t}$,  and then take gradient of $\theta$ in both sides of the above equality. We then have
	\begin{footnotesize}
		\begin{align}
		\nabla^2_{\theta^2} f_j (\hat{w}_{t}(\theta))|_{\theta_t} &=\frac{-\alpha_t}{n} \sum_{i=1}^n \left[ \frac{\partial G_{ij}}{\partial \theta}\frac{\partial \mathcal{A}(\theta)}{\partial \theta} + G_{ij} \frac{\partial \mathcal{A}^2(\theta)}{\partial \theta^2}   \right].
		\end{align}
	\end{footnotesize}
	For the first term in the right hand side, we have that
	\begin{tiny}
		\begin{align}\label{eqlemma1}
		\begin{split}
		&\left\|  \frac{\partial G_{ij}}{\partial \theta}\frac{\partial \mathcal{A}(\theta)}{\partial \theta}   \right\|  \leq \delta \left\| \frac{\partial f_j (\hat{w}_{t}(\theta))}{\partial \hat{w}_{t}(\theta) \partial \theta} \frac{\partial f_i(w_t)}{\partial w_t}  \right\| \\
		=&  \delta \left\|  \frac{\partial }{\partial \hat{w}_{t}(\theta)}  \left(\frac{-\alpha_t}{n} \sum_{i=1}^n \left(\frac{\partial f_j (\hat{w}_{t}(\theta))}{\partial \hat{w}_{t}(\theta)} \frac{\partial f_i(w_t)}{\partial w_t} \right)\frac{\partial \mathcal{A}(\theta)}{\partial \theta}\big|_{\theta_t} \right)     \frac{\partial f_i(w_t)}{\partial w_t}            \right\| \\
		=& \delta \left\|   \left(\frac{-\alpha_t}{n} \sum_{i=1}^n \left(\frac{\partial^2 f_j (\hat{w}_{t}(\theta))}{\partial \hat{w}_{t}^2(\theta)} \frac{\partial f_i(w_t)}{\partial w_t} \right)\frac{\partial \mathcal{A}(\theta)}{\partial \theta}\big|_{\theta_t} \right)     \frac{\partial f_i(w_t)}{\partial w_t}            \right\|  \leq  \alpha_t L\rho^2 \delta^2.
		\end{split}
		\end{align}
	\end{tiny}
	For the second term in the right hand side, we have that
	\begin{align}\label{eqlemma2}
	\left\| G_{ij} \frac{\partial \mathcal{A}^2(\theta)}{\partial \theta^2}  \right\| \leq \mathcal{B}\rho^2.
	\end{align}
	Combining the above two inequalities Eq.(\ref{eqlemma1}) and (\ref{eqlemma2}), we have
	\begin{align}
	\left\| \nabla_{\theta} f_j (\hat{w}_{t}(\theta))|_{\theta_t}  \right\| \leq \alpha\rho^2(\alpha_tL\delta^2 + \mathcal{B}).
	\end{align}
	Define $L_A = \alpha\rho^2(\alpha_tL\delta^2 + \mathcal{B})$, and based on the Lagrange mean value theorem, we have:
	\begin{align}
	\left\| \nabla f^{Val} (\hat{\mathbf{w}}_{t}(\theta_1)) - f^{Val} (\hat{\mathbf{w}}_{t}(\theta_2))   \right\| \leq L_A \left\| \theta_1 - \theta_2\right\|.
	\end{align}
	Thus the conclusion holds.
\end{proof}

\begin{algorithm}[t]
	\vspace{0mm}
	\renewcommand{\algorithmicrequire}{\textbf{Input:}}
	\renewcommand{\algorithmicensure}{\textbf{Output:}}
	\caption{Adam Algorithm}
	\label{algadam}
	\begin{algorithmic}[1]  \small
		\REQUIRE  $\theta_1  \in \mathbb{R}^{d'}$, learning rate $\{\eta_t\}_{t=1}^T$, decay parameters $0 \leq \beta_1,\beta_2 \leq 1, \epsilon>0$.
		\ENSURE  MLR-SNet parameter $\theta_T$
		\STATE Set $m_0=0,v_0=0$.
		\FOR{$t=0$ {\bfseries to} $T-1$}
		\STATE $D_n \leftarrow$ SampleMiniBatch($D_{Val},n$).
		\STATE Compute $g_t = \nabla_{\theta} f^{Val}(D_n,\theta_t)$.
		\STATE $m_t=\beta_1 m_{t-1}+(1-\beta_1)g_t$
		\STATE $v_t=v_{t-1}-(1-\beta_2)(v_{t-1}-g_t^2)$
		\STATE $\theta_{t+1}=\theta_t - \eta_t m_t/(\sqrt{v_t}+\epsilon)$
		\ENDFOR
	\end{algorithmic}
\end{algorithm}

Now we present the proof of Theorem 3.
\begin{proof}
	Suppose that we have a small validation set with $B$ samples $\{x_1,x_2,\cdots,x_M\}$, each associating with a validation loss function $\ell_i(w(\theta))$, where $w$ is the parameter of the model, and $\theta$ is the parameter of the MLR-SNet. The overall validation loss is then:
	\begin{align}
	f^{Val}(w) = \frac{1}{B} \sum_{i=1}^B f^{Val}_i (w(\theta)),
	\end{align}
	where $B$ is the minibatch size.
	According to the updating Algorithm 1, we have:
	\begin{align}  \label{eqval}
	\begin{split}
	&\mathbb{E}  f^{Val}(\hat{w}_{t+1}(\theta_{t+1}))-\mathbb{E} f^{Val}(\hat{w}_{t}(\theta_{t})) \\
	=&  \underbrace{\left\{\mathbb{E} f^{Val}(\hat{w}_{t+1}(\theta_{t+1})) - \mathbb{E} f^{Val}(\hat{w}_{t}(\theta_{t+1}))\right\}}_{(a)} \\ & + \underbrace{\left\{\mathbb{E}f^{Val}(\hat{w}_{t}(\theta_{t+1}))- \mathbb{E}f^{Val}(\hat{w}_{t}(\theta_{t}))\right\}}_{(b)}.
	\end{split}
	\end{align}
	For the above term (a), it holds that
	\begin{align} \label{app1}
	\begin{split}
	&\mathbb{E} f^{Val}(\hat{w}_{t+1}(\theta_{t+1})) - \mathbb{E} f^{Val}(\hat{w}_{t}(\theta_{t+1})) \\
	\leq & \!\left \langle \!\mathbb{E}\nabla_{w}f^{Val} (\hat{w}_{t+1}(\theta_{t+1})), \mathbb{E}\hat{w}_{t+1}(\theta_{t+1})- \mathbb{E}\hat{w}_{t}(\theta_{t+1}) \!\right\rangle\! \\ & +\frac{L}{2} \mathbb{E}\left\| \hat{w}_{t+1}(\theta_{t+1})- \hat{w}_{t}(\theta_{t+1}) \right\|_2^2.
	\end{split}
	\end{align}
	According to Eq (7) in the maintext, we have
	\begin{align*}
	\hat{w}_{t+1}(\theta_{t+1})- \hat{w}_{t}(\theta_{t+1}) = - \alpha_t\nabla_{w} f^{Tr}(\hat{w}_{t}(\theta_{t+1})).
	\end{align*}
	Then Eq (\ref{app1}) can be written as
	\begin{align*}
	a \leq & - \langle \mathbb{E}\nabla_{w}f^{Val} (\hat{w}_{t+1}(\theta_{t+1})), \alpha_t \mathbb{E} v_t \rangle +\frac{L}{2}\alpha_t^2 \mathbb{E} \|v_t\|^2 \\
	\leq&  - \langle \mathbb{E}\nabla_{w}f^{Val} (\hat{w}_{t+1}), \alpha_t \mathbb{E} v_t \rangle  +\frac{L}{2}\alpha_t^2  [(\kappa+1) \mathbb{E} \|\nabla f(w_t) \|^2 + \sigma] \\
	\leq & \alpha_t \rho^2  +\frac{L}{2}\alpha_t^2 [(1+\kappa)\rho^2+\sigma].
	\end{align*}
	%
	For the term (b) in Eq. (\ref{eqval}), according to Lemma 1, i.e., the validation loss is Lipschitz smooth with respect to the MLR-SNet parameter $\theta$ with $L$, we have
	\begin{align} \label{eqtheorem1}
	\begin{split}
	& \mathbb{E}f^{Val}(\hat{w}_{t}(\theta_{t+1}))- \mathbb{E}f^{Val}(\hat{w}_{t}(\theta_{t})) \\
	\!\leq &\! \left\langle\! \mathbb{E}\nabla_{\theta}f^{Val} (\hat{w}_{t}(\theta_{t})), \mathbb{E}\theta_{t+1}-\mathbb{E}\theta_{t}  \!\right\rangle\! + \frac{L}{2}\mathbb{E} \left\| \theta_{t+1}\!-\!\theta_{t}  \right\|_2^2.
	\end{split}
	\end{align}
	
	Here we adopt Adam algorithm \cite{kingma2015adam} (Algorithm \ref{algadam}) to update the parameter of MLR-SNet, $\theta_{t+1}-\theta_t$ in Eq.(\ref{eqtheorem1}) is updated by
	\begin{align}
	\theta_{t+1,i}= \theta_{t,i} -\eta_t \frac{g_{t,i}}{\sqrt{v_{t,i}}+\epsilon}, i=1,2,\cdots,d.
	\end{align}
	Now, we have
	\begin{align}
	\begin{split}
	b\leq & -\eta_t \sum_{i=1}^d \left\langle \mathbb{E}\nabla_{\theta}\mathcal{L}^i_{Val} (\hat{w}_{t}(\theta_{t})),  \mathbb{E}\frac{g_{t,i}}{\sqrt{v_{t,i}}+\epsilon} \right\rangle \\ & + \frac{L\eta_t^2}{2} \mathbb{E}\sum_{i=1}^d \frac{g_{t,i}^2}{(\sqrt{v_{t,i}}+\epsilon)^2}.
	\end{split}
	\end{align}
	Based on the proof process in \cite{zaheer2018adaptive} (Eq. (4) in pp. 13), we can deduce that
	\begin{align}
	\begin{split}
	b\leq &-\frac{\eta_t}{2(\sqrt{\beta_2}\rho+\epsilon)}\mathbb{E} \|\nabla_{\theta}f^{Val} (\hat{w}_{t}(\theta_{t}))\|_2^2 \\ & + \left(\frac{\eta \rho\sqrt{1-\beta_2}}{\epsilon^2}+ \frac{L\eta^2}{2\epsilon^2}\right) \frac{\sigma^2}{B}.
	\end{split}
	\end{align}
	Now Eq.(\ref{eqval}) can be reformulated as:
	\begin{small}
		\begin{align}\label{eqtheorem3}
		\begin{split}
		&\mathbb{E}  f^{Val}(\hat{w}_{t+1}(\theta_{t+1}))-\mathbb{E} f^{Val}(\hat{w}_{t}(\theta_{t})) \\ & \leq \alpha_t \rho^2  +\frac{L}{2}\alpha_t^2 [(1+\kappa)\rho^2+\sigma]-\frac{\eta_t}{2(\sqrt{\beta_2}\rho+\epsilon)} \\
		&   \mathbb{E}\|\nabla_{\theta}f^{Val} (\hat{w}_{t}(\theta_{t}))\|_2^2 + \left(\frac{\eta \rho\sqrt{1-\beta_2}}{\epsilon^2}+ \frac{L\eta^2}{2\epsilon^2}\right) \frac{\sigma^2}{B},
		\end{split}
		\end{align}
	\end{small}
	By rearranging the inequality (\ref{eqtheorem3}), we can then obtain:
	\begin{align*}
	&\mathbb{E}\left[\frac{\eta_t}{2(\sqrt{\beta_2}\rho+\epsilon)}\left\|\nabla_{\theta}\mathcal{L}_{Val} (\hat{w}_{t}(\theta_{t})) \right\|_2^2\right] \\
	\leq & \alpha_t \rho^2+ \frac{L}{2}\alpha_{t}^2 (\rho^2 +\sigma^2) -\mathbb{E}  f^{Val}(\hat{w}_{t+1}(\theta_{t+1})) \\ & + \mathbb{E} f^{Val}(\hat{w}_{t}(\theta_{t}))+ \left(\frac{\eta \rho\sqrt{1-\beta_2}}{\epsilon^2}+ \frac{L\eta^2}{2\epsilon^2}\right) \frac{\sigma^2}{B}.
	\end{align*}
	Using telscoping sum, we obtain
	\begin{small}
		\begin{align}
		\begin{split}
		&\sum_{t=1}^T \frac{\eta_t}{2(\sqrt{\beta_2}\rho+\epsilon)} \mathbb{E}\left\|\nabla_{\theta}f^{Val} (\hat{w}_{t}(\theta_{t})) \right\|_2^2 \\
		\leq  &\mathbb{E}f^{Val}(\hat{w}_{1}(\theta_1))-\mathbb{E}f^{Val}(\hat{w}_{T+1}(\theta_{T+1})) + \rho^2\sum_{t=1}^{T}\alpha_t \\ & + \frac{L}{2}(\rho^2 +\sigma^2)\sum_{t=1}^{T}\alpha_{t}^2 + \left(\frac{\eta \rho\sqrt{1-\beta_2}}{\epsilon^2}+ \frac{L\eta^2}{2\epsilon^2}\right) \frac{\sigma^2T}{B} \\
		\leq  &f^{Val}(\hat{w}_{1}(\theta_1))+ \rho^2\sum_{t=1}^{T}\alpha_t + \frac{L}{2}(\rho^2 +\sigma^2)\sum_{t=1}^{T}\alpha_{t}^2 \\ & + \left(\frac{\eta \rho\sqrt{1-\beta_2}}{\epsilon^2}+ \frac{L\eta^2}{2\epsilon^2}\right) \frac{\sigma^2T}{B}.
		\end{split}
		\end{align}
	\end{small}
	Therefore,
	\begin{align*}
	&\min_{t} \mathbb{E} \left[\left\|\nabla_{\theta}f^{Val} (\hat{w}_{t}(\theta_{t})) \right\|_2^2\right] \\ \leq & \frac{\sum_{t=1}^T \frac{\eta_t}{2(\sqrt{\beta_2}\rho+\epsilon)}\mathbb{E}\left\|\nabla_{\theta}f^{Val} (\hat{w}_{t}(\theta^{(t)})) \right\|_2^2}{\sum_{t=1}^T \frac{\eta_t}{2(\sqrt{\beta_2}\rho+\epsilon)}} \\
	\leq & \frac{f^{Val}(\hat{w}_{1}(\theta_1))-f^{Val}(\hat{w}_{T+1}(\theta_{T+1}))+ S}{1/2(\sqrt{\beta_2}\rho+\epsilon) \times \sum_{t=1}^{T}\eta_t} \\
	\leq & \frac{2(\sqrt{\beta_2}\rho+\epsilon)}{T \eta} \times\left\{ f^{Val}(\hat{w}_{1}(\theta_1)) +S \right\} ,
	\end{align*}
	where
	$S\!=\!\frac{L}{2}(\rho^2 +\sigma^2)\sum_{t=1}^{T}\alpha_{t}^2++\left(\frac{\eta \rho\sqrt{1-\beta_2}}{\epsilon^2}+ \frac{L\eta^2}{2\epsilon^2}\right) \frac{\sigma^2T}{B}+$\ $\rho^2\sum_{t=1}^{T}\alpha_t $. Taking a similar process as in Eq.(\ref{eqsumat}), we have that
	\begin{align*}
	\sum_{t=1}^T \alpha_t \leq & \frac{\ln(T/N)+T}{cL(1+\kappa)\ln(T/N)},\\
	\sum_{t=1}^T \alpha_t^2 \leq & \frac{2\ln(T/N)+T}{c^2L^2(1+\kappa)^2\ln(T/N)}.
	\end{align*}
	Therefore, we can obtain
	\begin{align*}
	\min_{t} \mathbb{E} \left\|\nabla_{\theta}f^{Val} (\hat{w}_{t}(\theta_{t})) \right\|_2^2\leq \mathcal{O}(\frac{1}{c^2\ln(T)}+\sigma^2 )
	\end{align*}
	Thus the conclusion holds.
\end{proof}

\newpage
\section{Pytorch implementation of MLR-SNet}
Here we also demonstrate the pseudo-code of the MLR-SNet for Pytorch implementation as follows, to make readers easily reproduce our algorithm.

\begin{python}
class LSTMCell(nn.Module):
def __init__(self, num_inputs, hidden_size):
super(LSTMCell, self).__init__()
self.hidden_size = hidden_size
self.fc_i2h = nn.Sequential(
nn.Linear(num_inputs, hidden_size),
nn.ReLU(),
nn.Linear(hidden_size, 4 * hidden_size))
self.fc_h2h = nn.Sequential(
nn.Linear(hidden_size, hidden_size),
nn.ReLU(),
nn.Linear(hidden_size, 4 * hidden_size))
def forward(self, inputs, state):
hx, cx = state
i2h = self.fc_i2h(inputs)
h2h = self.fc_h2h(hx)
x = i2h + h2h
gates = x.split(self.hidden_size, 1)
in_gate = torch.sigmoid(gates[0])
forget_gate = torch.sigmoid(gates[1])
out_gate = torch.sigmoid(gates[2])
in_transform = torch.tanh(gates[3])
cx = forget_gate * cx + in_gate * in_transform
hx = out_gate * torch.tanh(cx)
return hx, cx

class MLRNet(nn.Module):
def __init__(self, num_layers, hidden_size):
super(MLRNet, self).__init__()
self.hidden_size = hidden_size
self.layer1 = LSTMCell(1, hidden_size)
self.layer2 = nn.Linear(hidden_size, 1)
def forward(self, x, gamma):
self.hx, self.cx =
self.layer1(x, (self.hx, self.cx))
x = self.hx
x = self.layer2(x)
out = torch.sigmoid(x)
return gamma * out
\end{python}

\end{document}